\documentclass[10pt,twocolumn,letterpaper]{article}

\usepackage[pagenumbers]{cvpr} 
\usepackage{algorithm}
\usepackage{algorithmic}
\usepackage{multirow}
\usepackage{colortbl}
\usepackage{amsmath}
\usepackage[dvipsnames]{xcolor}

\usepackage[pagebackref,breaklinks,colorlinks]{hyperref}

\def\paperID{13078} 
\def\confName{CVPR}
\def\confYear{2024}

\title{Selection of Distinct Morphologies \\to Divide \& Conquer Gigapixel Pathology Images}

\author{ Abubakr Shafique$^{1,2}$, Saghir Alfasly$^{1}$,  Areej Alsaafin$^{1,2}$, Peyman Nejat$^{1}$, Jibran A. Khan$^{1}$,  \\H.R.Tizhoosh$^{1,2}$\thanks{Corresponding author} \\
		\small $^{1}$Rhazes Lab, Department of Artificial Intelligence \& Informatics, Mayo Clinic, Rochester, MN, USA\\
          \small $^{2}$Kimia Lab, University of Waterloo, Waterloo, ON, Canada\\
        \tt\small \{shafique.abubakr, tizhoosh.hamid\}@mayo.edu\\\
	}	

\begin{document}
\maketitle
\begin{abstract}

Whole slide images (WSIs) are massive digital pathology files illustrating intricate tissue structures. Selecting a small, representative subset of patches from each WSI is essential yet challenging. Therefore, following the ``Divide \& Conquer'' approach becomes essential to facilitate WSI analysis including the classification and the WSI matching in computational pathology. To this end, we propose a novel method termed ``Selection of Distinct Morphologies'' (SDM) to choose a subset of WSI patches. The aim is to encompass all inherent morphological variations within a given WSI while simultaneously minimizing the number of selected patches to represent these variations, ensuring a compact yet comprehensive set of patches. This systematically curated patch set forms what we term a ``montage''. We assess the representativeness of the SDM montage across various public and private histopathology datasets. This is conducted by using the \emph{leave-one-out} WSI search and matching evaluation method, comparing it with the state-of-the-art Yottixel’s mosaic. SDM demonstrates remarkable efficacy across all datasets during its evaluation. Furthermore, SDM eliminates the necessity for empirical parameterization, a crucial aspect of Yottixel's mosaic, by inherently optimizing the selection process to capture the distinct morphological features within the WSI.
\footnote{The code is available at: \url{https://github.com/RhazesLab/histopathology-SDM}}
\end{abstract}    
\section{Introduction}
\label{sec:intro}

\begin{figure}[t]
\centerline{\includegraphics[width =  1\columnwidth]{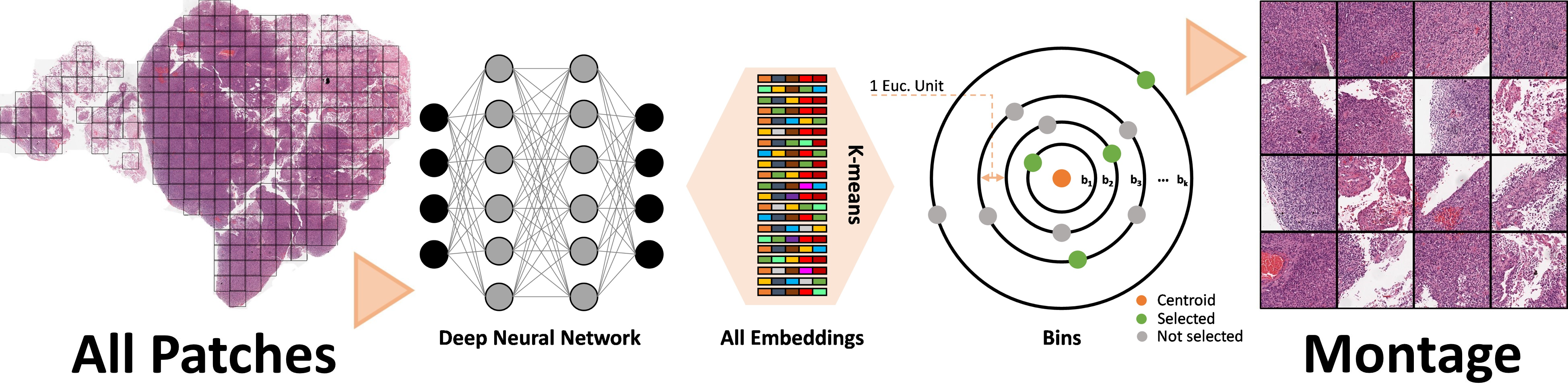}}
\caption{\textbf{Conceptual Overview.} Unsupervised Selection of Distinct Morphologies (SDM) through one-centroid clustering generates a montage to represent gigapixel WSI, enabling fast and efficient processing for downstream tasks in digital pathology. }
\label{fig:Montage}
\end{figure}

The progress in machine learning has demonstrated considerable potential in augmenting the efforts of healthcare practitioners~\cite{mukhopadhyay2018whole}. The emergence of digital pathology has opened new horizons for histopathology~\cite{baxi2022digital, niazi2019digital}. The volume of data encompassed within pathology archives is both remarkable and daunting in its scale~\cite{kalra2020Yottixel}. The representation of the whole slide images (WSIs) holds immense importance across a wide spectrum of applications within the fields of medicine and beyond~\cite{adnan2020representation, hemati2021, hemati2023learning}. WSIs are essentially high-resolution digital images that capture the entirety of a tissue sample, providing a comprehensive view of biopsy specimens under examination~\cite{shafique2021automatic}. Deep models, such as convolutional neural networks (CNNs)~\cite{he2016deep, xie2017aggregated, huang2017densely, howard2017mobilenets, tan2019efficientnet, liu2022convnet}, vision transformers (ViTs)~\cite{vaswani2017attention, dosovitskiy2020image, caron2021emerging} have been instrumental in extracting meaningful and interpretable features from the WSIs, leading to advanced applications in medicine~\cite{alsaafin2023learning}. Deep-learning-based representation of WSIs involves the use of neural networks to automatically learn hierarchical and abstract features from the vast amount of visual information contained in these high-resolution images~\cite{bidgoli2022evolutionary}. These learned representations enable computers to understand and interpret the complex structures and patterns present in human tissue. The applications of deep learning (DL) on WSI representations are diverse, ranging from automated disease diagnosis and prognosis prediction to drug discovery, telepathology consultations, and search \& matching techniques in content-based image retrieval (CBIR)~\cite{zheng2017size, li2018large, hegde2019, kalra2020pan, hemati2021, shafique2023preliminary}.

\begin{figure*}[ht]
\centerline{\includegraphics[width =  1\textwidth]{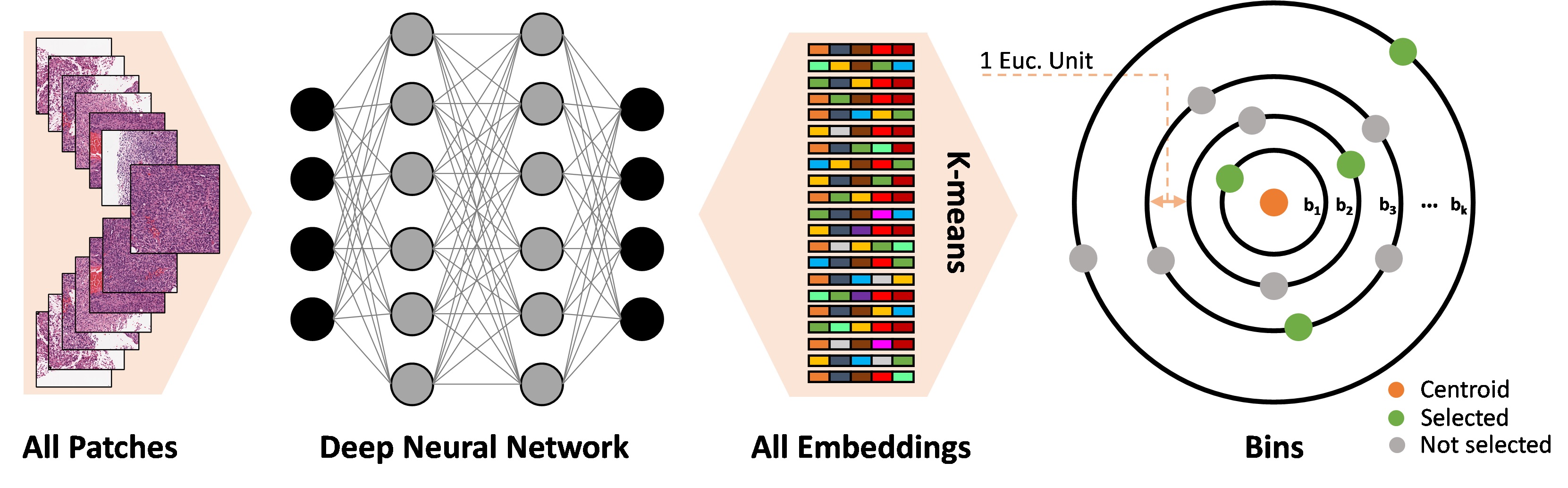}}
\caption{\textbf{The overall SDM process.} Commencing with the extraction of all patches from the WSI at low magnification (say at 2.5x), these patches subsequently undergo processing through a deep network (say DenseNet~\cite{huang2017densely}), resulting in the generation of embeddings for each patch. After obtaining all embeddings, k-means clustering is applied around a \emph{single centroid}, resulting in the calculation of the Euclidean distance of each patch from the centroid. Patches exhibiting similar Euclidean distances are organized into distinct Euclidean bins. Finally, one patch is selected from each bin to serve as part of the montage.}
\label{fig:SDM}
\end{figure*}

Second opinions (or \emph{consultations}) in histopathology are of paramount importance as they serve as a crucial quality control measure, enhancing diagnostic accuracy and reducing the risk of misdiagnosis, especially for rare or ambiguous cases~\cite{malarkey2015, tizhoosh2018, liu2019}. WSI search offers a valuable avenue for obtaining a virtual or computational second opinion~\cite{rasoolijaberi2022multi}. By leveraging advanced CBIR techniques, pathologists can compare a patient's WSI with a database of evidently diagnosed cases, aiding in the identification of similar patterns and anomalies~\cite{rasoolijaberi2022multi, hegde2019}. This approach provides a data-driven, objective perspective that complements the pathologist's evaluation, contributing to more reliable diagnoses and fostering an evidence-based approach to pathology~\cite{shafique2023immunohistochemistry, shafique2023preliminary}. The endeavor of conducting searches within an archive of gigapixel WSIs, akin to addressing large-scale big-data challenges, necessitates the design of a well-defined  ``Divide \& Conquer'' algorithm for WSIs~\cite{lin2018scannet, sharma2021cluster}.

\section{Related Work}
\label{sec:literature}

Despite the critical role of patch selection as an initial step in the analysis of WSI, this phase has not been extensively investigated. The predominant methods in the literature use brute force patching where the entire WSI is tiled into thousands of patches~\cite{lu2021data, shao2021transmil, hegde2019similar}. Leveraging the entirety of patches extracted from the WSI for retrieval tasks is computationally prohibitive for practical applications due to the substantial processing resources required. In the literature, the search engine Yottixel introduced an unsupervised cluster-based patching technique called \emph{mosaic}~\cite{kalra2020Yottixel}. Yottixel's mosaic functions as a pivotal component during the primary ``Divide" stage, effectively partitioning the formidable task of processing WSIs into discrete, manageable parts, with each part represented by an individual patch within the mosaic~\cite{kalra2020Yottixel}. Other search engines in the field have not offered new patch selection methods and have used Yottixel's patching scheme to divide the WSI~\cite{chen2022fast, sikaroudi2023comments, wang2023retccl}.

While Yottixel's mosaic method stands as a cutting-edge unsupervised approach for patch selection in the literature, it does incorporate certain empirical parameters, including the utilization of 9 clusters for k-means clustering and the selection of 5\% to 20\% of the total patches within each of the k=9 clusters~\cite{kalra2020Yottixel}. Given the intricate nature of tissue morphology~\cite{bankhead2017qupath, bychkov2018deep}, it is plausible that there may exist more or less than nine distinct tissue groups in a WSI. As well, determining the proper level of cluster sampling may not be straightforward in an automated fashion. All these considerations underscore the urgent need for the development of novel unsupervised patch selection methodologies capable of comprehensively representing all the diverse aspects and characteristics inherent in a WSI for all types of biopsy.

In this work, a novel unsupervised patch selection methodology is introduced which comprehensively captures the discrete attributes of a WSI without necessitating any empirical parameter setting by the user. The new ``Selection of Distinct Morphologies'' (SDM) will be explained in the methods Section~\ref{sec:method}. Furthermore, the evaluation of the proposed method is described in Section~\ref{sec:results} followed by the discussion and conclusion Section~\ref{sec:discussion}.

\section{Methodology}
\label{sec:method}

\begin{algorithm*}[ht]
\caption{Creation of the montage through selection of distinct morphologies.}\label{alg:montage}
\begin{algorithmic}[1]
\REQUIRE WSI Image
\ENSURE Set of selected patches $P_s$ as output
\STATE  $m$  $\gets$  The lower magnification for patching
\STATE  $s$  $\gets$  The patch size at low magnification
\STATE  $t$  $\gets$  A minimum tissue threshold for each patch
\STATE  $o$  $\gets$  The overlap percentage between each adjacent patch
\STATE  \textbf{Procedure}
\STATE  $I_m$ $\gets$ OpenWSI($m$) \hfill $\rhd$ Open the WSI at lower magnification ($m$)
\STATE  $M_{m}$ $\gets$ TissueSegmentation ($I_m$) \hfill $\rhd$ Extract the tissue regions at lower magnification ($m$)
\STATE  $T$ $\gets$ Patching ($I_m$, $M_{m}$, $s$, $o$) \hfill $\rhd$ Perform dense patching with $s$ size and $o$ overlap
\FOR {each $T$}
\STATE  $G$ $\gets$ TissuePercentage ($T$)  \hfill $\rhd$ Calculate tissue percentage for each patch
\STATE  $P$ $\gets$ $T$ if $G > t$ \hfill $\rhd$ Filter the patches using the tissue threshold $t$ 
\ENDFOR
\STATE  $E$  $\gets$  GetEmbeddings($P$) \hfill $\rhd$ Feed the patches $P_t$ to a deep network
\STATE  $C, D$  $\gets$  k-means($E$) \hfill $\rhd$ Get the centroid and the Euclidean distances for all the patches
\STATE  $D_r$ $\gets$ Round($D$) \hfill $\rhd$ Round off the distances to the nearest integer
\STATE  $B$ $\gets$ Binned($D_r$) \hfill $\rhd$ Generate the bin for each integer distance
\STATE  $P_s$ $\gets$ $B$ \hfill $\rhd$ Select a patch from each bin
\STATE  \textbf{Return} $P_s$ \hfill $\rhd$ Return the final selection of distinct patches
\STATE  \textbf{End Procedure}
\end{algorithmic}
\end{algorithm*}

Although it is important to have comprehensive annotations for the WSIs, manual delineations for a large number of WSIs are prohibitively time-consuming or even infeasible. Therefore, in most scenarios, the utilization of \emph{unsupervised patching} becomes inevitable. For this reason, an unsupervised technique is introduced to represent all distinct features of a WSI using fewer patches, termed a ``montage''. Building such montages serves as a fundamental component crucial for facilitating numerous downstream WSI operations, image search being just one of them. Figure~\ref{fig:Montage} shows the steps for producing a montage from a WSI using the proposed SDM method.

\subsection{SDM: Selection of Distinct Morphologies}

\begin{figure}[t]
\centerline{\includegraphics[width =  1\columnwidth]{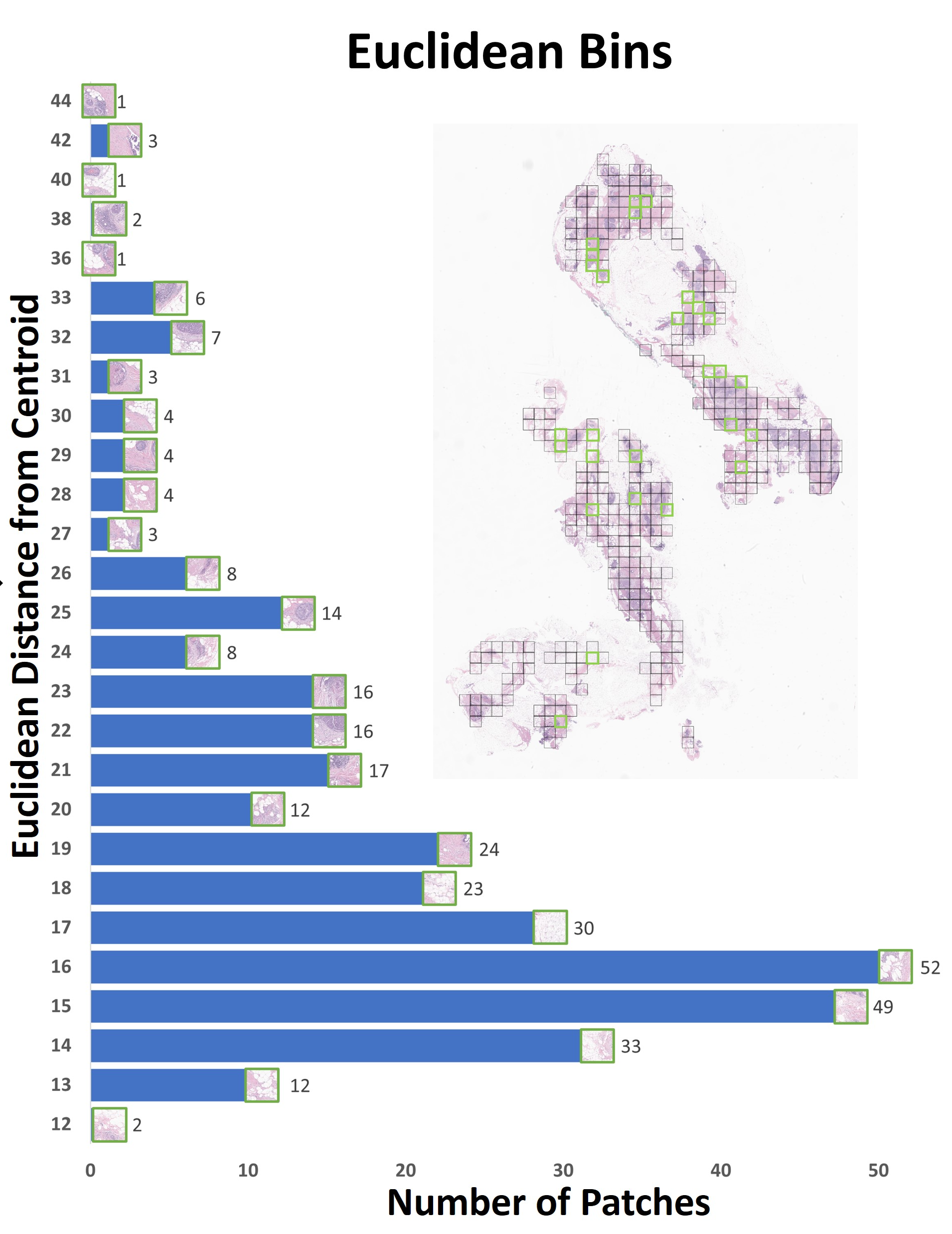}}
\caption{\textbf{Discrete Euclidean bins within SDM.} The bar chart visually represents the distribution of patches from the WSI across various Euclidean bins. Patches grouped within the same Euclidean bin exhibit similarity. Randomly selected patches (displayed at the top of each bin)  represent the montage.}
\label{fig:Euc_bins}
\end{figure}

Patch selection is a fundamental step in digital pathology for many computer-aided diagnosis (CAD) techniques, leading to enhanced diagnostic capabilities and improved patient care. To obtain representative patches that effectively represent the content of a WSI, the SDM is introduced in this work. SDM aims to create a ``montage'' comprising a rather small number of patches that exhibit diversity while maintaining their meaningfulness within the context of the WSI (see Figure~\ref{fig:Montage}). The SDM scheme for creating a montage is outlined in Algorithm~\ref{alg:montage}, and also illustrated in Figure~\ref{fig:SDM}.

Initially, we process the WSI $I_m$ at a low magnification level $m$, \eg, $m=2.5\times$. Tissue segmentation is performed to generate a binary tissue mask $M_{m}$, for instance using U-Net segmentation~\cite{riasatian2020}. According to the findings in the literature, a magnification of $2.5\times$ represents the minimum level at which it remains feasible to differentiate between tissue components and artifacts while also retaining some intricate details~\cite{riasatian2020}. Using the tissue mask $M_{m}$, dense patching is performed all over the tissue region to extract all the patches with patch size $s_l\times s_l$, and patch overlap $o$ at $2.5\times$. Empirically, we use $s_l=128$, $2.5\times$ magnification, and $o=5\%$. 

\begin{figure*}[t]
\centerline{\includegraphics[width =  1\textwidth]{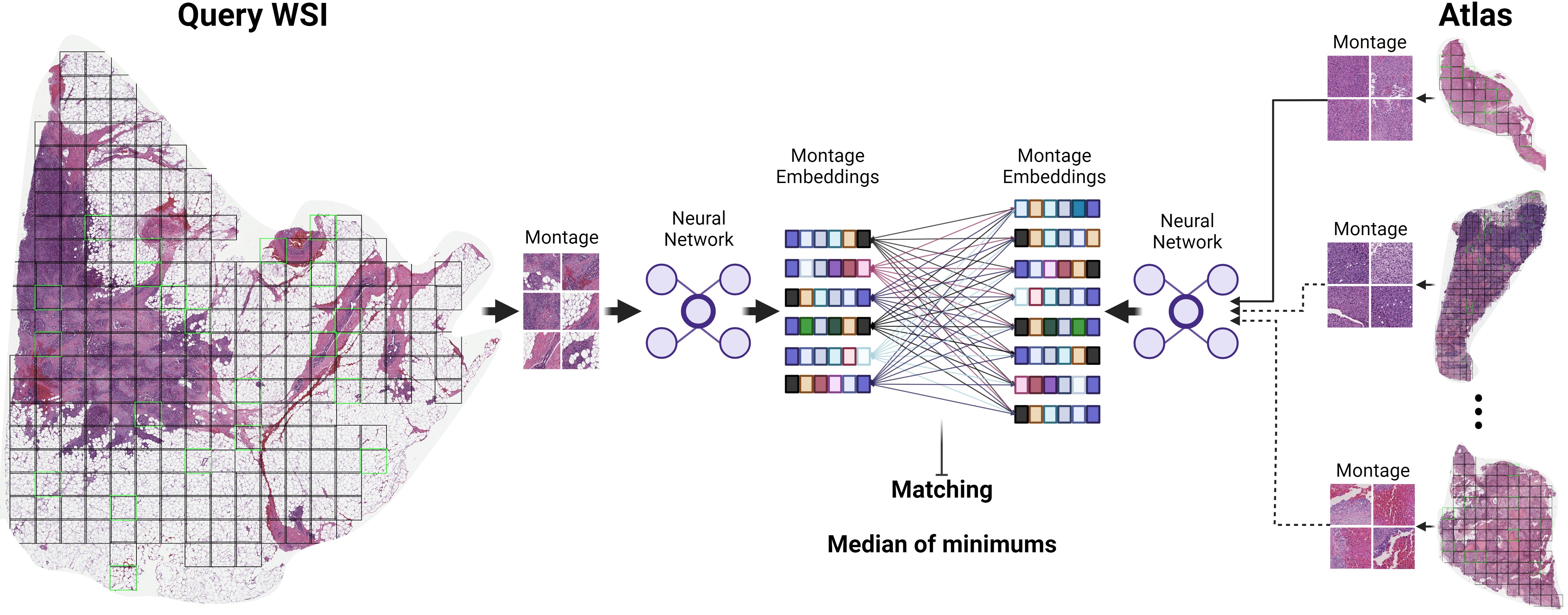}}
\caption{\textbf{WSI-Level Search}. The process involves matching one WSI to another using the \emph{median of minimum distances} \cite{kalra2020Yottixel}. For each query WSI, its patch embeddings are compared with the patch embeddings of every WSI in the archive.}
\label{fig:med_min}
\end{figure*}

In the literature \cite{kalra2020Yottixel,kalra2020pan}, the patch size of $1024\!\times\! 1024$ at $20\times$ magnification is used and thus we use the same. Once a WSI entirely tiled, a subset of patches $P = \{p_{1}, p_{2},\dotsc, p_{N}\}$ with tissue threshold $\geq t$ (\ie, $70\%$) are selected (here, $N$ is the total number of patches in subset $P$). Subsequently, these selected patches $P$ are fed into a deep neural network $f(\cdot)$ to extract the corresponding set of embeddings $E = \{\textbf{e}_{1}, \textbf{e}_{2}, \dotsc, \textbf{e}_{N}\}$. Empirically, we use DenseNet-121~\cite{huang2017densely} pre-trained on natural images of ImageNet~\cite{deng2009imagenet}. 

Here, the selection of DenseNet~\cite{huang2017densely} is a choice to mitigate any potential bias towards specific histological features (\ie, any properly trained network can be used). The primary goal is to identify various structural elements and edges within the WSI in order to effectively distinguish and capture the multitude of intricate tissue details.


All embedding vectors in $E$ are then used to get one centroid $\textbf{c}$ embedding vector computed as the mean of embedding vectors $\textbf{e}_i$, where $\textbf{e}_i \in E$, and $i = \{1, 2, \dotsc, N\}$. $\textbf{c}$ is computed as

\begin{equation}
\label{eq:centroid}
\textbf{c} = \frac{1}{|N|} \sum_{i=0}^{N} \textbf{e}_{i}. 
\end{equation}

Once, we calculated the centroid of the WSI, the set of Euclidean distances $D  = \{ d_{1}, d_{2}, \dotsc, d_{N}\}$ from the centroid is computed  for each patch in $P$. Euclidean distance is measured to quantify the degree of dissimilarity between patches. Individual distances $d_i$ are computed as

\begin{equation}
    d_i = \|\textbf{e}_{i} - \textbf{c}\|_2,
\end{equation}

\noindent where $d_i \in D$, and here $i = \{1,2,\dotsc, N\}$.

To compute the centroid $\textbf{c}$, we used the k-means algorithm with \underbar{only one centroid}. Subsequently, these distances $D$ are discretized by rounding them to the nearest integer $r(d_i)$.



Discretized patches that exhibit similar Euclidean distances are grouped together in the set Euclidean bins $B = \{b_1, b_2, \ldots, b_K\}$ since their proximity in terms of Euclidean distance suggests similarity (here, $K$ is the number of Euclidean bins which in turn represents the final number of selected patches). In this process, it is not required to manually specify the number of bins as this is the case for Yottixel's mosaic when it defines the number of clusters. By contrast, $K$ is dynamically determined based on the variability in the Euclidean distances among the patches. This adaptability allows the proposed method to effectively capture diverse numbers of distinct tissue regions within the WSI. Finally, a single patch is randomly chosen from each Euclidean bin, considering that all patches within the same bin are regarded as similar. Figure~\ref{fig:Euc_bins} shows the discrete Euclidean bins and selected patches from each bin. These selected set of patches $P_s = \{p_{s1}, p_{s2}, \dotsc, p_{sK} \}$ constitute distinct patches called WSI's \emph{montage}. 





\subsection{Atlas for WSI Matching}
After identifying a unique set of patches from the WSI at a lower magnification level (say $2.5\times$), these patches are subsequently extracted at higher magnification (say $20\times$) with a patch size of $1024\times1024$ pixels. This process generates a montage that contains fewer patches than contained in WSI. This approach enhances computational efficiency and minimizes storage space requirements for subsequent processing without compromising the distinct information in the WSI. The patches in a montage are converted to a set of barcodes using the MinMax algorithm \cite{tizhoosh2016minmax,kalra2020Yottixel}. To achieve this, the patches are initially converted into feature vectors using KimiaNet~\cite{riasatian2021}, which is a DenseNet-121~\cite{huang2017densely} model trained on histological data from TCGA. Global average pooling is applied to the feature maps obtained from this last convolutional layer, resulting in a feature vector with a dimension of 1024. Following feature extraction, we employ the discrete differentiation of the MinMax algorithm~\cite{tizhoosh2016minmax, kalra2020Yottixel}, to convert the feature vectors into binary representations known as a ``barcode''. This barcode  is lightweight and enables rapid Hamming distance-based searches~\cite{kalra2020Yottixel}. While it's possible to directly assess image similarity using deep features and metrics like the Euclidean distance, there is a notable concern regarding computational and storage efficiency, particularly when conducting searches within a large databases spanning various primary sites. Following the processing and binarization of all WSIs using the SDM method, the resulting barcodes are preserved as a reference ``atlas'' (structured database of patients with known outcomes). This atlas can subsequently be employed for the matching process when handling new patients, enabling efficient search and matching applications.

Matching WSI to one another poses significant challenges due to various factors. One key challenge arises from the inherent variability in the number of patches extracted from different WSIs. Since WSIs can vary in size and complexity, the number of patches derived from them can differ substantially. Additionally, factors such as variations in tissue preparation, staining quality, and imaging conditions can introduce further complexity. All these factors make it challenging to establish WSI-to-WSI matching, requiring sophisticated computational methods to address these variations and ensure robust matching in histopathological analysis. 

To overcome this challenge, Kalra et al.~\cite{kalra2020Yottixel} introduced a novel approach called the ``\emph{median of minimum}'' distances within the search engine Yottixel. This technique aims to enhance the robustness of WSI-to-WSI matching. It does so by considering the minimum distances between patches in two WSIs and then selecting the median of these minimum distances as a representative measure of WSI similarity (see Figure~\ref{fig:med_min}). 
In this study, we adopted the median-of-minimum method to perform WSI-to-WSI matching within the atlas.

\section{Experiments \& Results}
\label{sec:results}

\begin{table*}[ht]
\centering
\resizebox{1\linewidth}{!}{\begin{tabular}{l|cccccc|cccccc|cccccc|cc|cc}
\hline
\multicolumn{1}{c|}{\multirow{3}{*}{Datasets}} & \multicolumn{6}{c|}{Accuracy}                                                    & \multicolumn{6}{c|}{Macro Average}                                               & \multicolumn{6}{c|}{Weighted Average}                                            & \multicolumn{2}{c|}{\multirow{2}{*}{\begin{tabular}[c]{@{}c@{}}Patches per WSI\\  (median$\pm$std.)\end{tabular}}} & \multicolumn{2}{c}{\multirow{2}{*}{\begin{tabular}[c]{@{}c@{}}Number of \\Missed WSIs \end{tabular}}} \\ \cline{2-19}
\multicolumn{1}{c|}{}                          & \multicolumn{2}{c}{Top-1} & \multicolumn{2}{c}{MV@3} & \multicolumn{2}{c|}{MV@5} & \multicolumn{2}{c}{Top-1} & \multicolumn{2}{c}{MV@3} & \multicolumn{2}{c|}{MV@5} & \multicolumn{2}{c}{Top-1} & \multicolumn{2}{c}{MV@3} & \multicolumn{2}{c|}{MV@5} & \multicolumn{2}{c|}{} & \multicolumn{2}{c}{}  \\ \cline{2-23} 
\multicolumn{1}{c|}{} & Yottixel       & SDM      & Yottixel     & SDM      & Yottixel       & SDM      & Yottixel      & SDM      & Yottixel      & SDM      & Yottixel       & SDM      & Yottixel       & SDM      & Yottixel      & SDM      & Yottixel       & SDM      & Yottixel  & SDM & Yottixel  & SDM\\ \hline
TCGA~\cite{riasatian2020}   & 81 & 81 & 82 & 82 & 81 & \textbf{82} & 75 & 75 & 75 & 75 & 72  & \textbf{74}   & 80 & \textbf{81} & 82 & 82 & 80 & \textbf{82}  & 33$\pm$21 & \textbf{24$\pm$4} & 4 & \textbf{0} \\
BRACS~\cite{brancati2022bracs}  & \textbf{62} & 61 & 65  & \textbf{66} & 65 & \textbf{66} & 54 & \textbf{55} & 57 & \textbf{59} & 57 & \textbf{58} & \textbf{62} & 61 & 64 & \textbf{66} & 64 & \textbf{65} & 21$\pm$16 & \textbf{30$\pm$5} & 20 & \textbf{0} \\
PANDA~\cite{bulten2022artificial}  & 58 & \textbf{59} & 57 & \textbf{58} & 57 & 57 & 59 & 59 & 57 & \textbf{58} & 56 & 56 & 58 & \textbf{59} & 58 & 58 & 57 & 57 & \textbf{9$\pm$2} & 12$\pm$3 & 120 & \textbf{8}  \\
CRC    & 60 & \textbf{66} & 60 & \textbf{70} & 60 & \textbf{68} & 60 & \textbf{66} & 61 & \textbf{70} & 60 & \textbf{69} & 58 & \textbf{66} & 59 & \textbf{70} & 59 & \textbf{68} & 17$\pm$15 & \textbf{21$\pm$4} & 0 & 0 \\
Liver  & 76  & \textbf{77} & 79 & 79 & 80 & 80 & 62 & 62 & 67 & \textbf{68} & 65 & \textbf{66} & 75 & 75 & \textbf{79} & 78 & 79 & 79 & \textbf{9$\pm$3} & 17$\pm$4 & 2 & \textbf{0} \\
BC     & 55 & \textbf{64}  & -  & - & -  & - & 51  & \textbf{55} & -  & - & - & - & 52 & \textbf{59} & - & - & - & - & \textbf{11$\pm$9}  & 27$\pm$5 & 1 & \textbf{0}  \\ \hline
\end{tabular}}
\caption{The collective accuracy, both macro and weighted averages at top-1, MV@3, and MV@5 retrievals using both Yottixel mosaic~\cite{kalra2020Yottixel}  and SDM montage methods across all datasets employed for evaluation. The number of patches per WSI for each dataset is documented, inclusive of the associated standard deviation. Additionally, the number of missed WSIs for each dataset is also presented when using Yottixel's  mosaic~\cite{kalra2020Yottixel}  and SDM's montage.}
\label{tab:all_results}
\end{table*}

The verification and validation of histological similarity pose formidable challenges. A comprehensive validation scenario would ideally entail the comparison of numerous patients across diverse healthcare institutions, involving multiple pathologists conducting visual inspections over an extended timeframe. In this research, as in many other works, the performance of the search task was quantified by approaching it as a classification problem. One of the primary advantages of employing classification methodologies lies in their ease of validation; each image can be categorized as either belonging to a specific class or not, a binary concept that allows for performance quantification through tallying misclassified instances. Nonetheless, it's essential to acknowledge that the notion of similarity in image search is a fundamentally continuous subject matter (in many cases, a straightforward yes/no answer may be a coarse oversimplification) and predominantly a matter of degree (ranging from \emph{almost identical} to \emph{utterly dissimilar}). Moreover, distance measures, such as Euclidean distance, which assess dissimilarity between two feature vectors representing images, are typically used to gauge the extent of similarity (or dissimilarity) between images~\cite{radenovic2018fine, mazaheri2023ranking}. 

All experiments have been conducted on Dell PowerEdge XE8545 with 2$\times$ AMD EPYC 7413 CPUs, 1023 GB RAM, and 4$\times$ NVIDIA A100-SXM4-80GB using TensorFlow (TF) version of DenseNet and KimiaNet. We used TF 2.12.0, Python 3.9.16, CUDA 11.8, and CuDNN 8.6 on a Linux operating system. In all experiments, two patch selection methodologies were employed: Yottixel's \textbf{mosaic} and SDM's \textbf{montage}. Subsequently, patches were extracted at 20× magnification, with dimensions measuring 1000 × 1000 for the mosaic and 1024 × 1024 for the montage. 
This particular size facilitates computational efficiency and is aligned with architectural requirements, particularly for ViTs~\cite{dosovitskiy2020image, caron2021emerging}. Following patching, feature extraction was executed using KimiaNet \cite{riasatian2021}. These features were subsequently transformed into barcodes, characterized by their lightweight nature and ability to facilitate swift Hamming distance-based searches~\cite{tizhoosh2016minmax, kalra2020Yottixel}. The barcodes from all the WSIs are subsequently used for the creation of an atlas, an indexed dataset for a specific disease. This atlas functions as a fundamental asset, tested via a ``leave-one-patient-out'' search and matching experiment, a notably rigorous method particularly suited for datasets of small to medium size, with the aim of retrieving the highest-ranking matching WSIs. The computer vision literature typically emphasizes \textbf{top-n} accuracy, where search is deemed successful when any one of the top-n search results is correct. In contrast, our approach relies on more rigorous ``\textbf{majority-n} accuracy'', which we find to be a significantly more dependable validation scheme for medical imaging \cite{kalra2020Yottixel,kalra2020pan}. Under this scheme, a search is deemed correct only when the majority of the top-n search results are correct. The advantage of a search process lies in its capability to retrieve multiple top-matching results, thereby enabling the achievement of consensus among the top retrievals to solidify the decision-making rationale.

Once, the top matching results (through majority voting at top-n -- MV@n) are compiled, then the most commonly used evaluation metrics for verifying the performance of image search and retrieval algorithms are average precision, recall, and F1-scores~\cite{dubey2021decade, kalra2020Yottixel, riasatian2021}.

\subsection{Results}

\begin{figure*}[t]
\centerline{\includegraphics[width =  1\textwidth]{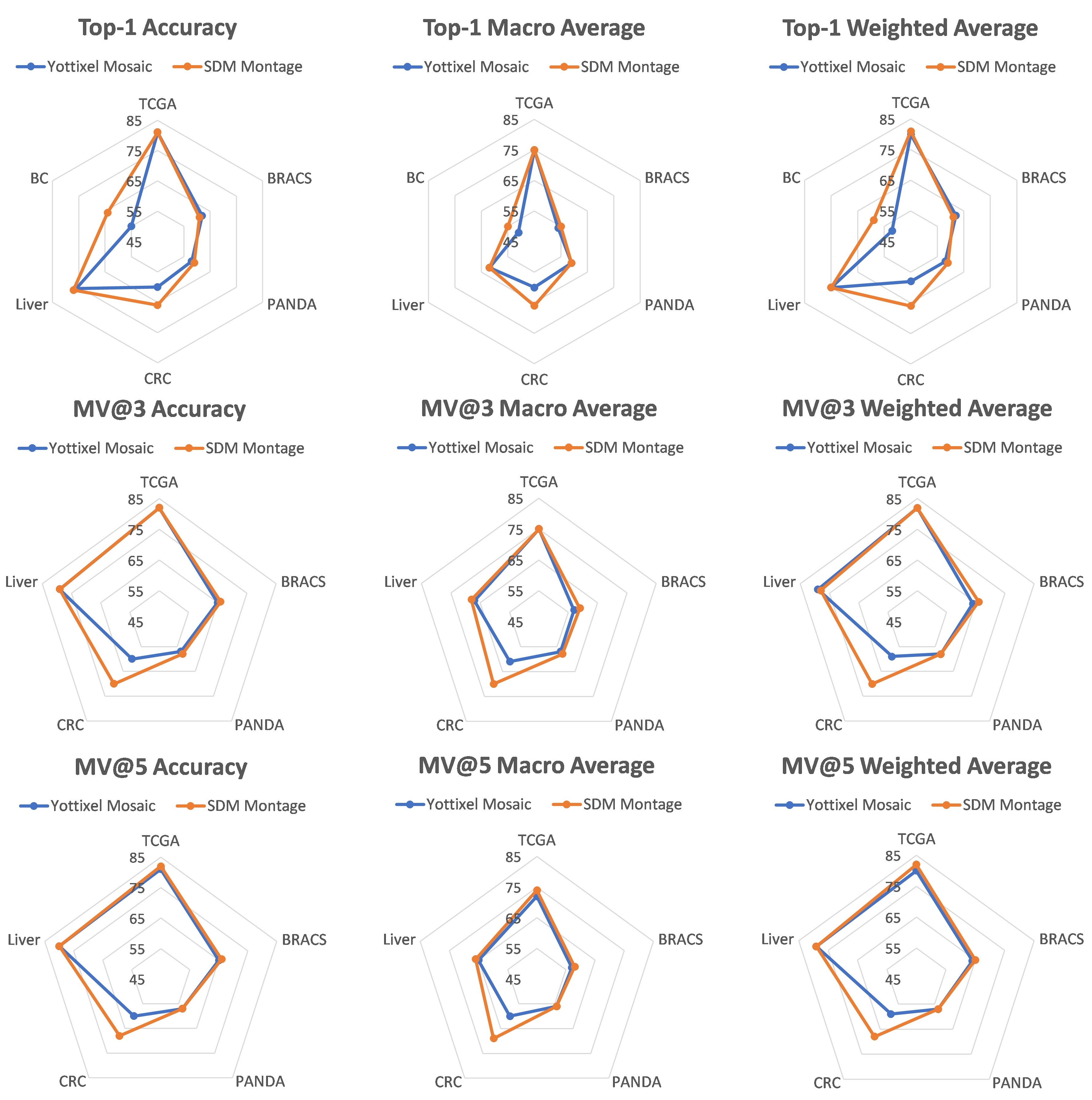}}
\caption{\textbf{Overall Results.} The collective accuracy, both macro and weighted averages, at top-1, MV@3, and MV@5 using both Yottixel's mosaic~\cite{kalra2020Yottixel} and SDM's montage across all datasets.}
\label{fig:SDM_Combined}
\end{figure*}


\begin{figure*}[t]
\centerline{\includegraphics[width =  1\textwidth]{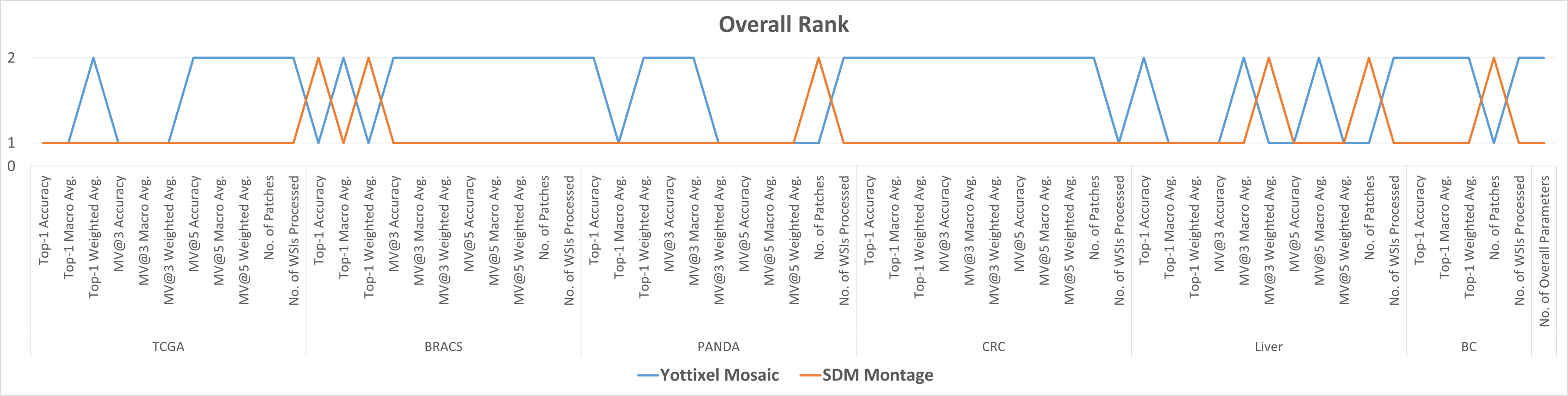}}
\caption{\textbf{Comprehensive Ranking Scheme.} A comprehensive ranking scheme was devised to evaluate the performance of the two methods: Yottixel's mosaic~\cite{kalra2020Yottixel} and SDM's montage. In this scheme, a rank of `1' signifies superior performance of a method relative to the other, a rank of `2' indicates inferior performance, and identical ranks of `1' for both methods denote comparable performance. After aggregating the results across all metrics, Yottixel mosaic achieved an average rank of 1.64, while SDM montage recorded a more favorable score of 1.09. }
\label{fig:SDM_Rank}
\end{figure*}

SDM's montage has been extensively evaluated on various public and private histopathology datasets using a ``leave-one-out'' WSI search and matching as a downstream task on each dataset and compared with the state-of-the-art Yottixel's mosaic. For public dataset evaluation, the following datasets have been used:  The Cancer Genome Atlas (TCGA)~\cite{tomczak2015review}, BReAst Carcinoma Subtyping (BRACS)~\cite{brancati2022bracs}, and Prostate cANcer graDe Assessment (PANDA)~\cite{bulten2022artificial}. On the other hand, for the private dataset evaluation, we have used Alcoholic Steatohepatitis (ASH) and Non-alcoholic Steatohepatitis (NASH) Liver, Colorectal Cancer (CRC), and Breast Cancer (BC) datasets from our hospital. The extended details about each dataset are provided in the supplementary file.

\noindent \textbf{TCGA -- } It is the largest public and comprehensive repository in the field of cancer research. A set of 1466 out of 1553 slides were used. These WSIs  were not involved in the fine-tuning of KimiaNet~\cite{riasatian2021}.

\noindent \textbf{BRACS -- } The BRACS dataset comprises a total of 547 WSIs derived from 189 distinct patients~\cite{brancati2022bracs}. 
The dataset is categorized into two main subsets: WSI and Region of Interest (ROI). Within the WSI subset, there are three primary tumor Groups (Atypical, Benign, and Malignant tumors)~\cite{brancati2022bracs} that we used for validation. 

\noindent \textbf{PANDA -- } It is the largest publicly available dataset of prostate biopsies, put together for a global AI competition~\cite{bulten2022artificial}. 
In our experiment, we used the publicly available training cohort of 10,616 WSIs with their International Society of Urological Pathology (ISUP) scores for an extensive leave-one-out search and matching experiment~\cite{epstein20052005, bulten2022artificial}.

\noindent \textbf{Private CRC -- } The CRC dataset, sourced from our hospital, encompasses a collection of 209 WSIs, with a primary focus on colorectal histopathology. This dataset is categorized into three distinct groups as WSI labels. 

\noindent \textbf{Private Liver -- } 326 Liver biopsy slides were acquired from patients who had been diagnosed with either ASH or NASH at our hospital. Our cohort also includes some normal WSIs, facilitating the differentiation between neoplastic and non-neoplastic tissue specimens. 

\noindent \textbf{Private BC -- } 74 Breast tumor slides were acquired from patients at our hospital. There are 16 different subtypes of breast tumors were employed in this experiment. 

To evaluate the performance of the SDM's montage against Yottixel's mosaic for all datasets, we retrieved the top similar cases using leave-one-out evaluation. The assessments rely on several retrieval criteria, including the top-1 retrieval, the majority vote among the top 3 retrievals (MV@3), and the majority vote among the top 5 retrievals (MV@5). The accuracy, macro average, and weighted average for top-1, MV@3, and MV@5 are reported in Table~\ref{tab:all_results}. For better visualization, Figure~\ref{fig:SDM_Combined} shows an overall comparison of accuracy, macro average, and weighted average at top-1, MV@3, and MV@5 using both Yottixel mosaic~\cite{kalra2020Yottixel} and SDM montage methods across all datasets used in this experiment. In addition to these performance metrics, a comparative analysis of the number of patches extracted per WSI by each respective method,  as well as documenting the count of WSIs that each method was unable to process. These comparative metrics are systematically presented in Table~\ref{tab:all_results} (also see the supplementary file for extended evaluation results).

In our experiments, the overall performance of SDM's montage showcased superior performance when compared with Yottixel's mosaic as we can see in Table~\ref{tab:all_results} and Figure~\ref{fig:SDM_Combined}. For the TCGA, SDM's montage demonstrated superior performance by +2\% in macro average of F1-scores, and +1\% in accuracy and weighted average of F1-score as compared to the Yottixel mosaic when it came to the MV@5 retrievals. SDM exhibited improvements of +1\%, +2\%, and +1\% in the macro average of F1-scores concerning top-1, MV@3, and MV@5 retrievals, respectively, when experimenting with the BRACS dataset. When evaluating PANDA images, SDM exhibited comparable performance to the Yottixel concerning accuracy at MV@5. However, a noteworthy distinction emerged when considering accuracy at top-1 and MV@3 retrievals. Regarding the macro-averaged F1-scores for MV@3,  SDM  demonstrates an improvement of 1\%. For CRC evaluation, SDM displayed superior performance in the macro-average of F1-scores by +6\%, +9\%, and +9\% for top-1, MV@3, and MV@5, respectively. From an accuracy perspective, the SDM demonstrated improvements of +6\%, +10\%, and +8\% for the top-1, MV@3, and MV@5 retrievals, respectively. For the Liver, SDM and Yottixel have demonstrated a comparable performance. However, there was an enhancement of +1\% in the macro-average of F1-scores when using the SDM. Finally, for the BC dataset, SDM evidently showcased superior performance in the top-1 retrieval result by +9\% in accuracy, +4\% in macro average of F1-scores, and +7\% in the weighted average. In the assessment of BC, the evaluation is confined to the top 1 retrieval for each query, a decision driven by the limited number of WSIs available per tumor type. Moreover, It has been observed that Yottixel demonstrates a propensity to omit some WSIs across the majority of datasets. In contrast, the SDM is capable of effectively processing the preponderance of WSIs from all evaluated datasets (see Table~\ref{tab:all_results} for details). Additionally, our findings indicate that for excisional biopsy samples, the SDM can represent the entire WSI using a fewer number of patches in comparison to 5\% of the patches selected by Yottixel. Conversely, in the context of needle biopsies, Yottixel demonstrates an efficiency in selecting a lesser number of patches than SDM. 

\section{Discussions}
\label{sec:discussion}

Unsupervised WSI-to-WSI search holds a significant importance, particularly when searching through large archives of medical images. It offers the invaluable capability of generating a computational second opinion based on previously established and evidently diagnosed cases. By leveraging unsupervised search techniques, medical practitioners can efficiently compare a new WSI to a repository of historical cases without requiring pre-labeled data. 
To execute WSI-to-WSI search effectively, it is imperative to employ a sophisticated divide-and-conquer strategy. WSIs are typically gigapixel files and intricate images that are impractical to be processed in their entirety. Therefore, the divide-and-conquer approach involves breaking down the WSI into smaller, more manageable patches to compare WSIs. Relying on a small number of  patches is a crucial aspect of practical WSI-to-WSI matching. Incorporating a diverse range of patches from the entire WSIs is critical for capturing the rich tissue information contained within image. 
By capturing the inherent diversity within WSIs, utilizing a varied set of patches can boost diagnostic accuracy. This approach not only refines the quality of research insights but also strengthens the ability to generalize findings across a wider array of cases.

For the specific objective at hand, we have introduced a methodology referred to as ``Selection of Distinct Morphologies (SDM)'' (presented in Section~\ref{sec:method}). The primary aim of SDM is to systematically choose a small set of patches from a larger pool, with the intention of encompassing all unique morphological characteristics present within a given WSI. These meticulously selected patches collectively constitute what we term a ``montage''. The proposed methodology has undergone rigorous testing across six distinct datasets, comprising three publicly available datasets and three privately acquired datasets. In the evaluation process, we conducted a comprehensive comparative analysis with the Yottixel's mosaic~\cite{kalra2020Yottixel}, which is the sole existing patch selection method reported in the literature. This extensive testing thoroughly assesses the effectiveness and performance of our approach in relation to the established benchmark provided by Yottixel's  mosaic~\cite{kalra2020Yottixel}. In Figure~\ref{fig:SDM_Rank}, a  ranking methodology is presented to assess the efficacy of two distinct methods, Yottixel's mosaic and SDM's  montage, across multiple datasets, employing a range of evaluation metrics. The criteria employed to evaluate and rank the algorithms encompass various metrics, including accuracy values, macro averages, weighted averages, the number of WSIs successfully processed per dataset, the number of patches extracted for each dataset, and the cumulative number of parameters essential for the algorithm's operation. Within this ranking paradigm, a designation of `1' denotes that a method exhibits a performance edge over its counterpart, while a `2' suggests subpar performance. Receiving identical rankings of `1' for both methods suggests they exhibit parity in their performance outcomes. Upon consolidating the rankings overall metrics, Yottixel's mosaic registered an average ranking of 1.64, in contrast to the SDM's montage which secured a more commendable average of 1.09. An inspection of the figure clearly illustrates the SDM's montage consistently achieving a `1' rank more often than the Yottixel mosaic. 

\section{Conclusions}
Our investigations underscored the paramount significance of an adept patch selection strategy in the context of WSI representation. The robustness and precision of classification and search hinge on the ability to meticulously curate informative patches from the gigapixel WSIs. In this regard, our proposed approach, SDM, has demonstrated remarkable efficacy through extensive experimentation on diverse datasets, including both publicly available and privately acquired datasets. Throughout our evaluations, it has been consistently discerned that the proposed methodology outperforms the prevailing state-of-the-art patch selection technique, as epitomized by Yottixel's mosaic. The Yottixel approach necessitates the specification of certain empirical parameters, such as the percentage of patch selection and the number of color clusters, which introduces some unwanted variability. In contrast, the SDM approach obviates the need for such empirical parameter settings, inherently optimizing the selection to capture the distinct morphological features present in the WSI. Taken together, our findings affirm that a robust patch selection strategy is indispensable for enhancing the effectiveness of WSI classification and matching applications, with our proposed method showcasing substantial advancements in this critical domain.

\noindent \textbf{Limitations:} In histological pattern recognition, the Field of View (FoV) plays a crucial role~\cite{neary2023minimum}, as different patterns necessitate varying FoV widths for accurate identification. In this study, we analyze a 1024$\times$1024 FoV at a 20$\times$ magnification. Using different FoVs at different magnifications for different tumour types may expand the insights into different aspects of computational pathology.

\noindent \textbf{Broader Impacts:} The proposed method for WSI matching has the potential to be used as a virtual second opinion by search \& matching with evidently diagnosed previous cases. With the widespread use of computational pathology in clinical practice, these methods can reduce the workload and human errors of pathologists. Furthermore, reducing intra- and inter-observer variability. As well, it is conceivable that the proposed \emph{divide \& conquer} scheme may be applicable to other fields that also employ gigapixel images, e.g., satellite imaging and remote sensing.

\noindent\textbf{Acknowledgments:}  The authors thank 
Ghazal Alabtah, Saba Yasir, Tiffany Mainella, Lisa Boardman, Chady Meroueh, Vijay H. Shah, and Joaquin J. Garcia for their valuable insights, discussions, and suggestions.

{
    \small
    \bibliographystyle{unsrt}
    \bibliography{main}
}

\clearpage
\setcounter{page}{1}
\maketitlesupplementary

\noindent Content of Supplementary File:\\
\\
\\
\noindent Extended Results: Section~\ref{sec:Exd_Results}
\\
\\
\begin{itemize}
    \item Public -- TCGA~\ref{sec:Exd_Results_TCGA}
    \item Public -- BRACS~\ref{sec:Exd_Results_BRACS}
    \item Public -- PANDA~\ref{sec:Exd_Results_PANDA}
    \item Private -- CRC~\ref{sec:Exd_Results_CRC}
    \item Private -- ASH \& NASH~\ref{sec:Exd_Results_Liver}
    \item Private -- BC~\ref{sec:Exd_Results_BC}
\end{itemize}

\onecolumn

\section{Extended Results}
\label{sec:Exd_Results}
Expanded results stemming from a comparative analysis of the Selection of Distinct Morphologies (SDM) and Yottixel, utilizing six diverse datasets, are presented in this supplementary file. SDM montage has been extensively evaluated on various public and private histopathology datasets using a ``leave-one-out'' WSI search and matching as a downstream task on each dataset and compared with the state-of-the-art Yottixel's mosaic. For public dataset evaluation, the following datasets have been used:  The Cancer Genome Atlas (TCGA)~\cite{tomczak2015review}, BReAst Carcinoma Subtyping (BRACS)~\cite{brancati2022bracs}, and Prostate cANcer graDe Assessment (PANDA)~\cite{bulten2022artificial}. On the other hand, for the private dataset evaluation, we have used Alcoholic Steatohepatitis (ASH) and Non-alcoholic Steatohepatitis (NASH) Liver, Colorectal Cancer (CRC), and Breast Cancer (BC) datasets from our hospital. 


\subsection{Public -- The Cancer Genome Atlas (TCGA)}
\label{sec:Exd_Results_TCGA}
The Cancer Genome Atlas (TCGA) is a public and comprehensive repository in the field of cancer research. Established by the National Institutes of Health (NIH) and the National Cancer Institute (NCI), TCGA represents a collaborative effort involving numerous research institutions. Its primary mission is to analyze and catalog genomic and clinical data from a wide spectrum of cancer types. It is the largest publicly available dataset for cancer research. The dataset contains 25 anatomic sites with 32 cancer subtypes of almost 33,000 patients.

The KimiaNet~\cite{riasatian2020} underwent a training process utilizing the TCGA dataset, using the ImageNet weights from DenseNet as initial values. This process involved the utilization of 7,375 diagnostic H\&E slides to extract a substantial dataset of over 240,000 patches, each with dimensions measuring $1000 \times 1000$, for training KimiaNet. Additionally, a set of 1553 slides was set aside for evaluation purposes, comprising a test dataset consisting of 777 slides and a validation dataset encompassing 776 slides.

From 1553 evaluation slides that were not involved in the fine-tuning of KimiaNet~\cite{riasatian2021}, 1466 were used in the evaluation of this study (see Table.~\ref{tab:TCGA} for a detailed breakdown of the dataset). 

\begin{figure*}[!]
\centerline{\includegraphics[width =  1\textwidth]{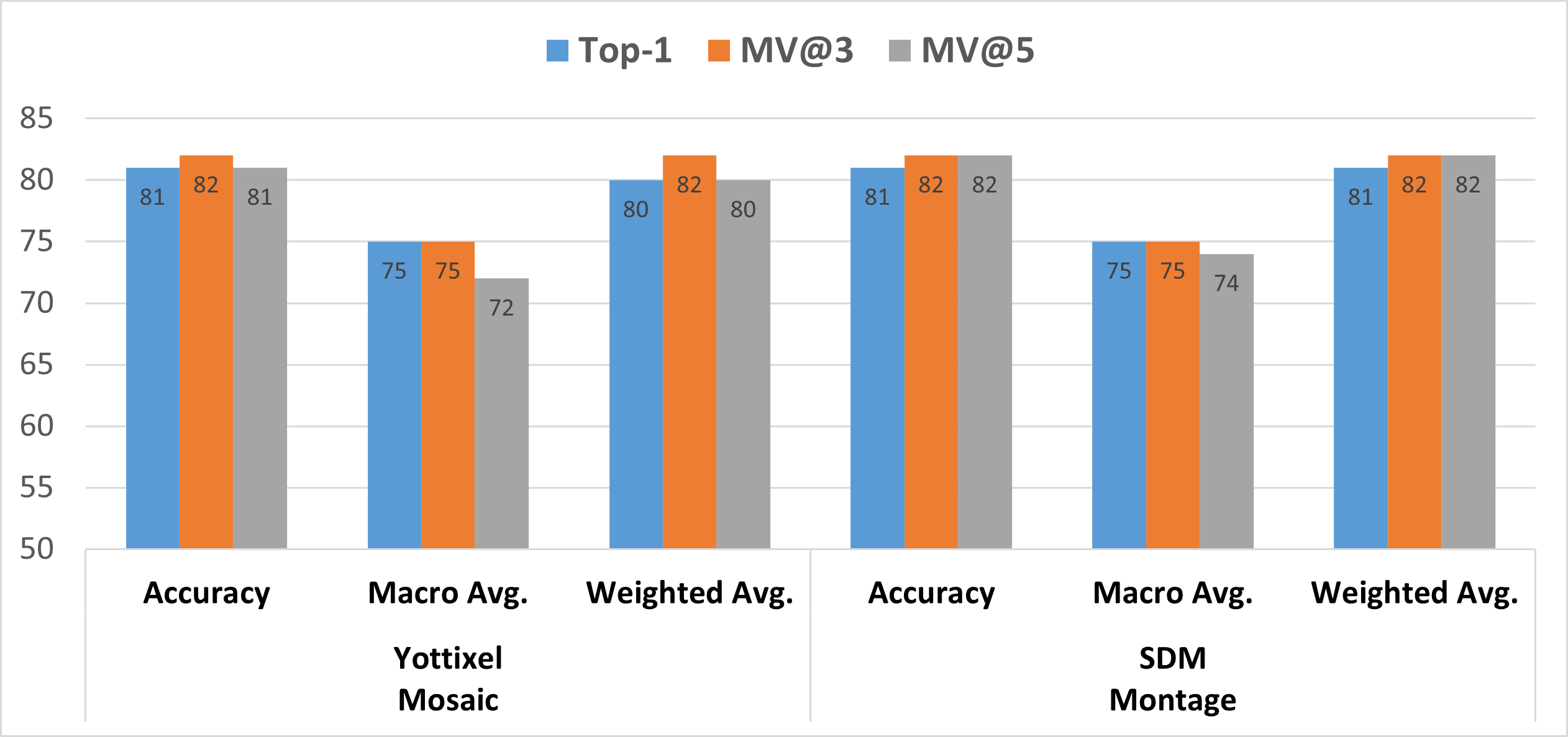}}
\caption{Accuracy, macro average of f1-scores, and weighted average of f1-scores are shown from Yottixel mosaic, and SDM montage. The evaluations are based on the top 1 retrieval, the majority among the top 3 retrievals, and the majority among the top 5 retrievals using the TCGA dataset.}
\label{fig:TCGA_Accuracy}
\end{figure*}

\begin{figure*}[!]
\centerline{\includegraphics[width =  0.6\textwidth]{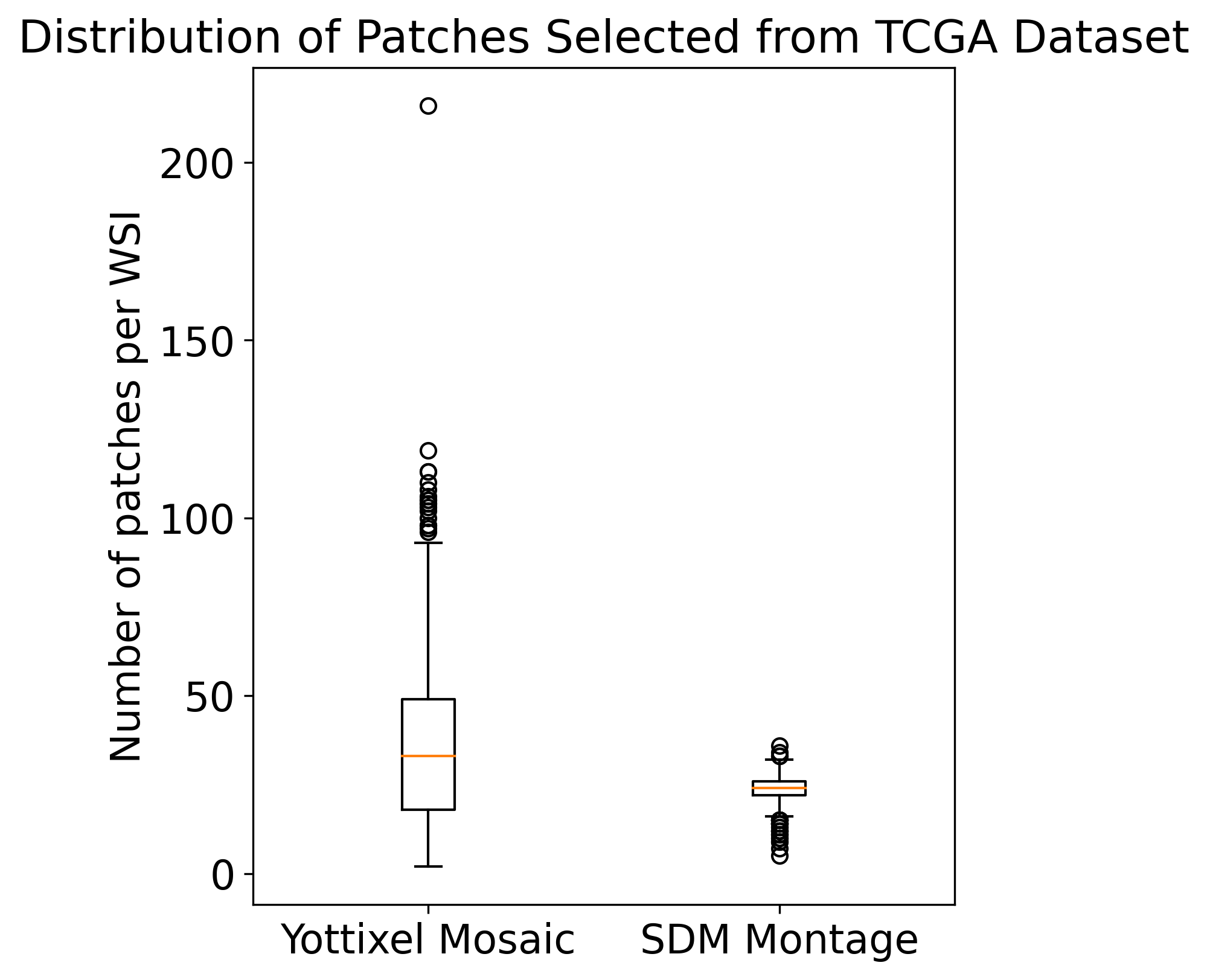}}
\caption{The boxplot illustrates the distribution of patches selected for each WSI in the TCGA dataset from both the Yottixel Mosaic and SDM Montage. Additionally, it provides statistical measures for these distributions. Specifically, for the Yottixel Mosaic, the median number of selected patches is $33\pm21$. Conversely, for the SDM Montage, the median number of selected patches is $24\pm4$.}
\label{fig:TCGA_Patch_boxplot}
\end{figure*}

\begin{figure*}[!]
\centerline{\includegraphics[width =  1\textwidth]{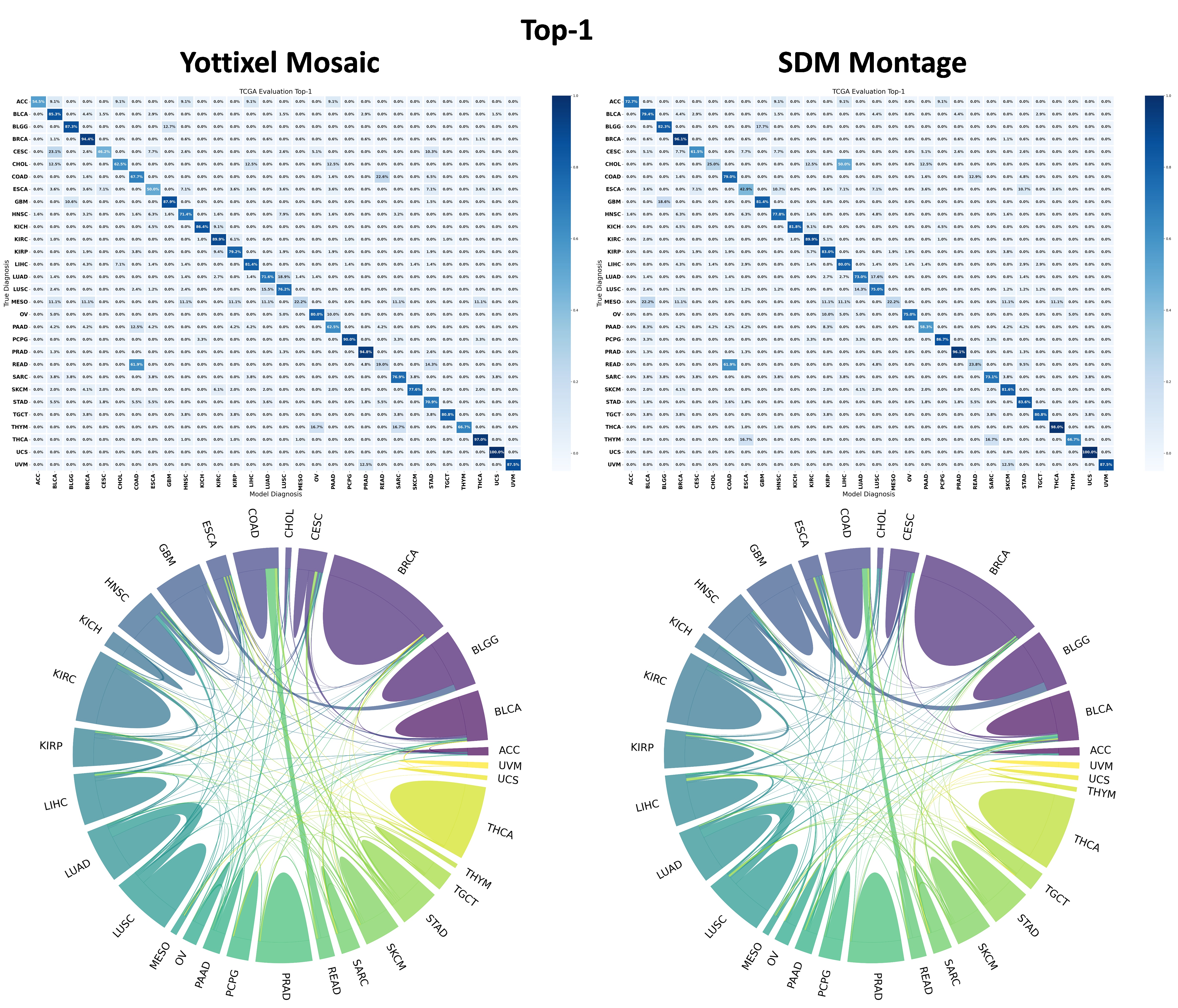}}
\caption{Confusion matrices and chord diagrams from Yottixel mosaic (left column), and SDM montage (right column). The evaluations are based on the top 1 retrieval when evaluating the TCGA dataset.}
\label{fig:TCGA_Chord1}
\end{figure*}

\begin{figure*}[!]
\centerline{\includegraphics[width =  1\textwidth]{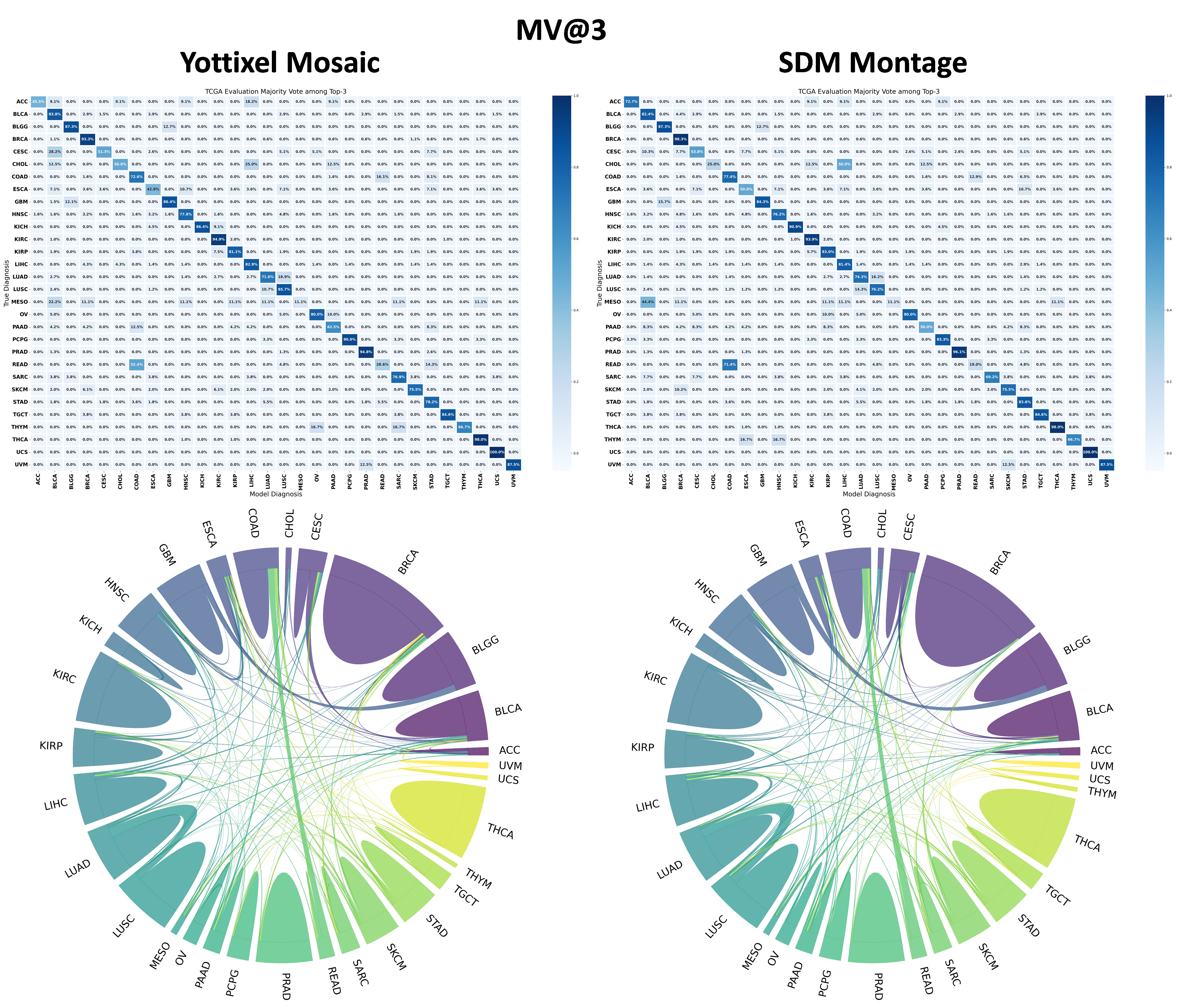}}
\caption{Confusion matrices and chord diagrams from Yottixel mosaic (left column), and SDM montage (right column). The evaluations are based on the majority of the top 3 retrievals when evaluating the TCGA dataset.}
\label{fig:TCGA_Chord3}
\end{figure*}

\begin{figure*}[!]
\centerline{\includegraphics[width =  1\textwidth]{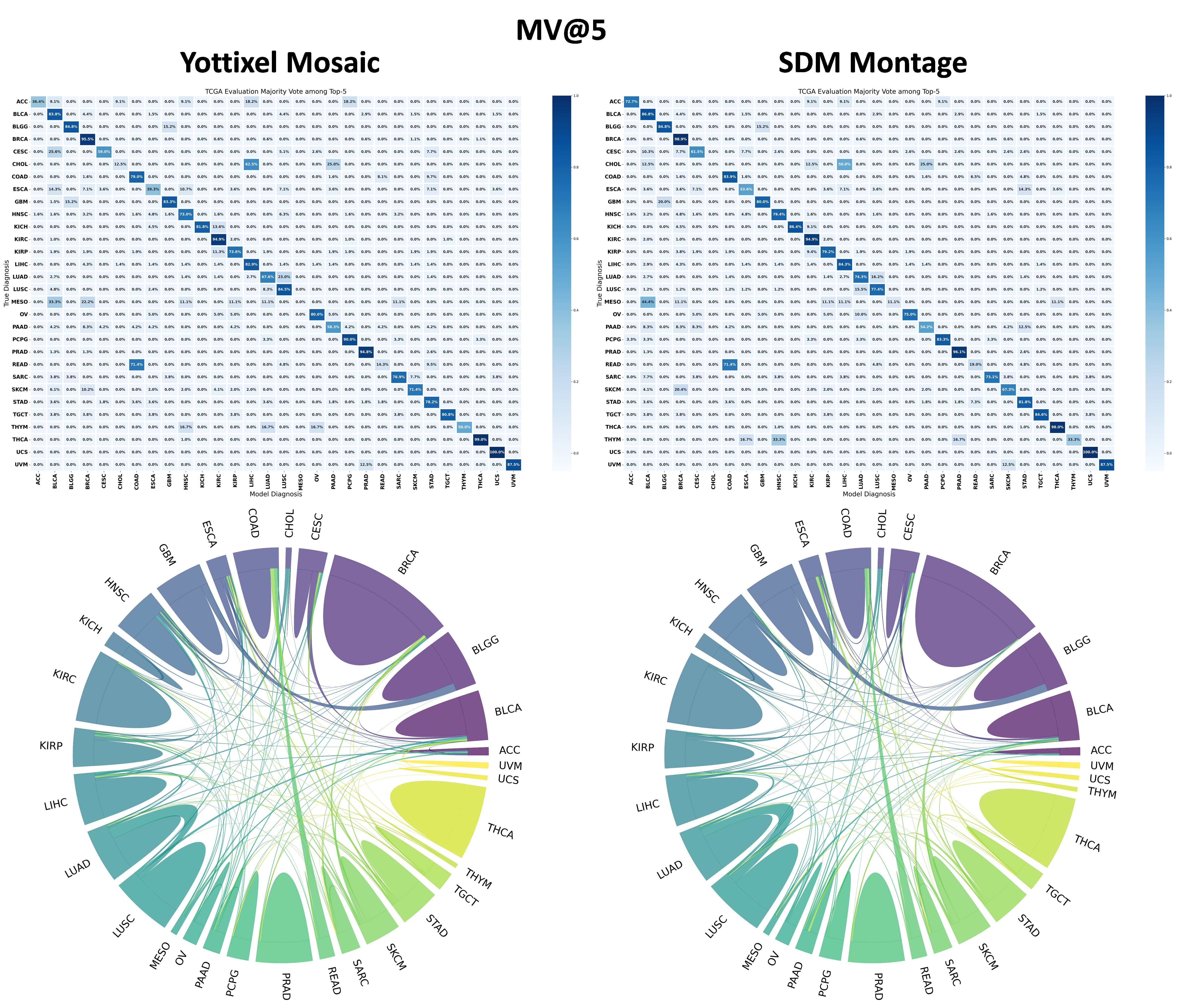}}
\caption{Confusion matrices and chord diagrams from Yottixel mosaic (left column), and SDM montage (right column). The evaluations are based on the majority of the top 5 retrievals when evaluating the TCGA dataset.}
\label{fig:TCGA_Chord5}
\end{figure*}

\begin{figure*}[!]
\centerline{\includegraphics[width =  1\textwidth]{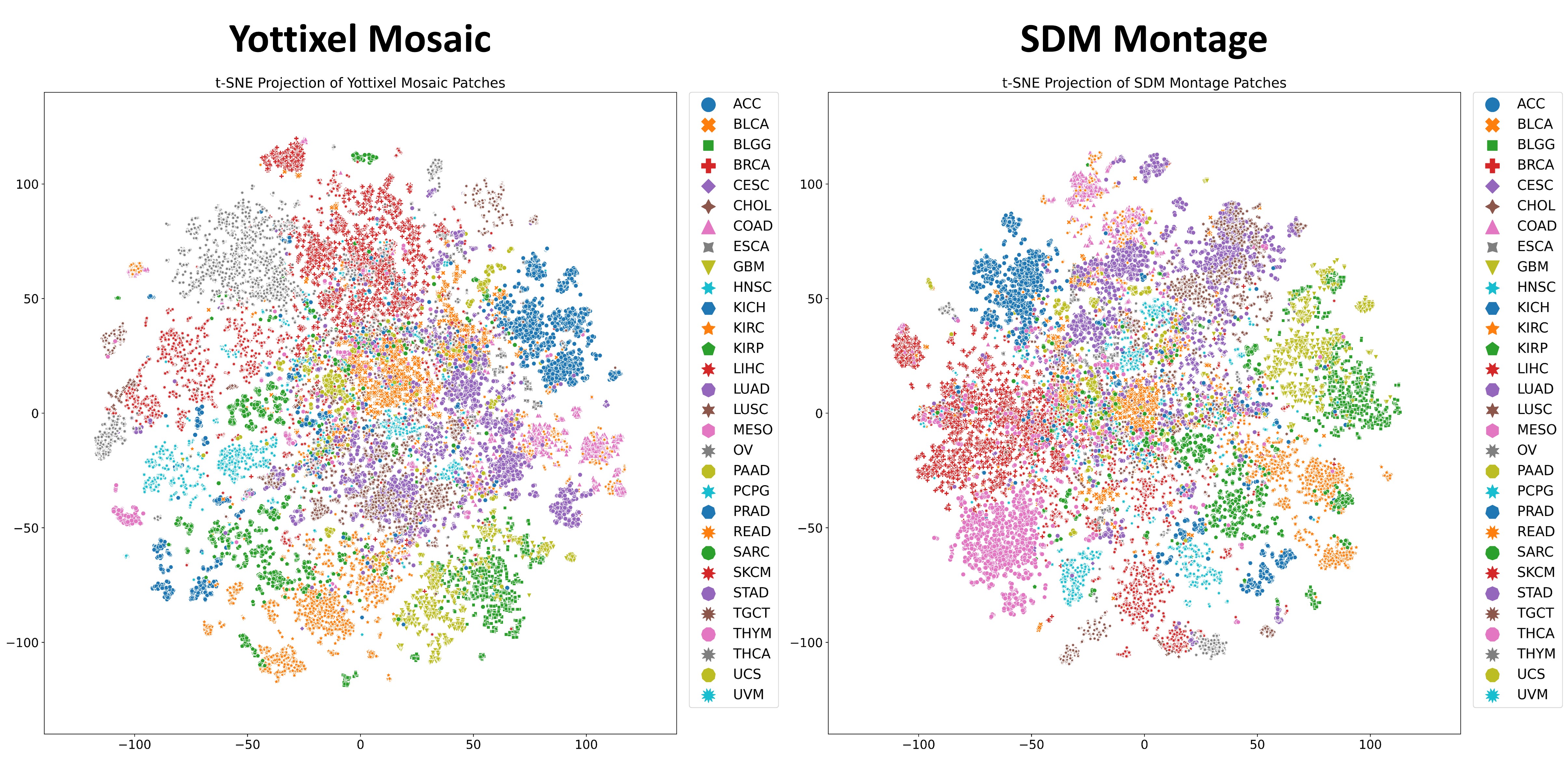}}
\caption{The t-SNE projection displays the embeddings of all patches extracted from the TCGA dataset using Yottixel's mosaic (left) and SDM's montage (right).}
\label{fig:TCGA_TSNE}
\end{figure*}

To assess how the performance of the SDM montage compares to Yottixel's mosaic, a leave-one-out evaluation was conducted to retrieve the most similar cases. The evaluation involved multiple retrieval criteria, including the top-1, MV@3, and MV@5. The accuracy, macro average, and weighted average at top-1, MV@3, and MV@5 are reported in Figure~\ref{fig:TCGA_Accuracy}. Moreover, confusion matrices and chord diagrams at top-1, MV@3. and MV@5 retrievals are illustrated in Figure~\ref{fig:TCGA_Chord1}, \ref{fig:TCGA_Chord3}, and ~\ref{fig:TCGA_Chord5}, respectively. Table~\ref{tab:TCGA_results_YT} and \ref{tab:TCGA_results_SDM} show the detailed results including precision, recall, and f1-score for Yottixel's mosaic and SDM's montage, respectively. In addition to conventional accuracy metrics, we also conducted a comparative analysis of the number of patches extracted per WSI by each method (see the boxplots in Figure~\ref{fig:TCGA_Patch_boxplot} for the depiction of the patch distribution per WSI). To visually represent the extracted patches, t-distributed Stochastic Neighbor Embedding (t-SNE) projections of these patches are also provided in Figure~\ref{fig:TCGA_TSNE}.

\begin{table}[!]
\centering
\resizebox{\columnwidth}{!}{\begin{tabular}{llr}
\hline
\begin{tabular}[c]{@{}l@{}}Primary Diagnoses\end{tabular}                                            & Acronym & Slides \\ \hline
Adrenocortical Carcinoma                                                                             & ACC     & 11     \\
Bladder Urothelial Carcinoma                                                                         & BLCA    & 68     \\
Brain Lower Grade Glioma                                                                             & BLGG    & 79     \\
Breast Invasive Carcinoma                                                                            & BRCA    & 178     \\
\begin{tabular}[c]{@{}l@{}}Cervical Squamous Cell Carcinoma and Endocervical Adenocarcinoma\end{tabular} & CESC    & 39     \\
Cholangiocarcinoma                                                                                   & CHOL    & 8     \\
Colon Adenocarcinoma                                                                                 & COAD    & 62    \\
Esophageal Carcinoma                                                                                 & ESCA    &  28      \\
Glioblastoma Multiforme                                                                              & GBM     &   70     \\
\begin{tabular}[c]{@{}l@{}}Head and Neck Squamous Cell Carcinoma\end{tabular}                        & HNSC    &    63    \\
Kidney Chromophobe                                                                                   & KICH    &   22     \\
\begin{tabular}[c]{@{}l@{}}Kidney Renal Clear Cell Carcinoma\end{tabular}                            & KIRC    &    99    \\
\begin{tabular}[c]{@{}l@{}}Kidney Renal Papillary Cell Carcinoma\end{tabular}                        & KIRP    &   53     \\
Liver Hepatocellular Carcinoma                                                                       & LIHC    &  70      \\
Lung Adenocarcinoma                                                                                  & LUAD    &   74     \\
Lung Squamous Cell Carcinoma                                                                         & LUSC    &     84   \\
Mesothelioma                                                                                         & MESO    &    9    \\
Ovarian Serous Cystadenocarcinoma                                                                    & OV      &    20    \\
Pancreatic Adenocarcinoma                                                                            & PAAD    &    24    \\
\begin{tabular}[c]{@{}l@{}}Pheochromocytoma and Paraganglioma\end{tabular}                           & PCPG    &    30    \\
Prostate Adenocarcinoma                                                                              & PRAD    &   77     \\
Rectum Adenocarcinoma                                                                                & READ    &   21     \\
Sarcoma                                                                                              & SARC    &    26    \\
Skin Cutaneous Melanoma                                                                              & SKCM    &   49     \\
Stomach Adenocarcinoma                                                                               & STAD    &   55     \\
Testicular Germ Cell Tumors                                                                          & TGCT    &   26     \\
Thymoma                                                                                              & THYM    &   6     \\
Thyroid Carcinoma                                                                                    & THCA    &    101    \\
Uterine Carcinosarcoma                                                                               & UCS     &   6     \\
Uveal Melanoma                                                                                       & UVM     &   8     \\ \hline
\end{tabular}}
\caption{Comprehensive details regarding the TCGA dataset utilized in this study, encompassing the corresponding acronyms and the number of slides attributed to each primary diagnosis.}\label{tab:TCGA}
\end{table}

Through this experiment, we observed that SDM exhibited comparable performance to the Yottixel mosaic concerning top-1 retrieval and the majority agreement among the top 3 retrievals. However, notably, the SDM montage demonstrated superior performance by +2\% in the macro avg. of f1-scores, and +1\% in accuracy and weighted avg. of f1-scores as compared to the Yottixel mosaic when it came to the majority agreement among the top 5 retrievals, highlighting its effectiveness in capturing relevant information in this specific retrieval context (see Figure.~\ref{fig:TCGA_Accuracy}). Another notable advantage of employing the SDM montage method becomes evident when examining Figure~\ref{fig:TCGA_Patch_boxplot}, which illustrates the number of patches selected. In comparison to the Yottixel mosaic, SDM proves to be more efficient by selecting a fewer number of patches. This not only conserves storage space but also eliminates the redundancy \& need for empirical determination of the ideal number of patches to select. Additionally, it has come to our attention that Yottixel is more prone to overlooking WSIs in comparison to SDM. Specifically, our observations reveal that Yottixel processed 1462 WSIs, whereas SDM successfully processed the entirety of 1466 WSIs.

\begin{table}[!]
\centering
\resizebox{\columnwidth}{!}{\begin{tabular}{llllllllllr}

\multicolumn{11}{c}{\textbf{Yottixel Mosaic~\cite{kalra2020Yottixel}}}                                                                                                                                                                                                                                                                                                                                                    \\ \hline
\multicolumn{1}{l|}{}                                                            & \multicolumn{3}{c|}{Top-1}                                                                 & \multicolumn{3}{c|}{MV@3}                                                                  & \multicolumn{3}{c|}{MV@5}                                                                  &               \\ \hline
\multicolumn{1}{l|}{\begin{tabular}[c]{@{}l@{}}Primary\\ Diagnoses\end{tabular}} & \multicolumn{1}{c}{Precision} & \multicolumn{1}{c}{Recall} & \multicolumn{1}{c|}{f1-score} & \multicolumn{1}{c}{Precision} & \multicolumn{1}{c}{Recall} & \multicolumn{1}{c|}{f1-score} & \multicolumn{1}{c}{Precision} & \multicolumn{1}{c}{Recall} & \multicolumn{1}{c|}{f1-score} & Slides        \\ \hline
\multicolumn{1}{l|}{ACC} & 0.86 & 0.55 & \multicolumn{1}{l|}{0.67}     & 0.83                          & 0.45                       & \multicolumn{1}{l|}{0.59}     & 0.80                          & 0.36                       & \multicolumn{1}{l|}{0.50}     & 11            \\
\multicolumn{1}{l|}{BLCA} & 0.68 & 0.85 & \multicolumn{1}{l|}{0.76}     & 0.63                          & 0.84                       & \multicolumn{1}{l|}{0.72}     & 0.61                          & 0.84                       & \multicolumn{1}{l|}{0.70}     & 68            \\
\multicolumn{1}{l|}{BLGG} & 0.90 & 0.87 & \multicolumn{1}{l|}{0.88}     & 0.88                          & 0.87                       & \multicolumn{1}{l|}{0.88}     & 0.86                          & 0.85                       & \multicolumn{1}{l|}{0.85}     & 79            \\
\multicolumn{1}{l|}{BRCA} & 0.91 & 0.94 & \multicolumn{1}{l|}{0.93}     & 0.92                          & 0.93                       & \multicolumn{1}{l|}{0.92}     & 0.88                          & 0.96                       & \multicolumn{1}{l|}{0.92}     & 178           \\
\multicolumn{1}{l|}{CESC} & 0.78 & 0.46 & \multicolumn{1}{l|}{0.58}     & 0.87                          & 0.51                       & \multicolumn{1}{l|}{0.65}     & 0.88                          & 0.59                       & \multicolumn{1}{l|}{0.71}     & 39            \\
\multicolumn{1}{l|}{CHOL} & 0.45 & 0.62 & \multicolumn{1}{l|}{0.53}     & 0.50                          & 0.50                       & \multicolumn{1}{l|}{0.50}     & 0.33                          & 0.12                       & \multicolumn{1}{l|}{0.18}     & 8             \\
\multicolumn{1}{l|}{COAD} & 0.64 & 0.68 & \multicolumn{1}{l|}{0.66}     & 0.70                          & 0.73                       & \multicolumn{1}{l|}{0.71}     & 0.71                          & 0.79                       & \multicolumn{1}{l|}{0.75}     & 62            \\
\multicolumn{1}{l|}{ESCA} & 0.45 & 0.50 & \multicolumn{1}{l|}{0.47}     & 0.52                          & 0.43                       & \multicolumn{1}{l|}{0.47}     & 0.44                          & 0.39                       & \multicolumn{1}{l|}{0.42}     & 28            \\
\multicolumn{1}{l|}{GBM} & 0.84 & 0.88 & \multicolumn{1}{l|}{0.86}     & 0.84                          & 0.86                       & \multicolumn{1}{l|}{0.85}     & 0.80                          & 0.83                       & \multicolumn{1}{l|}{0.81}     & 66            \\
\multicolumn{1}{l|}{HNSC} & 0.79 & 0.71 & \multicolumn{1}{l|}{0.75}     & 0.84                          & 0.78                       & \multicolumn{1}{l|}{0.81}     & 0.82                          & 0.73                       & \multicolumn{1}{l|}{0.77}     & 63            \\
\multicolumn{1}{l|}{KICH} & 0.90 & 0.86 & \multicolumn{1}{l|}{0.88}     & 1.00                          & 0.86                       & \multicolumn{1}{l|}{0.93}     & 1.00                          & 0.82                       & \multicolumn{1}{l|}{0.90}     & 22            \\
\multicolumn{1}{l|}{KIRC} & 0.87 & 0.90 & \multicolumn{1}{l|}{0.89}     & 0.89                          & 0.95                       & \multicolumn{1}{l|}{0.92}     & 0.85                          & 0.95                       & \multicolumn{1}{l|}{0.90}     & 99            \\
\multicolumn{1}{l|}{KIRP} & 0.78 & 0.79 & \multicolumn{1}{l|}{0.79}     & 0.84                          & 0.81                       & \multicolumn{1}{l|}{0.83}     & 0.83                          & 0.74                       & \multicolumn{1}{l|}{0.78}     & 53            \\
\multicolumn{1}{l|}{LIHC} & 0.90 & 0.81 & \multicolumn{1}{l|}{0.86}     & 0.85                          & 0.83                       & \multicolumn{1}{l|}{0.84}     & 0.85                          & 0.83                       & \multicolumn{1}{l|}{0.84}     & 70            \\
\multicolumn{1}{l|}{LUAD} & 0.75 & 0.72 & \multicolumn{1}{l|}{0.73}     & 0.77                          & 0.72                       & \multicolumn{1}{l|}{0.74}     & 0.78                          & 0.68                       & \multicolumn{1}{l|}{0.72}     & 74            \\
\multicolumn{1}{l|}{LUSC} & 0.72 & 0.76 & \multicolumn{1}{l|}{0.74}     & 0.73                          & 0.86                       & \multicolumn{1}{l|}{0.79}     & 0.70                          & 0.85                       & \multicolumn{1}{l|}{0.77}     & 84            \\
\multicolumn{1}{l|}{MESO} & 0.40 & 0.22 & \multicolumn{1}{l|}{0.29}     & 0.50                          & 0.11                       & \multicolumn{1}{l|}{0.18}     & 0.00                          & 0.00                       & \multicolumn{1}{l|}{0.00}     & 9             \\
\multicolumn{1}{l|}{OV} & 0.80 & 0.80 & \multicolumn{1}{l|}{0.80}     & 0.80                          & 0.80                       & \multicolumn{1}{l|}{0.80}     & 0.84                          & 0.80                       & \multicolumn{1}{l|}{0.82}     & 20            \\
\multicolumn{1}{l|}{PAAD} & 0.60 & 0.62 & \multicolumn{1}{l|}{0.61}     & 0.62                          & 0.62                       & \multicolumn{1}{l|}{0.62}     & 0.61                          & 0.58                       & \multicolumn{1}{l|}{0.60}     & 24            \\
\multicolumn{1}{l|}{PCPG} & 0.93 & 0.90 & \multicolumn{1}{l|}{0.92}     & 0.93                          & 0.90                       & \multicolumn{1}{l|}{0.92}     & 0.82                          & 0.90                       & \multicolumn{1}{l|}{0.86}     & 30            \\
\multicolumn{1}{l|}{PRAD} & 0.92 & 0.95 & \multicolumn{1}{l|}{0.94}     & 0.94                          & 0.95                       & \multicolumn{1}{l|}{0.94}     & 0.94                          & 0.95                       & \multicolumn{1}{l|}{0.94}     & 77            \\
\multicolumn{1}{l|}{READ} & 0.18 & 0.19 & \multicolumn{1}{l|}{0.19}     & 0.32                          & 0.29                       & \multicolumn{1}{l|}{0.30}     & 0.30                          & 0.14                       & \multicolumn{1}{l|}{0.19}     & 21            \\
\multicolumn{1}{l|}{SARC} & 0.77 & 0.77 & \multicolumn{1}{l|}{0.77}     & 0.77                          & 0.77                       & \multicolumn{1}{l|}{0.77}     & 0.80                          & 0.77                       & \multicolumn{1}{l|}{0.78}     & 26            \\
\multicolumn{1}{l|}{SKCM} & 0.95 & 0.78 & \multicolumn{1}{l|}{0.85}     & 0.88                          & 0.76                       & \multicolumn{1}{l|}{0.81}     & 0.83                          & 0.71                       & \multicolumn{1}{l|}{0.77}     & 49            \\
\multicolumn{1}{l|}{STAD} & 0.66 & 0.71 & \multicolumn{1}{l|}{0.68}     & 0.68                          & 0.78                       & \multicolumn{1}{l|}{0.73}     & 0.69                          & 0.78                       & \multicolumn{1}{l|}{0.74}     & 55            \\
\multicolumn{1}{l|}{TGCT} & 0.95 & 0.81 & \multicolumn{1}{l|}{0.88}     & 0.96                          & 0.85                       & \multicolumn{1}{l|}{0.90}     & 0.95                          & 0.81                       & \multicolumn{1}{l|}{0.88}     & 26            \\
\multicolumn{1}{l|}{THYM} & 1.00 & 0.67 & \multicolumn{1}{l|}{0.80}     & 1.00                          & 0.67                       & \multicolumn{1}{l|}{0.80}     & 1.00                          & 0.50                       & \multicolumn{1}{l|}{0.67}     & 6             \\
\multicolumn{1}{l|}{THCA} & 0.94 & 0.97 & \multicolumn{1}{l|}{0.96}     & 0.94                          & 0.98                       & \multicolumn{1}{l|}{0.96}     & 0.97                          & 0.99                       & \multicolumn{1}{l|}{0.98}     & 101           \\
\multicolumn{1}{l|}{UCS} & 0.67 & 1.00 & \multicolumn{1}{l|}{0.80}     & 0.67                          & 1.00                       & \multicolumn{1}{l|}{0.80}     & 0.67                          & 1.00                       & \multicolumn{1}{l|}{0.80}     & 6             \\
\multicolumn{1}{l|}{UVM} & 1.00 & 0.88 & \multicolumn{1}{l|}{0.93}     & 1.00                          & 0.88                       & \multicolumn{1}{l|}{0.93}     & 1.00                          & 0.88                       & \multicolumn{1}{l|}{0.93}     & 8             \\ \hline
\multicolumn{1}{l|}{\textbf{Total Slides}}                                       &                               &                            & \multicolumn{1}{l|}{}         &                               &                            & \multicolumn{1}{l|}{}         &                               &                            & \multicolumn{1}{l|}{}         & \textbf{1462} \\ \hline
\end{tabular}}
\caption{Detailed precision, recall, f1-score, and the number of slides processed for each subtype are shown in this table using the Yottixel mosaic. The evaluations are based on the top 1 retrieval, the majority among the top 3 retrievals, and the majority among the top 5 retrievals using the TCGA dataset.}\label{tab:TCGA_results_YT}
\end{table}

\begin{table}[!]
\centering
\resizebox{\columnwidth}{!}{\begin{tabular}{llllllllllr}

\multicolumn{11}{c}{\textbf{SDM Montage}}                                                                                                                                                                                                                                                                                                                                                        \\ \hline
\multicolumn{1}{l|}{}                                                            & \multicolumn{3}{c|}{Top-1}                                                                 & \multicolumn{3}{c|}{MV@3}                                                                  & \multicolumn{3}{c|}{MV@5}                                                                  &               \\ \hline
\multicolumn{1}{l|}{\begin{tabular}[c]{@{}l@{}}Primary\\ Diagnoses\end{tabular}} & \multicolumn{1}{c}{Precision} & \multicolumn{1}{c}{Recall} & \multicolumn{1}{c|}{f1-score} & \multicolumn{1}{c}{Precision} & \multicolumn{1}{c}{Recall} & \multicolumn{1}{c|}{f1-score} & \multicolumn{1}{c}{Precision} & \multicolumn{1}{c}{Recall} & \multicolumn{1}{c|}{f1-score} & Slides        \\ \hline
\multicolumn{1}{l|}{ACC} & 0.89 & 0.73 & \multicolumn{1}{l|}{0.80}     & 0.80                          & 0.73                       & \multicolumn{1}{l|}{0.76}     & 0.80                          & 0.73                       & \multicolumn{1}{l|}{0.76}     & 11            \\
\multicolumn{1}{l|}{BLCA} & 0.74 & 0.79 & \multicolumn{1}{l|}{0.77}     & 0.68                          & 0.82                       & \multicolumn{1}{l|}{0.75}     & 0.66                          & 0.87                       & \multicolumn{1}{l|}{0.75}     & 68            \\
\multicolumn{1}{l|}{BLGG} & 0.82 & 0.82 & \multicolumn{1}{l|}{0.82}     & 0.86                          & 0.87                       & \multicolumn{1}{l|}{0.87}     & 0.83                          & 0.85                       & \multicolumn{1}{l|}{0.84}     & 79            \\
\multicolumn{1}{l|}{BRCA} & 0.89 & 0.96 & \multicolumn{1}{l|}{0.92}     & 0.88                          & 0.98                       & \multicolumn{1}{l|}{0.93}     & 0.84                          & 0.99                       & \multicolumn{1}{l|}{0.91}     & 178           \\
\multicolumn{1}{l|}{CESC} & 0.80 & 0.62 & \multicolumn{1}{l|}{0.70}     & 0.66                          & 0.54                       & \multicolumn{1}{l|}{0.59}     & 0.75                          & 0.62                       & \multicolumn{1}{l|}{0.68}     & 39            \\
\multicolumn{1}{l|}{CHOL} & 0.50 & 0.25 & \multicolumn{1}{l|}{0.33}     & 0.67                          & 0.25                       & \multicolumn{1}{l|}{0.36}     & 0.00                          & 0.00                       & \multicolumn{1}{l|}{0.00}     & 8             \\
\multicolumn{1}{l|}{COAD} & 0.71 & 0.79 & \multicolumn{1}{l|}{0.75}     & 0.70                          & 0.77                       & \multicolumn{1}{l|}{0.73}     & 0.71                          & 0.84                       & \multicolumn{1}{l|}{0.77}     & 62            \\
\multicolumn{1}{l|}{ESCA} & 0.43 & 0.43 & \multicolumn{1}{l|}{0.43}     & 0.54                          & 0.50                       & \multicolumn{1}{l|}{0.52}     & 0.58                          & 0.54                       & \multicolumn{1}{l|}{0.56}     & 28            \\
\multicolumn{1}{l|}{GBM} & 0.80 & 0.81 & \multicolumn{1}{l|}{0.81}     & 0.86                          & 0.84                       & \multicolumn{1}{l|}{0.85}     & 0.81                          & 0.80                       & \multicolumn{1}{l|}{0.81}     & 70            \\
\multicolumn{1}{l|}{HNSC} & 0.82 & 0.78 & \multicolumn{1}{l|}{0.80}     & 0.83                          & 0.76                       & \multicolumn{1}{l|}{0.79}     & 0.88                          & 0.79                       & \multicolumn{1}{l|}{0.83}     & 63            \\
\multicolumn{1}{l|}{KICH} & 0.95 & 0.82 & \multicolumn{1}{l|}{0.88}     & 0.95                          & 0.91                       & \multicolumn{1}{l|}{0.93}     & 1.00                          & 0.86                       & \multicolumn{1}{l|}{0.93}     & 22            \\
\multicolumn{1}{l|}{KIRC} & 0.91 & 0.90 & \multicolumn{1}{l|}{0.90}     & 0.93                          & 0.94                       & \multicolumn{1}{l|}{0.93}     & 0.88                          & 0.95                       & \multicolumn{1}{l|}{0.91}     & 99            \\
\multicolumn{1}{l|}{KIRP} & 0.75 & 0.83 & \multicolumn{1}{l|}{0.79}     & 0.79                          & 0.83                       & \multicolumn{1}{l|}{0.81}     & 0.84                          & 0.79                       & \multicolumn{1}{l|}{0.82}     & 53            \\
\multicolumn{1}{l|}{LIHC} & 0.82 & 0.80 & \multicolumn{1}{l|}{0.81}     & 0.84                          & 0.81                       & \multicolumn{1}{l|}{0.83}     & 0.84                          & 0.84                       & \multicolumn{1}{l|}{0.84}     & 70            \\
\multicolumn{1}{l|}{LUAD} & 0.76 & 0.73 & \multicolumn{1}{l|}{0.74}     & 0.71                          & 0.74                       & \multicolumn{1}{l|}{0.73}     & 0.75                          & 0.74                       & \multicolumn{1}{l|}{0.75}     & 74            \\
\multicolumn{1}{l|}{LUSC} & 0.72 & 0.75 & \multicolumn{1}{l|}{0.74}     & 0.77                          & 0.76                       & \multicolumn{1}{l|}{0.77}     & 0.78                          & 0.77                       & \multicolumn{1}{l|}{0.78}     & 84            \\
\multicolumn{1}{l|}{MESO} & 0.67 & 0.22 & \multicolumn{1}{l|}{0.33}     & 1.00                          & 0.11                       & \multicolumn{1}{l|}{0.20}     & 1.00                          & 0.11                       & \multicolumn{1}{l|}{0.20}     & 9             \\
\multicolumn{1}{l|}{OV} & 0.88 & 0.75 & \multicolumn{1}{l|}{0.81}     & 0.84                          & 0.80                       & \multicolumn{1}{l|}{0.82}     & 0.83                          & 0.75                       & \multicolumn{1}{l|}{0.79}     & 20            \\
\multicolumn{1}{l|}{PAAD} & 0.64 & 0.58 & \multicolumn{1}{l|}{0.61}     & 0.60                          & 0.50                       & \multicolumn{1}{l|}{0.55}     & 0.68                          & 0.54                       & \multicolumn{1}{l|}{0.60}     & 24            \\
\multicolumn{1}{l|}{PCPG} & 0.90 & 0.87 & \multicolumn{1}{l|}{0.88}     & 0.93                          & 0.83                       & \multicolumn{1}{l|}{0.88}     & 0.96                          & 0.83                       & \multicolumn{1}{l|}{0.89}     & 30            \\
\multicolumn{1}{l|}{PRAD} & 0.93 & 0.96 & \multicolumn{1}{l|}{0.94}     & 0.95                          & 0.96                       & \multicolumn{1}{l|}{0.95}     & 0.94                          & 0.96                       & \multicolumn{1}{l|}{0.95}     & 77            \\
\multicolumn{1}{l|}{READ} & 0.31 & 0.24 & \multicolumn{1}{l|}{0.27}     & 0.31                          & 0.19                       & \multicolumn{1}{l|}{0.24}     & 0.33                          & 0.19                       & \multicolumn{1}{l|}{0.24}     & 21            \\
\multicolumn{1}{l|}{SARC} & 0.83 & 0.73 & \multicolumn{1}{l|}{0.78}     & 0.86                          & 0.69                       & \multicolumn{1}{l|}{0.77}     & 0.90                          & 0.73                       & \multicolumn{1}{l|}{0.81}     & 26            \\
\multicolumn{1}{l|}{SKCM} & 0.80 & 0.82 & \multicolumn{1}{l|}{0.81}     & 0.86                          & 0.76                       & \multicolumn{1}{l|}{0.80}     & 0.87                          & 0.67                       & \multicolumn{1}{l|}{0.76}     & 49            \\
\multicolumn{1}{l|}{STAD} & 0.74 & 0.84 & \multicolumn{1}{l|}{0.79}     & 0.72                          & 0.84                       & \multicolumn{1}{l|}{0.77}     & 0.74                          & 0.82                       & \multicolumn{1}{l|}{0.78}     & 55            \\
\multicolumn{1}{l|}{TGCT} & 0.81 & 0.81 & \multicolumn{1}{l|}{0.81}     & 0.85                          & 0.85                       & \multicolumn{1}{l|}{0.85}     & 0.88                          & 0.85                       & \multicolumn{1}{l|}{0.86}     & 26            \\
\multicolumn{1}{l|}{THYM} & 0.80 & 0.67 & \multicolumn{1}{l|}{0.73}     & 1.00                          & 0.67                       & \multicolumn{1}{l|}{0.80}     & 1.00                          & 0.33                       & \multicolumn{1}{l|}{0.50}     & 6             \\
\multicolumn{1}{l|}{THCA} & 0.98 & 0.98 & \multicolumn{1}{l|}{0.98}     & 0.98                          & 0.98                       & \multicolumn{1}{l|}{0.98}     & 0.98                          & 0.98                       & \multicolumn{1}{l|}{0.98}     & 101           \\
\multicolumn{1}{l|}{UCS} & 0.75 & 1.00 & \multicolumn{1}{l|}{0.86}     & 0.75                          & 1.00                       & \multicolumn{1}{l|}{0.86}     & 0.75                          & 1.00                       & \multicolumn{1}{l|}{0.86}     & 6             \\
\multicolumn{1}{l|}{UVM} & 1.00 & 0.88 & \multicolumn{1}{l|}{0.93}     & 1.00                          & 0.88                       & \multicolumn{1}{l|}{0.93}     & 1.00                          & 0.88                       & \multicolumn{1}{l|}{0.93}     & 8             \\ \hline
\multicolumn{1}{l|}{\textbf{Total Slides}}                                       &                               &                            & \multicolumn{1}{l|}{}         &                               &                            & \multicolumn{1}{l|}{}         &                               &                            & \multicolumn{1}{l|}{}         & \textbf{1466} \\ \hline
\end{tabular}}
\caption{Detailed precision, recall, f1-score, and the number of slides processed for each subtype are shown in this table using the SDM Montage. The evaluations are based on the top 1 retrieval, the majority among the top 3 retrievals, and the majority among the top 5 retrievals using the TCGA dataset.}\label{tab:TCGA_results_SDM}
\end{table}

\clearpage\subsection{Public -- BReAst Carcinoma Subtyping (BRACS)}
\label{sec:Exd_Results_BRACS}
The BRACS dataset comprises a total of 547 WSIs derived from 189 distinct patients~\cite{brancati2022bracs}. In the context of the leave-one-out search and matching experiment, all 547 WSIs were employed from the dataset. Notably, all slides have been scanned utilizing an Aperio AT2 scanner, with a resolution of 0.25 $\mu m$ per pixel and a magnification factor of $40\times$. The dataset is categorized into two main subsets: WSI and Region of Interest (ROI). Within the WSI subset, there are three primary tumor Groups~\cite{brancati2022bracs}. Whereas, the ROI subset is divided into seven distinct tumor types~\cite{brancati2022bracs}. For this study, since we are conducting a WSI-to-WSI matching, we utilized the WSI subset to perform histological matching. Table~\ref{tab:BRACS} shows more details about the data used in this experiment (see Table~\ref{tab:BRACS} for more details).

\begin{table}[t] 
\centering
\resizebox{\columnwidth}{!}{\begin{tabular}{llcccc}
\hline
Primary Diagnoses           & Acronyms & Slides & Group                                                                        & \begin{tabular}[c]{@{}c@{}}Group\\ Acronyms\end{tabular} & Slides               \\ \hline
Atypical Ductal Hyperplasia & ADH      & 48     & \multirow{2}{*}{\begin{tabular}[c]{@{}c@{}}Atypical\\ Tumours\end{tabular}}  & \multirow{2}{*}{AT}                                      & \multirow{2}{*}{89}  \\
Flat Epithelial Atypia      & FEA      & 41     &                                                                              &                                                          &                      \\ \hline
Normal                      & N        & 44     & \multirow{3}{*}{\begin{tabular}[c]{@{}c@{}}Benign\\ Tumours\end{tabular}}    & \multirow{3}{*}{BT}                                      & \multirow{3}{*}{265} \\
Pathological Benign         & PB       & 147    &                                                                              &                                                          &                      \\
Usual Ductal Hyperplasia    & UDH      & 74     &                                                                              &                                                          &                      \\ \hline
Ductal Carcinoma in Situ    & DCIS     & 61     & \multirow{2}{*}{\begin{tabular}[c]{@{}c@{}}Malignant\\ Tumours\end{tabular}} & \multirow{2}{*}{MT}                                      & \multirow{2}{*}{193} \\
Invasive Carcinoma          & IC       & 132    &                                                                              &                                                          &                      \\ \hline
\end{tabular}}
\caption{Information concerning the BRACS dataset employed in this experiment, inclusive of the respective acronyms and the number of slides associated with each primary diagnosis and group.}\label{tab:BRACS}
\end{table}

\begin{figure*}[t]
\centerline{\includegraphics[width =  1\textwidth]{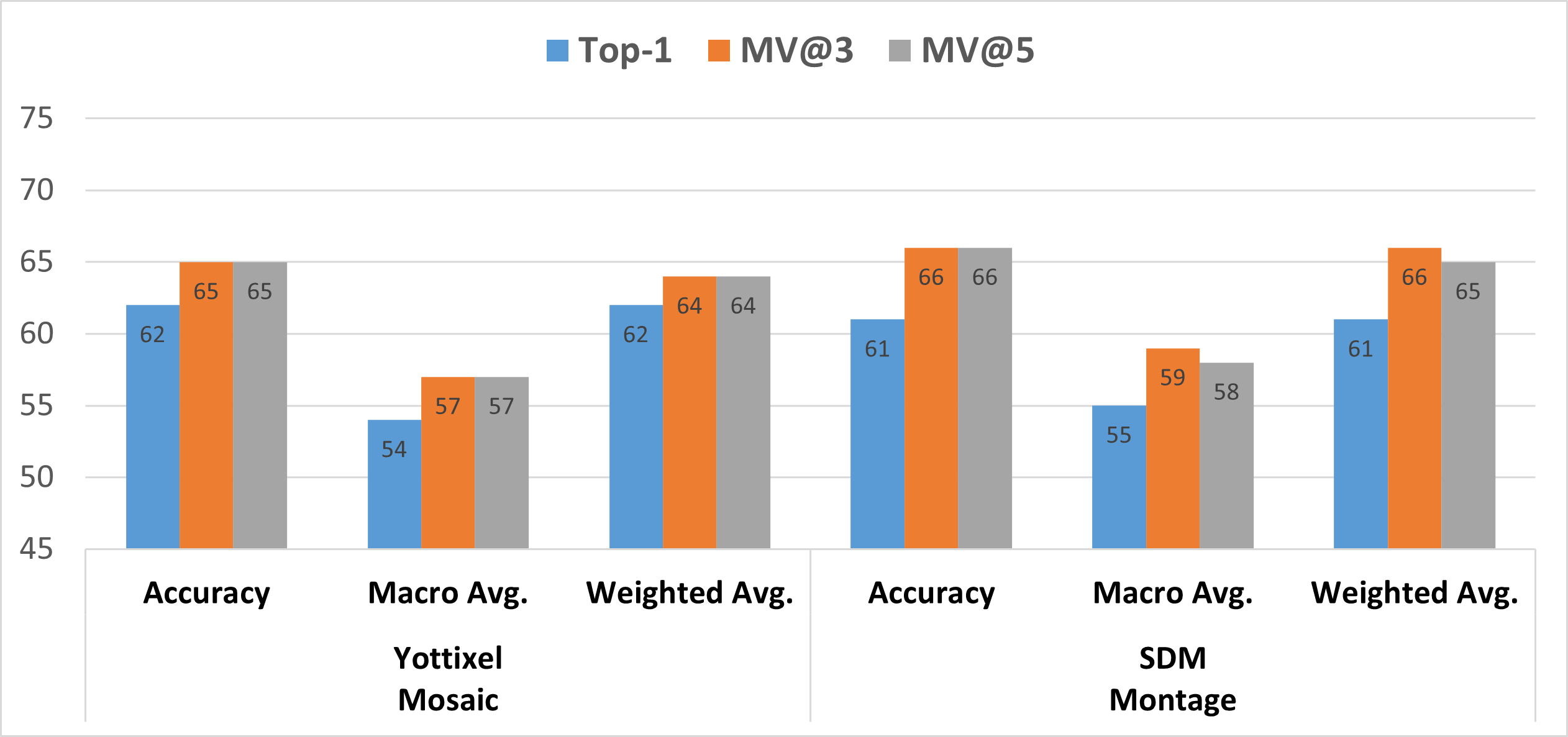}}
\caption{Accuracy, macro average of f1-scores, and weighted average of f1-scores are reported from Yottixel mosaic, and SDM montage. The evaluations are based on the top 1 retrieval, the majority among the top 3 retrievals, and the majority among the top 5 retrievals using the BRACS dataset.}
\label{fig:Bracs_Accuracy}
\end{figure*}

\begin{figure*}[t]
\centerline{\includegraphics[width =  0.6\textwidth]{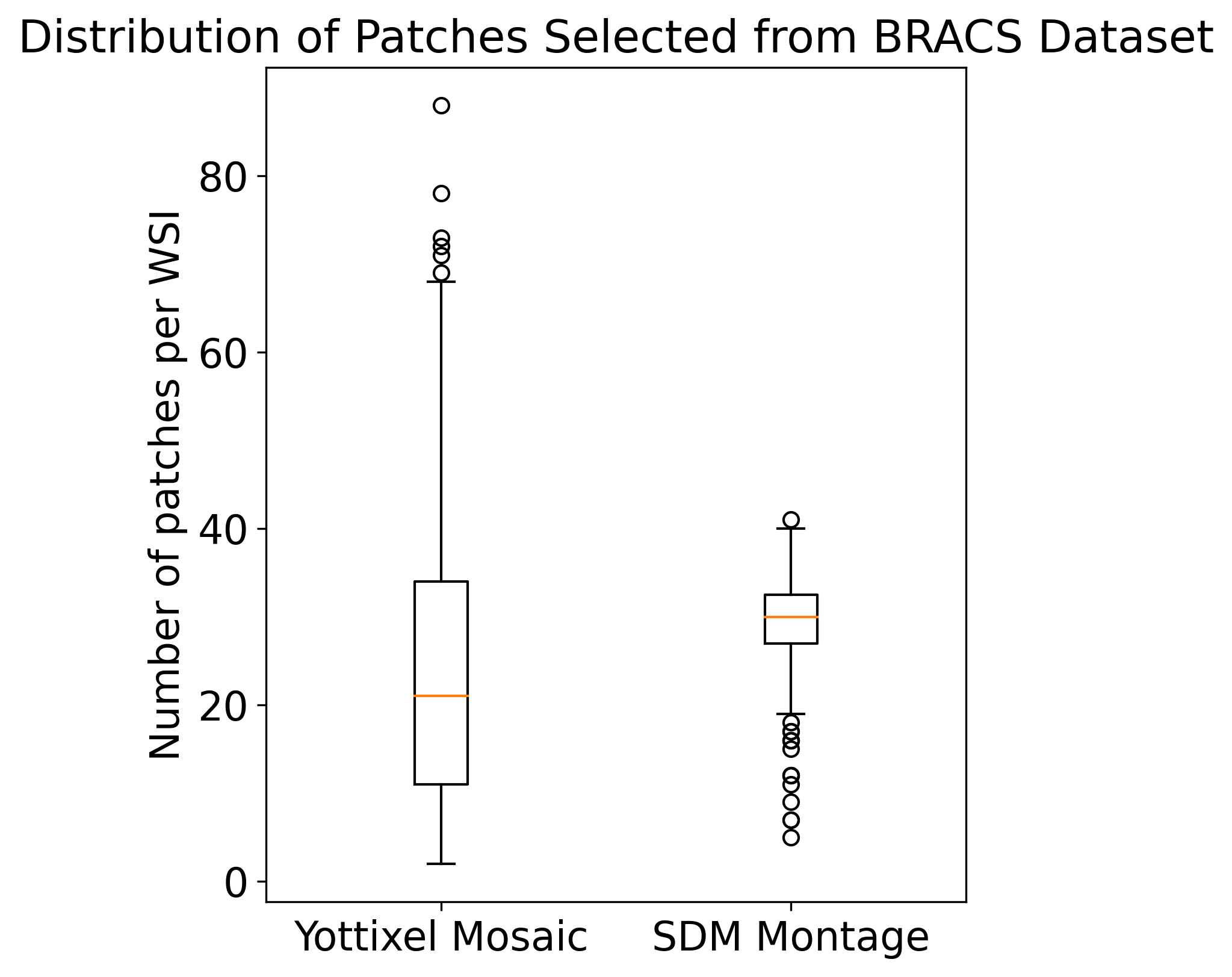}}
\caption{The boxplot illustrates the distribution of patches selected for each WSI in the BRACS dataset from both the Yottixel Mosaic and SDM Montage. Additionally, it provides statistical measures for these distributions. Specifically, for the Yottixel Mosaic, the median number of selected patches is $21\pm16$. On the other hand, for the SDM Montage, the median number of selected patches is $30\pm5$.}
\label{fig:BRACS_Patch_boxplot}
\end{figure*}

\begin{figure*}[t]
\centerline{\includegraphics[width =  1\textwidth]{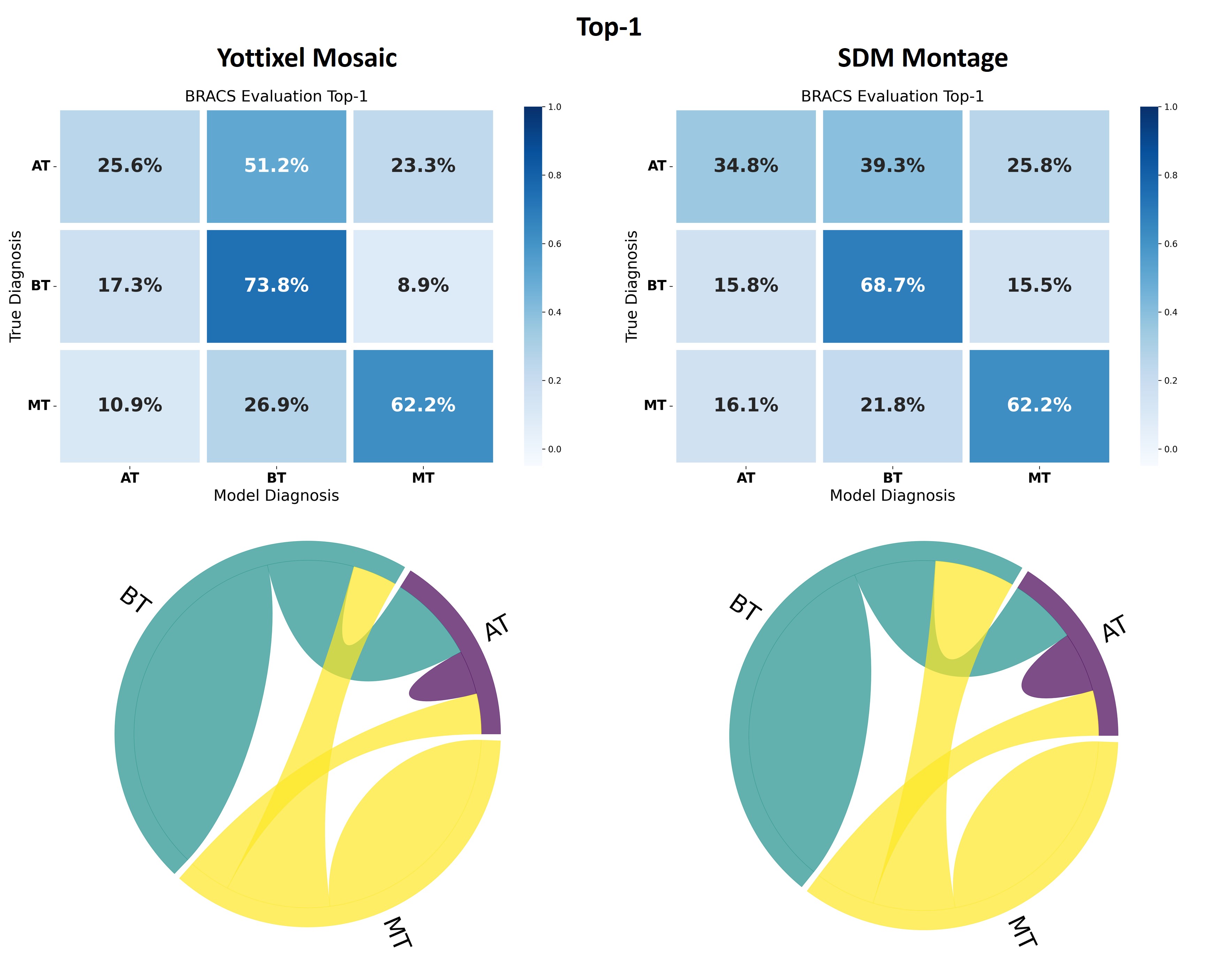}}
\caption{Confusion matrices and chord diagrams from Yottixel mosaic (left column), and SDM montage (right column). The evaluations are based on the top 1 retrieval when evaluating the BRACS dataset.}
\label{fig:Bracs_Chord1}
\end{figure*}

\begin{figure*}[t]
\centerline{\includegraphics[width =  1\textwidth]{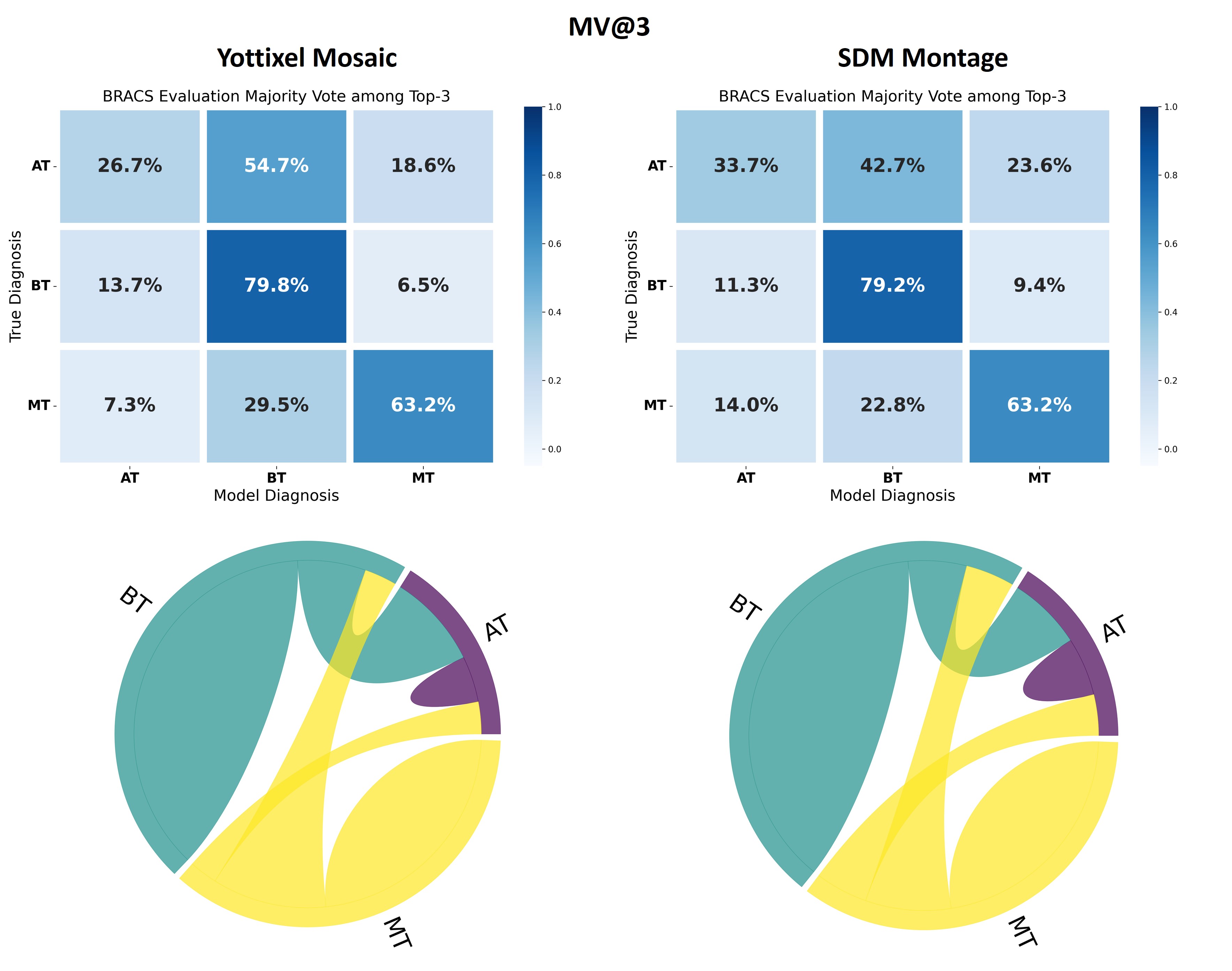}}
\caption{Confusion matrices and chord diagrams from Yottixel mosaic (left column), and SDM montage (right column). The evaluations are based on the majority of the top 3 retrievals when evaluating the BRACS dataset.}
\label{fig:Bracs_Chord3}
\end{figure*}

\begin{figure*}[t]
\centerline{\includegraphics[width =  1\textwidth]{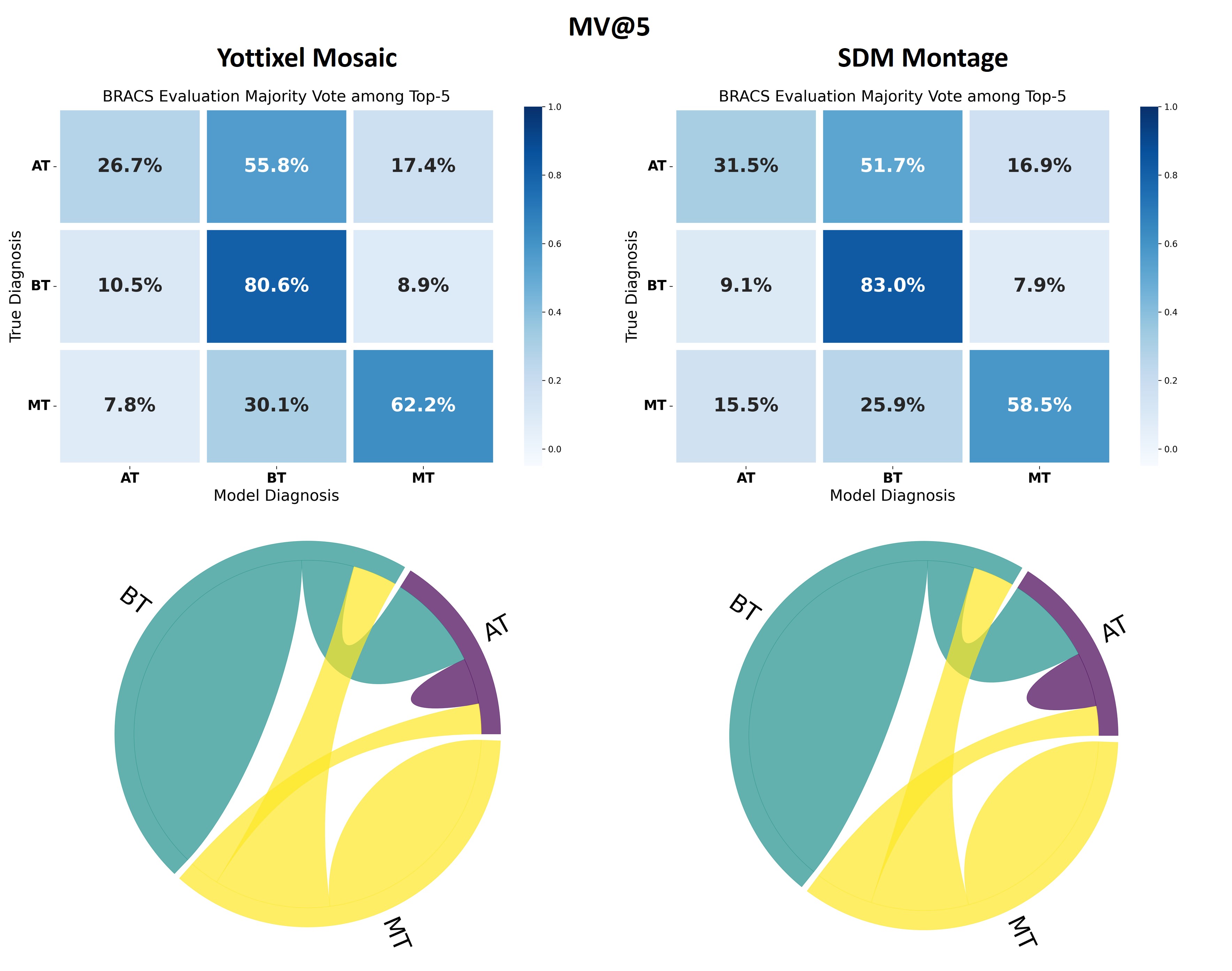}}
\caption{Confusion matrices and chord diagrams from Yottixel mosaic (left column), and SDM montage (right column). The evaluations are based on the majority of the top 5 retrievals when evaluating the BRACS dataset.}
\label{fig:Bracs_Chord5}
\end{figure*}

\begin{figure*}[t]
\centerline{\includegraphics[width =  1\textwidth]{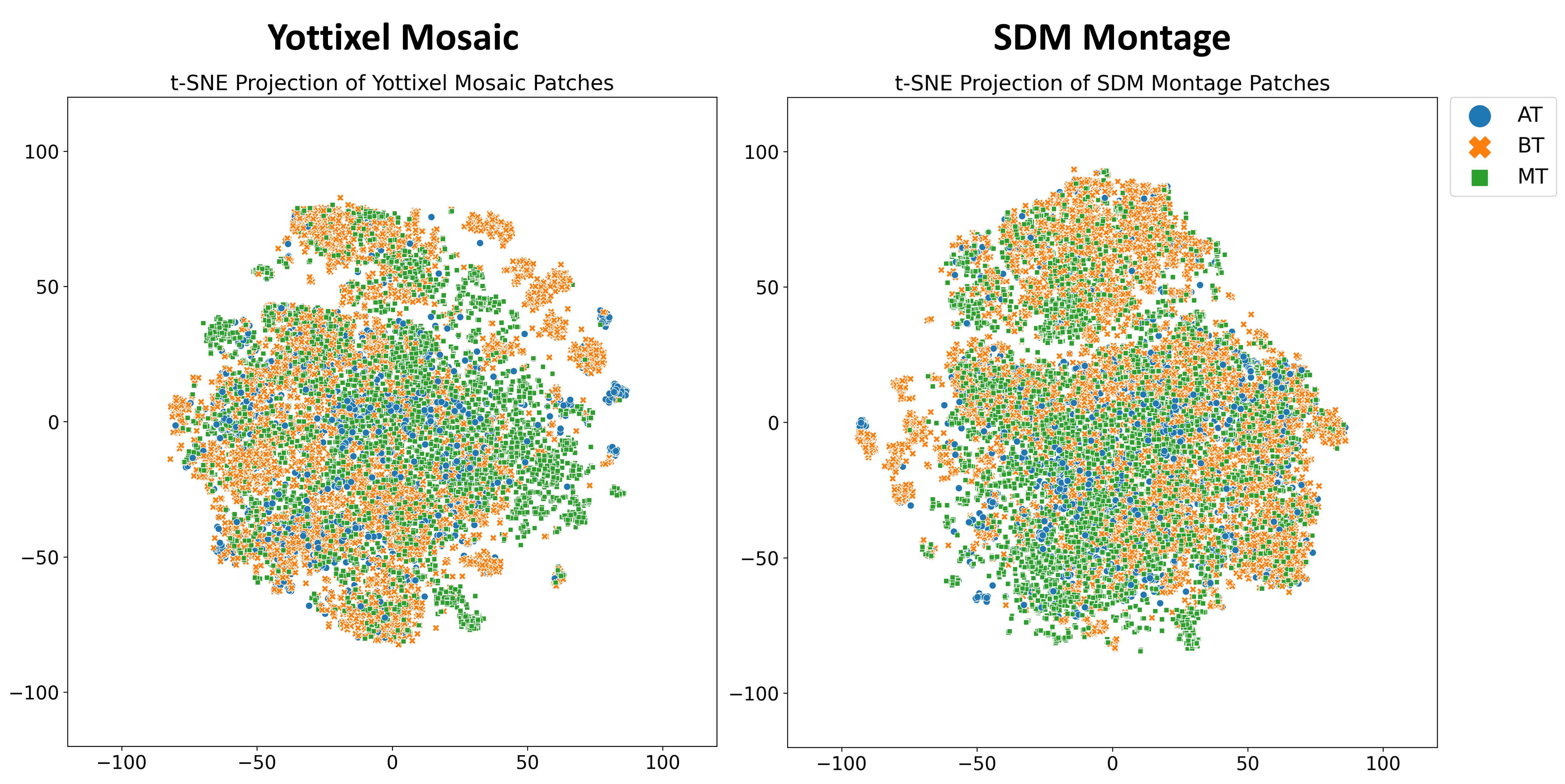}}
\caption{The t-SNE projection displays the embeddings of all patches extracted from the BRACS dataset using Yottixel's mosaic (left) and SDM's montage (right).}
\label{fig:BRACS_TSNE}
\end{figure*}

To evaluate the performance of the SDM montage against Yottixel's mosaic, we retrieved the top similar cases using leave-one-out evaluation. The assessments rely on several retrieval criteria, including the top-1, MV@3, and MV@5. The accuracy, macro average, and weighted average at top-1, MV@3, and MV@5 are shown in Figure~\ref{fig:Bracs_Accuracy}. Table~\ref{tab:BRACS_results} shows the detailed results including precision, recall, and f1-score. Moreover, confusion matrices and chord diagrams at Top-1, MV@3, and MV@5 are shown in Figure~\ref{fig:Bracs_Chord1}, \ref{fig:Bracs_Chord3}, and ~\ref{fig:Bracs_Chord5}, respectively. In addition to these accuracy metrics, a comparative analysis of the number of patches extracted per WSI by each respective method is also presented in Figure~\ref{fig:BRACS_Patch_boxplot} for a visual representation of the distribution over the entire dataset. To visually illustrate the extracted patches, we used t-SNE projections, as demonstrated in Figure~\ref{fig:BRACS_TSNE}.

\begin{table*}[t]
\centering
\resizebox{\textwidth}{!}{\begin{tabular}{c|l|lll|lll|lll|c}
\hline
                                 &                                                             & \multicolumn{3}{c|}{\textbf{Top-1}}    & \multicolumn{3}{c|}{\textbf{MV@3}}     & \multicolumn{3}{c|}{\textbf{MV@5}}     &        \\ \hline
                                 & \begin{tabular}[c]{@{}l@{}}Groups\end{tabular} & Precision & Recall & f1-score & Precision & Recall & f1-score & Precision & Recall & f1-score & Slides \\ \hline
\multirow{3}{*}{\rotatebox[origin=c]{0}{\textbf{\begin{tabular}[c]{@{}c@{}}Yottixel\\Mosaic~\cite{kalra2020Yottixel}\end{tabular}}}} 
& AT  & 0.26     & 0.26   & 0.26      & 0.32    &  0.27    &  0.29     & 0.36     & 0.27    &  0.31     & 86     \\
& BT  & 0.66      & 0.74   & 0.69     & 0.66    &  0.80    &  0.72     & 0.65     & 0.81    &  0.72     & 248     \\
& MT   & 0.74      & 0.62   & 0.68    & 0.79    &  0.63    &  0.70     & 0.76     & 0.62    &  0.69     & 193     \\ \hline
 &\textbf{Total Slides}&&&&&&&&&&  \textbf{527}\\ \hline
\multirow{3}{*}{\rotatebox[origin=c]{0}{\textbf{\begin{tabular}[c]{@{}c@{}}SDM\\Montage\end{tabular}}}}  
& AT  & 0.30  &   0.35 &   0.32   &     0.34       &  0.34  &   0.34   &    0.34     &  0.31   &  0.33   &   89     \\
 & BT  &    0.70       &    0.69    &    0.69      &   0.72  &   0.79   &    0.75      & 0.70      &  0.83      &  0.76         &   265     \\
 & MT   &     0.65      &   0.62     &    0.64      &   0.73 &    0.63  &    0.68      &   0.76     &   0.59     &    0.66       &    193    \\  \hline
&\textbf{Total Slides}&&&&&&&&&&  \textbf{547}\\ \hline
\end{tabular}}
\caption{Precision, recall, f1-score, and the number of slides processed for each subtype are reported in this table using Yottixel mosaic, and SDM montage. The evaluations are based on the top 1 retrieval, the majority among the top 3 retrievals, and the majority among the top 5 retrievals using the BRACS dataset.}\label{tab:BRACS_results}
\end{table*}

 In the course of this experiment, the SDM montage demonstrated the performance advantage over the Yottixel mosaic. Notably, it exhibited improvements of +1\%, +2\%, and +1\% in the macro average of F1-scores concerning top-1 retrieval, majority agreement among the top 3 retrievals, and majority agreement among the top 5 retrievals, respectively. In terms of accuracy, SDM underperforms at top-1 retrieval by one percent whereas it outperforms at MV@3 and MV@5 retrievals by one percent. These findings underscore the method's effectiveness in capturing relevant information within the specific context of retrieval, as visualized in Figure~\ref{fig:Bracs_Accuracy}. Furthermore, our analysis unveiled an important aspect of Yottixel's behavior in comparison to SDM. Specifically, our investigation revealed that Yottixel failed to process some WSIs and it processed a total of 527 WSIs, whereas SDM demonstrated a more comprehensive approach by successfully processing all 547 WSIs as shown in Table~\ref{tab:BRACS_results}. This observation highlights the robustness and completeness of the SDM method in managing the entire dataset, further emphasizing its advantages in applications related to the analysis and retrieval of WSIs. In contrast to the Yottixel mosaic, SDM exhibits reduced variability in the number of patches per WSI as seen in Figure~\ref{fig:BRACS_Patch_boxplot}. This is attributed to the absence of an empirical parameter dictating patch selection, as opposed to Yottixel's approach of utilizing 5\% of the total patches. Such a methodological shift not only optimizes storage utilization but also curtails redundancy and obviates the necessity for empirical determination of an optimal patch count.

\clearpage\subsection{Public -- Prostate cANcer graDe Assessment (PANDA)}
\label{sec:Exd_Results_PANDA}
PANDA is the largest publicly available dataset of prostate biopsies, put together for a global AI competition~\cite{bulten2022artificial}. The data is provided by Karolinska Institute, Solna, Sweden, and Radboud University Medical Center (RUMC), Nijmegen, Netherlands. All slides from RUMC were scanned at $20\times$ using a 3DHistech Pannoramic Flash II 250 scan. On the other hand, all the WSIs from Karolinska Institute were digitized at $20\times$ using a Hamamatsu C9600-12 scanner, and an Aperio ScanScope AT2 scanner. In entirety, a dataset comprising 12,625 whole slide images (WSIs) of prostate biopsies was amassed and partitioned into 10,616 WSIs for training and 2,009 WSIs for evaluation purposes. In our experiment, we used the publicly available training cohort of 10,616 WSIs with their International Society of Urological Pathology (ISUP) scores for an extensive leave-one-out search and matching experiment (see Table.~\ref{tab:PANDA} for more details). 

In recent years, there have been significant advancements in both the diagnosis and treatment of prostate cancer. As we entered the new millennium, there was a significant effort to update and modernize the Gleason system. In 2005, the ISUP organized a consensus conference. The gathering attempted to provide a clearer understanding of the patterns that make up different Gleason grades. It also established practical guidelines for how to apply these patterns and introduced what is now known as the ISUP score from zero to five based on the severity of the cancer~\cite{epstein20052005, bulten2022artificial}.

\begin{table}[t] 
\centering
\begin{tabular}{cr}
\hline
\begin{tabular}[c]{@{}l@{}}ISUP Grade\end{tabular} & Slides \\ \hline
0      & 2889     \\
1      & 2665     \\
2      & 1343     \\ 
3      & 1242    \\
4      & 1246     \\
5      & 1223     \\  \hline
\end{tabular}
\caption{Comprehensive dataset particulars pertaining to the Prostate cANcer graDe Assessment (PANDA) dataset, encompassing relevant ISUP grade and the number of slides attributed to each grade.}\label{tab:PANDA}
\end{table}

\begin{figure*}[t]
\centerline{\includegraphics[width =  1\textwidth]{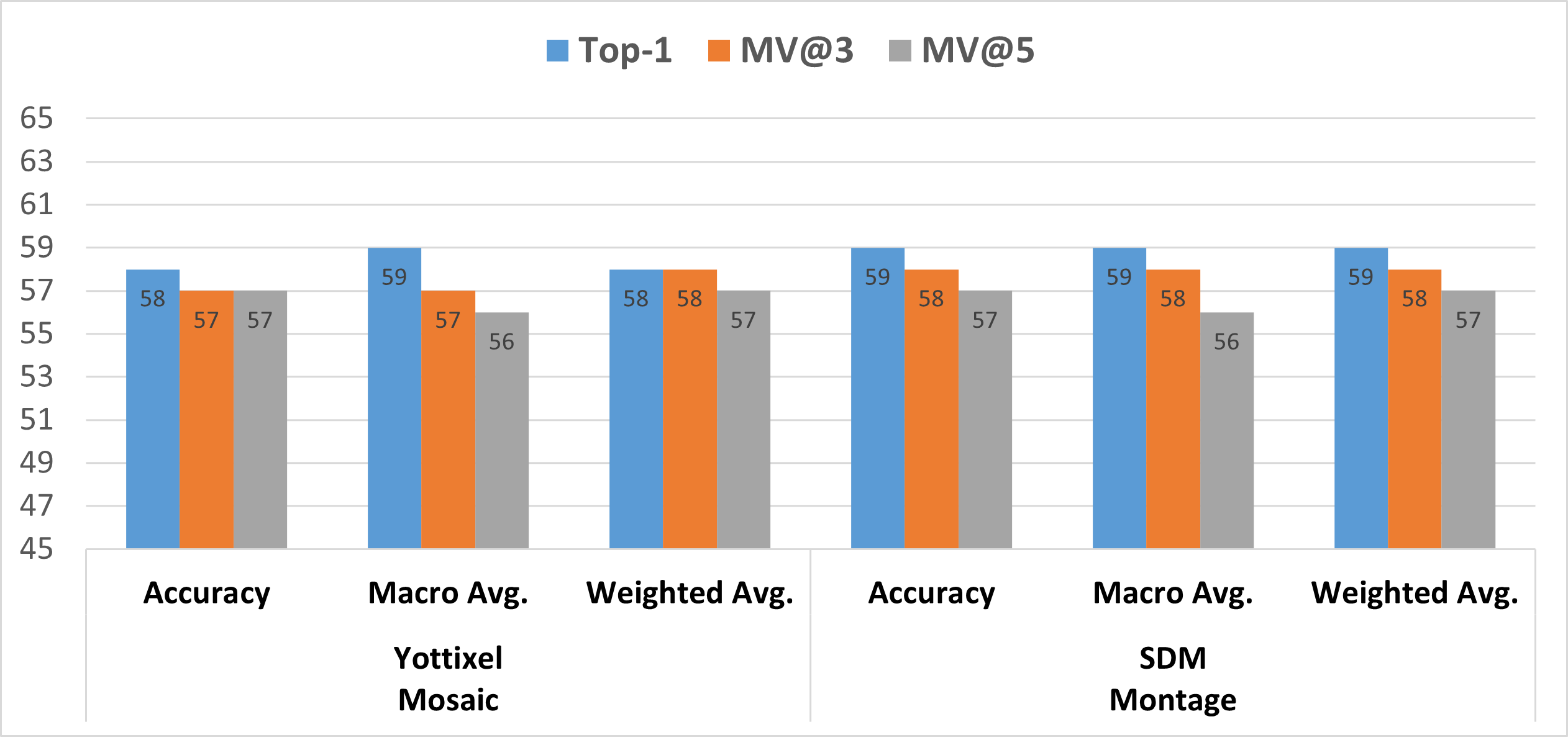}}
\caption{Accuracy, macro average of f1-scores, and weighted average of f1-scores are shown from Yottixel mosaic, and SDM montage. The evaluations are based on the top 1 retrieval, the majority among the top 3 retrievals, and the majority among the top 5 retrievals in the PANDA dataset.}
\label{fig:PANDA_Accuracy}
\end{figure*}

\begin{figure*}[t]
\centerline{\includegraphics[width =  0.6\textwidth]{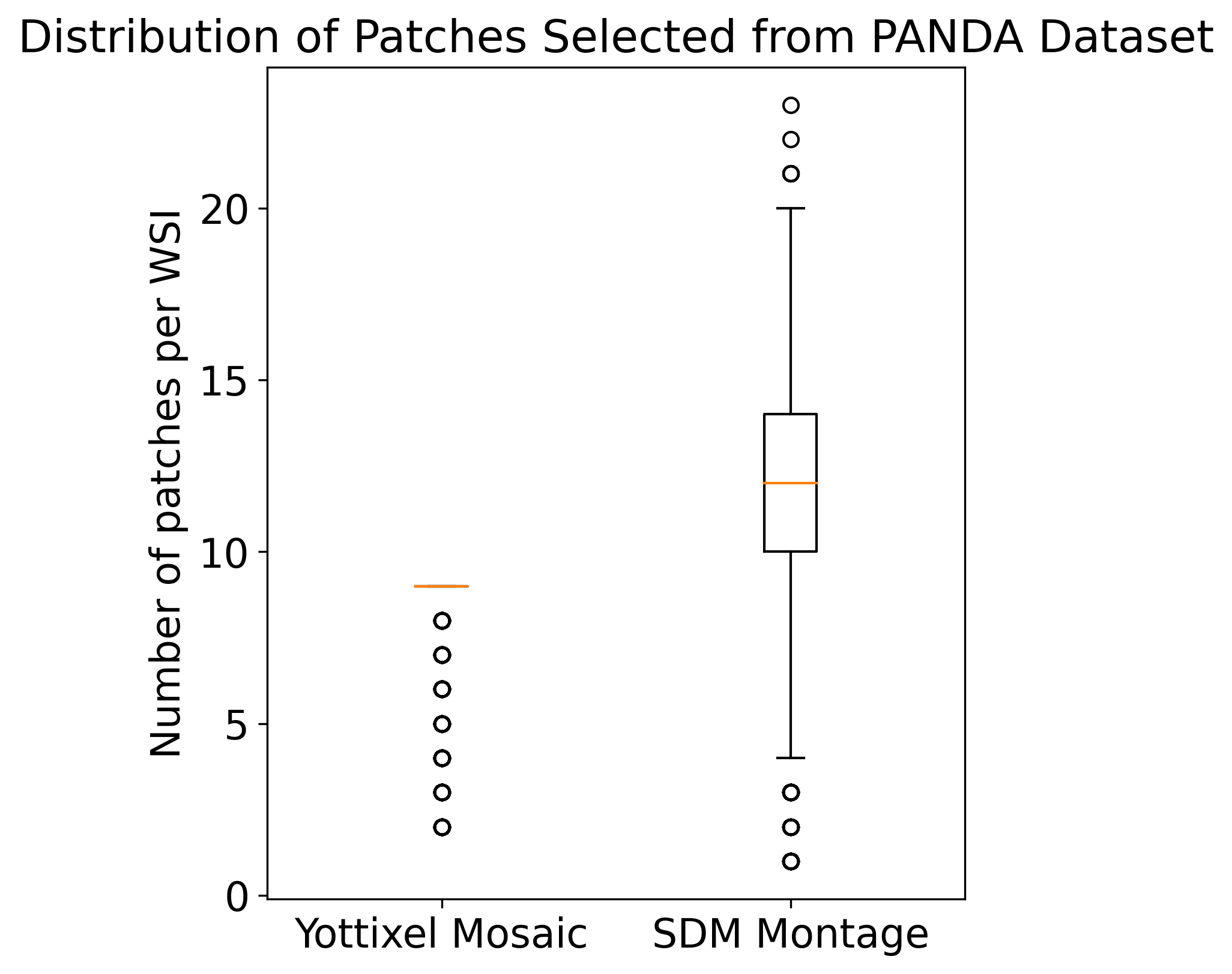}}
\caption{The boxplot illustrates the distribution of patches selected for each WSI in the PANDA dataset from both the Yottixel Mosaic and SDM Montage. Additionally, it provides statistical measures for these distributions. Specifically, for the Yottixel Mosaic, the median number of selected patches is $9\pm2$. On the other hand, for the SDM Montage, the median number of selected patches is $12\pm3$.}
\label{fig:PANDA_Patch_boxplot}
\end{figure*}

\begin{figure*}[t]
\centerline{\includegraphics[width =  1\textwidth]{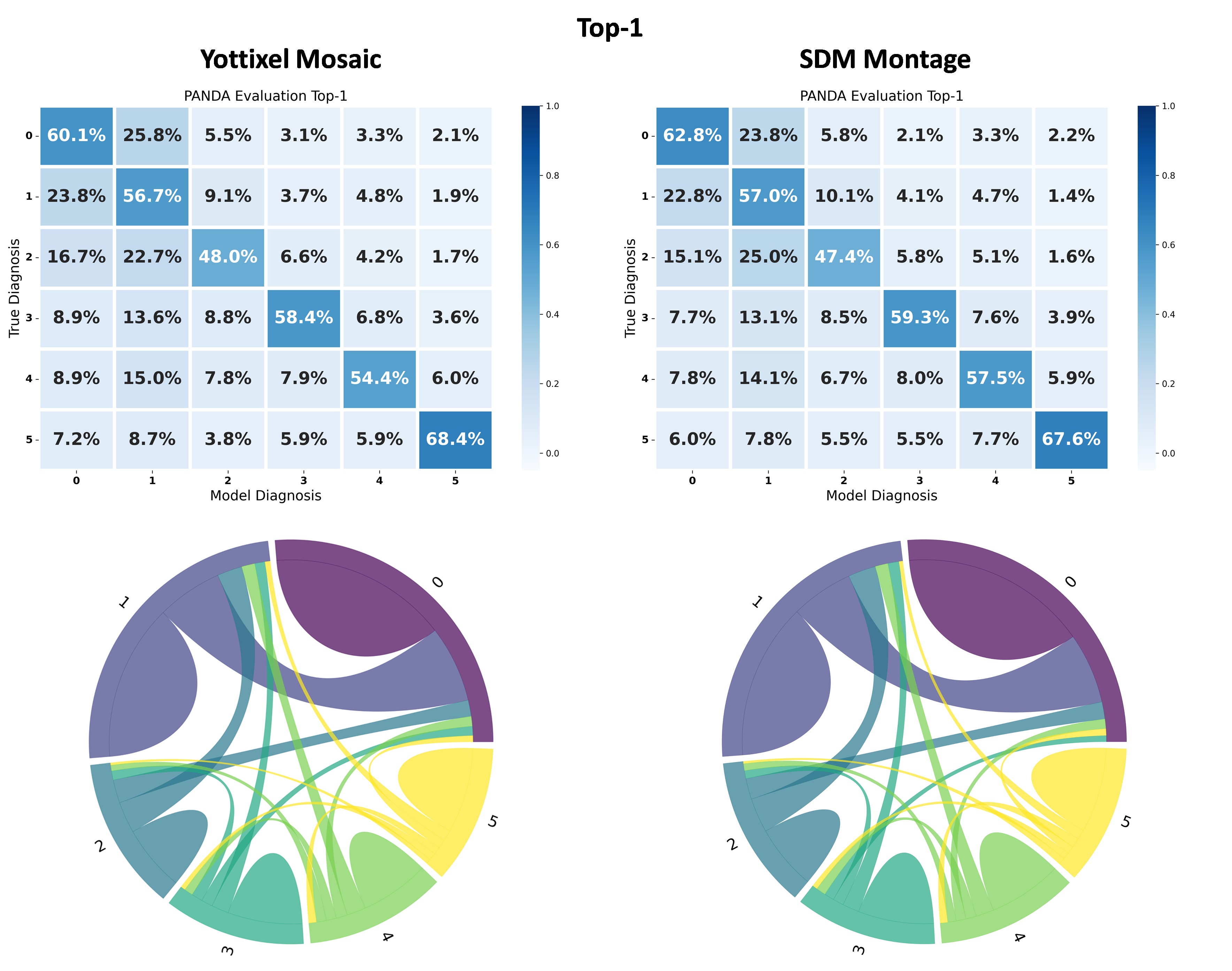}}
\caption{Confusion matrices and chord diagrams from Yottixel mosaic (left column), and SDM montage (right column). The evaluations are based on the top 1 retrieval when evaluating the PANDA dataset.}
\label{fig:PANDA_Chord1}
\end{figure*}

\begin{figure*}[t]
\centerline{\includegraphics[width =  1\textwidth]{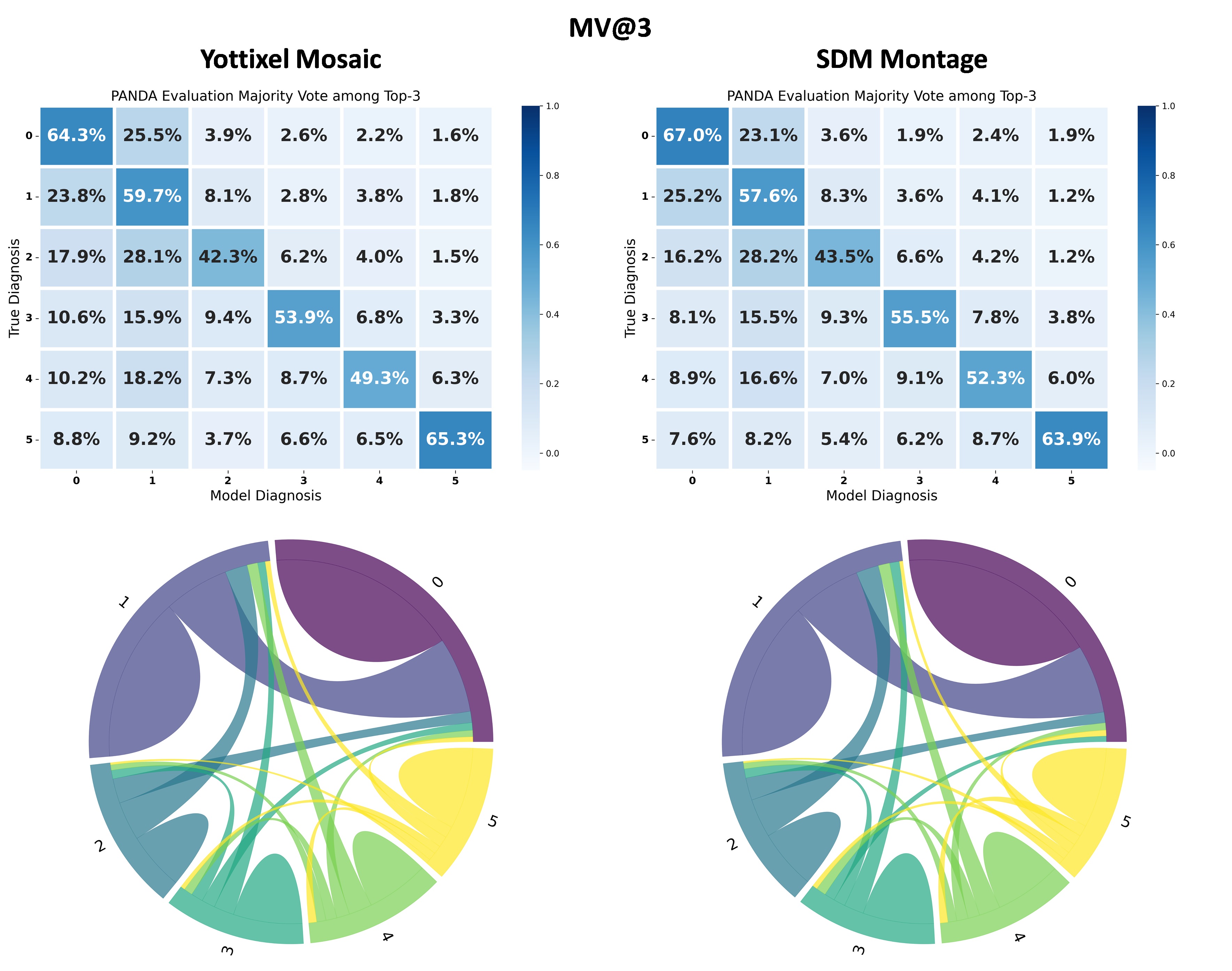}}
\caption{Confusion matrices and chord diagrams from Yottixel mosaic (left column), and SDM montage (right column). The evaluations are based on the majority of the top 3 retrievals when evaluating the PANDA dataset.}
\label{fig:PANDA_Chord3}
\end{figure*}

\begin{figure*}[t]
\centerline{\includegraphics[width =  1\textwidth]{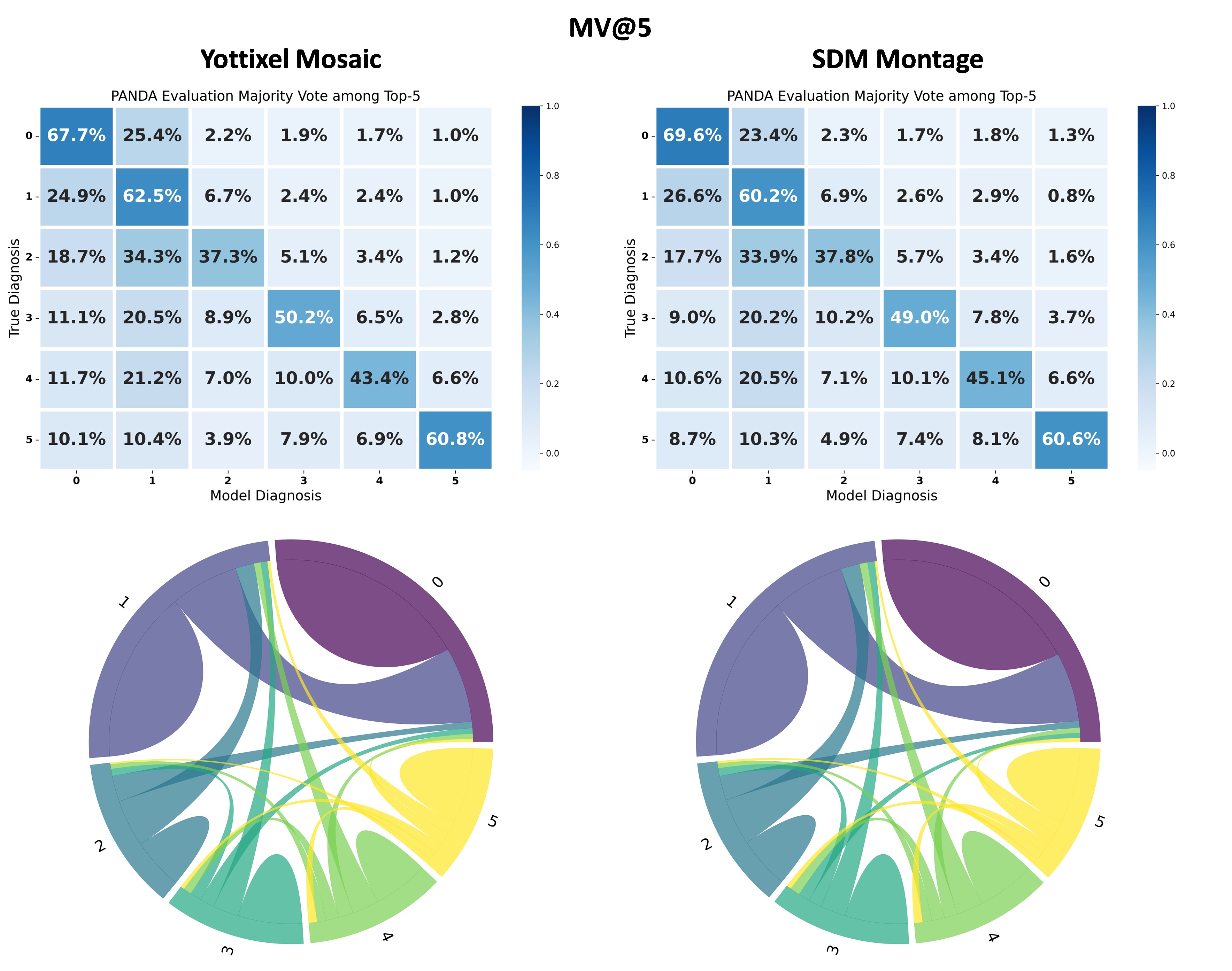}}
\caption{Confusion matrices and chord diagrams from Yottixel mosaic (left column), and SDM montage (right column). The evaluations are based on the majority of the top 5 retrievals when evaluating the PANDA dataset.}
\label{fig:PANDA_Chord5}
\end{figure*}

\begin{figure*}[t]
\centerline{\includegraphics[width =  1\textwidth]{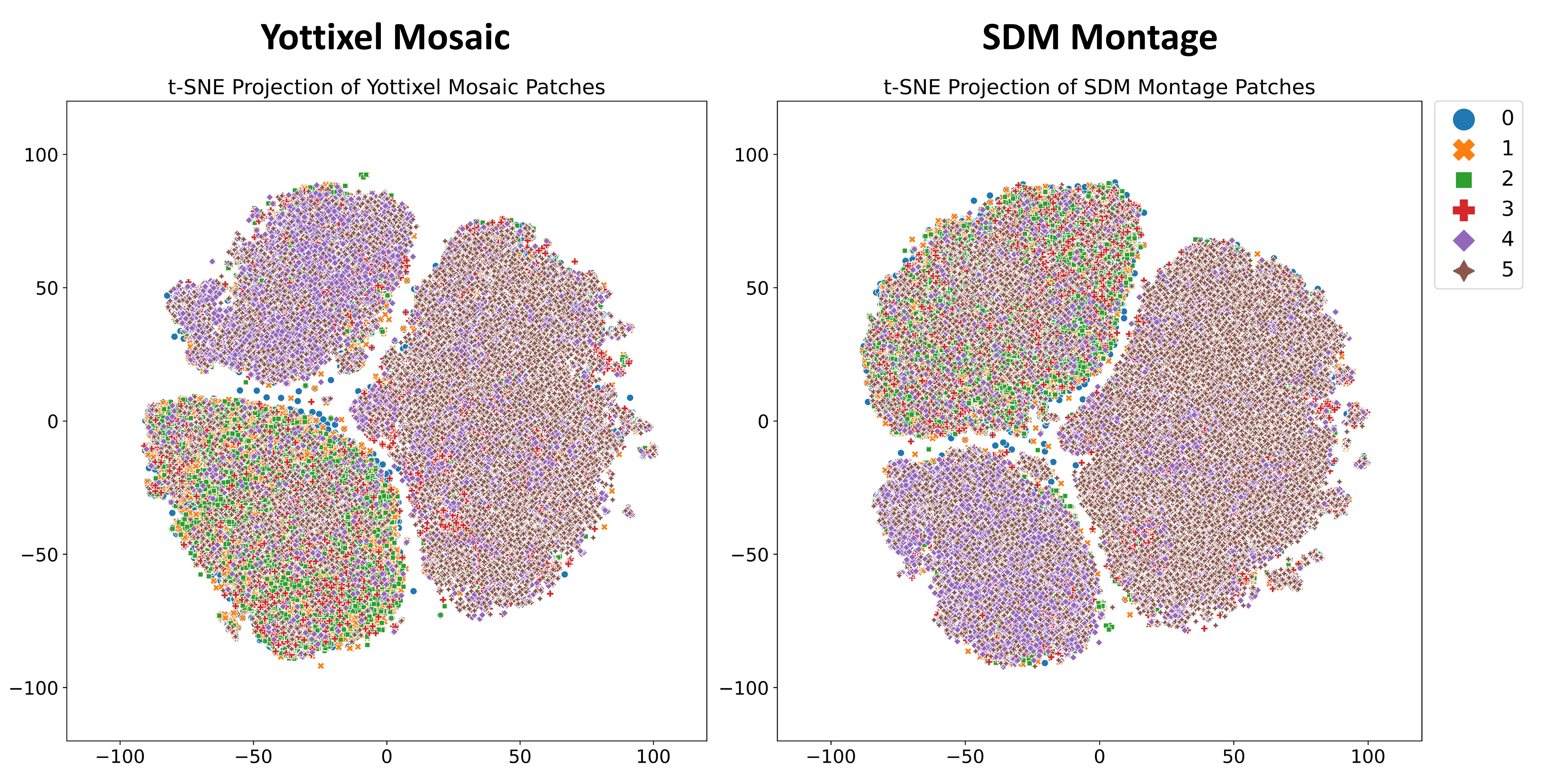}}
\caption{The t-SNE projection displays the embeddings of all patches extracted from the PANDA dataset using Yottixel's mosaic (left) and SDM's montage (right).}
\label{fig:PANDA_TSNE}
\end{figure*}

To assess the performance of the SDM montage in comparison to Yottixel's mosaic, we conducted a leave-one-out evaluation to retrieve the most similar cases. This evaluation involves multiple criteria for retrieval assessment, including the top-1, MV@3, and MV@5. The results include accuracy, macro average, and weighted average scores for each of these criteria, as depicted in Figure~\ref{fig:PANDA_Accuracy}. Table~\ref{tab:PANDA_results} shows the detailed statistical results including precision, recall, and f1-score. Moreover, confusion matrices and chord diagrams at Top-1, MV@3, and MV@5 are shown in Figure.~\ref{fig:PANDA_Chord1}, \ref{fig:PANDA_Chord3}, and ~\ref{fig:PANDA_Chord5}, respectively. In addition to these accuracy metrics, a comparative analysis of the number of patches extracted per WSI by each respective method is also presented in Figure~\ref{fig:PANDA_Patch_boxplot} for a visual representation of the distribution over the entire dataset. To visually illustrate the extracted patches, we used t-SNE projections, as demonstrated in Figure~\ref{fig:PANDA_TSNE}.

\begin{table*}[t]
\centering
\resizebox{\textwidth}{!}{\begin{tabular}{l|l|lll|lll|lll|l}
\hline
                                 &                                                             & \multicolumn{3}{c|}{\textbf{Top-1}}    & \multicolumn{3}{c|}{\textbf{MV@3}}     & \multicolumn{3}{c|}{\textbf{MV@5}}    &        \\ \hline
                                 & \begin{tabular}[c]{@{}l@{}}ISUP Grade\end{tabular} & Precision & Recall & f1-score & Precision & Recall & f1-score & Precision & Recall & f1-score & Slides \\ \hline
\multirow{6}{*}{\rotatebox[origin=c]{0}{\textbf{\begin{tabular}[c]{@{}c@{}}Yottixel\\Mosaic~\cite{kalra2020Yottixel}\end{tabular}}}} 
& 0  & 0.60   & 0.60  & 0.60  & 0.60  & 0.64   & 0.62  & 0.60   & 0.68  & 0.63  & 2853     \\
& 1  & 0.50   & 0.57  & 0.53  & 0.49  & 0.60   & 0.54  & 0.48   & 0.62  & 0.54  & 2655     \\
& 2  & 0.50   & 0.48  & 0.49  & 0.49  & 0.42   & 0.46  & 0.51   & 0.37  & 0.43  & 1332     \\
& 3  & 0.62   & 0.58  & 0.60  & 0.62  & 0.54   & 0.57  & 0.60   & 0.50  & 0.55  & 1230     \\
& 4  & 0.61   & 0.54  & 0.57  & 0.62  & 0.49   & 0.55  & 0.62   & 0.43  & 0.51  & 1225     \\
& 5  & 0.77   & 0.68  & 0.72  & 0.77  & 0.65   & 0.71  & 0.80   & 0.61  & 0.69  & 1201    \\ \hline
 &\textbf{Total Slides}&&&&&&&&&&  \textbf{10496}\\ \hline
\multirow{6}{*}{\rotatebox[origin=c]{0}{\textbf{\begin{tabular}[c]{@{}c@{}}SDM\\Montage\end{tabular}}}}  
& 0  & 0.63   & 0.63  & 0.63  & 0.62  & 0.67   & 0.64  & 0.61   & 0.70  & 0.65  & 2889     \\
& 1  & 0.51   & 0.57  & 0.54  & 0.50  & 0.58   & 0.53  & 0.48   & 0.60  & 0.53  & 2665     \\
& 2  & 0.48   & 0.47  & 0.48  & 0.50  & 0.43   & 0.46  & 0.49   & 0.38  & 0.43  & 1343     \\
& 3  & 0.64   & 0.59  & 0.62  & 0.62  & 0.55   & 0.58  & 0.60   & 0.49  & 0.54  & 1242     \\
& 4  & 0.60   & 0.58  & 0.59  & 0.60  & 0.52   & 0.56  & 0.60   & 0.45  & 0.52  & 1246     \\
& 5  & 0.77   & 0.68  & 0.72  & 0.77  & 0.64   & 0.70  & 0.78   & 0.61  & 0.68  & 1223    \\ \hline
&\textbf{Total Slides}&&&&&&&&&&  \textbf{10608}\\ \hline
\end{tabular}}
\caption{Precision, recall, f1-score, and the number of slides processed for each sub-type are shown in this table using Yottixel mosaic, and SDM montage. The evaluations are based on the top 1 retrieval, the majority among the top 3 retrievals, and the majority among the top 5 retrievals in the PANDA dataset.}\label{tab:PANDA_results}
\end{table*}

PANDA is one of the most extensive publicly available datasets for prostate cancer analysis. In this research, our empirical findings shed light on the comparative efficacy of our proposed method when compared to the Yottixel mosaic. Specifically, our findings indicate that SDM exhibited comparable performance to the Yottixel mosaic concerning accuracy with majority agreement among the top 5 retrievals. However, a noteworthy distinction emerged when considering accuracy at top-1 and the majority agreement among the top 3 retrievals. Regarding the macro-averaged F1-scores, both top-1 and MV@5 exhibit analogous outcomes. However, for MV@3, the SDM method demonstrates a 1\% enhancement, as depicted in the Figure~\ref{fig:PANDA_Accuracy}. This highlights the proficiency of the SDM method in assimilating pertinent information for retrieval tasks without the reliance on empirical parameters, a contrast to the Yottixel approach. Specifically, Yottixel necessitates predefined settings for both cluster count and patch selection percentage. Moreover, our analysis revealed an intriguing facet of Yottixel's behavior in comparison to SDM. Specifically, it has come to our attention that Yottixel exhibits a tendency to overlook certain WSIs within the dataset. Our observations indicate that Yottixel processed a total of 10,496 WSIs, while SDM demonstrated a more comprehensive approach, successfully processing 10,608 WSIs out of the 10,616 WSIs as shown in Table~\ref{tab:PANDA_results}. This observation underscores the robustness and completeness of the SDM method in managing the entire dataset, further emphasizing its advantages in applications related to the analysis and retrieval of WSIs in the context of prostate cancer research. A notable inference from the box plot depicted in Figure~\ref{fig:PANDA_Patch_boxplot}reveals that for fine needle biopsies (which constitute a significant portion of the PANDA dataset), the Yottixel 5\% methodology selects a reduced number of patches in comparison to the SDM approach.

\clearpage\subsection{Private -- Colorectal Cancer (CRC)}
\label{sec:Exd_Results_CRC}
The Colorectal Cancer (CRC) dataset, sourced from our hospital, encompasses a collection of 209 WSIs, with a primary focus on colorectal histopathology. This dataset is categorized into three distinct groups, specifically Cancer Adjacent polyps (CAP), Non-recurrent polyps (POP-NR), and Recurrent polyps (POP-R), all of which pertain to colorectal pathology. Importantly, all the slides in this dataset were subjected to scanning at a magnification level of 40x (see Table~\ref{tab:CRC} for more details).

\begin{table}[t] 
\centering
\begin{tabular}{llr}
\hline
\begin{tabular}[c]{@{}l@{}}Primary Diagnoses\end{tabular} & Acronyms & Slides \\ \hline
Cancer Adjacent Polyps              & CAP     & 63     \\
Non-recurrent Polyps              & POP-NR    & 63     \\
Recurrent Polyps                  & POP-R     & 83     \\ \hline
\end{tabular}
\caption{Comprehensive dataset particulars pertaining to the Colorectal Cancer dataset utilized in this experiment, encompassing relevant acronyms and the number of slides attributed to each primary diagnosis.}\label{tab:CRC}
\end{table}

\begin{figure*}[t]
\centerline{\includegraphics[width =  1\textwidth]{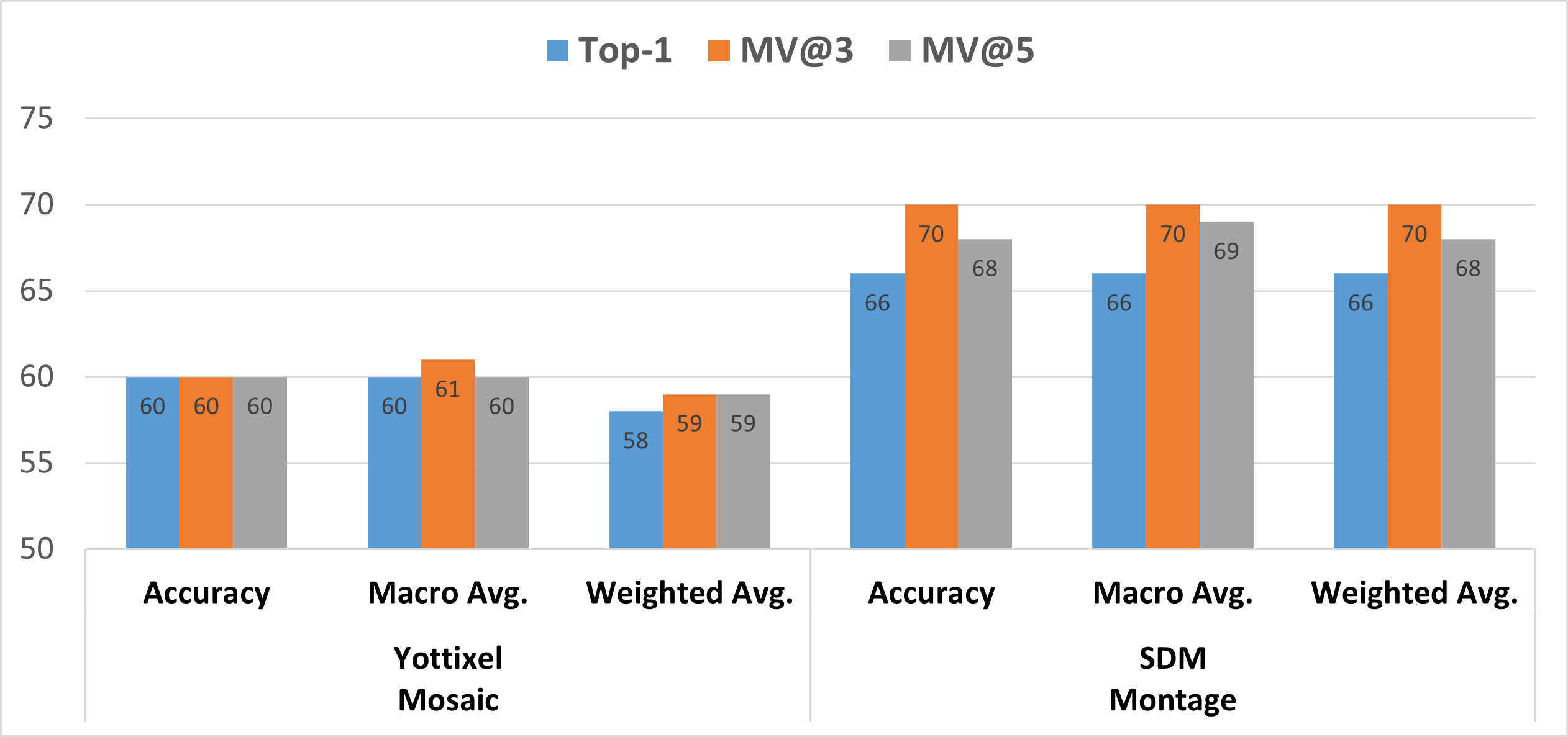}}
\caption{Accuracy, macro average of f1-scores, and weighted average of f1-scores are shown from Yottixel mosaic, and SDM montage. The evaluations are based on the top 1 retrieval, the majority among the top 3 retrievals, and the majority among the top 5 retrievals using the CRC dataset.}
\label{fig:CRC_Accuracy}
\end{figure*}

\begin{figure*}[t]
\centerline{\includegraphics[width =  0.6\textwidth]{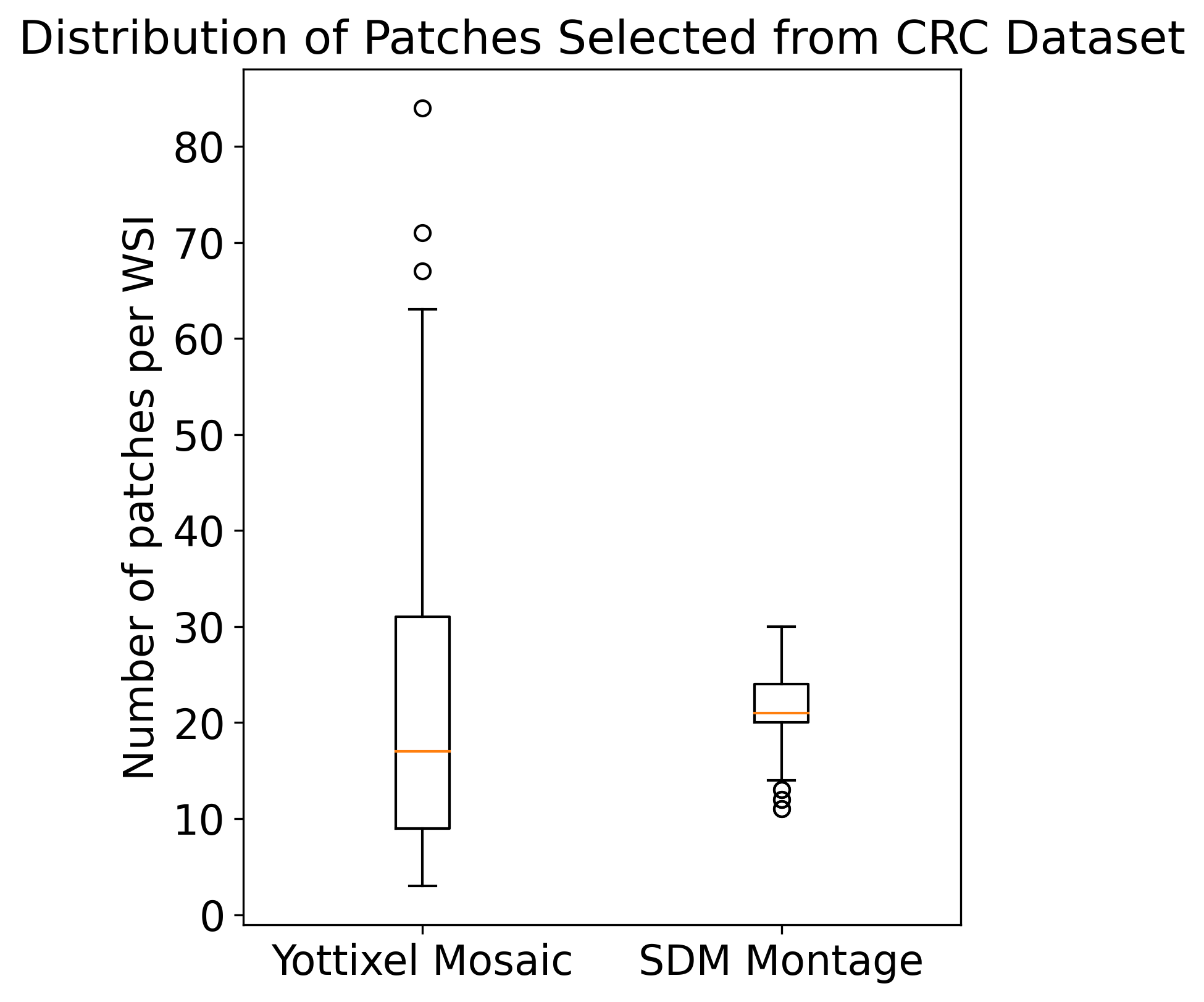}}
\caption{The boxplot illustrates the distribution of patches selected for each WSI in the CRC dataset from both the Yottixel Mosaic and SDM Montage. Additionally, it provides statistical measures for these distributions. Specifically, for the Yottixel Mosaic, the median number of selected patches is $17\pm15$. On the other hand, for the SDM Montage, the median number of selected patches is $21\pm4$.}
\label{fig:CRC_Patch_boxplot}
\end{figure*}

\begin{figure*}[t]
\centerline{\includegraphics[width =  1\textwidth]{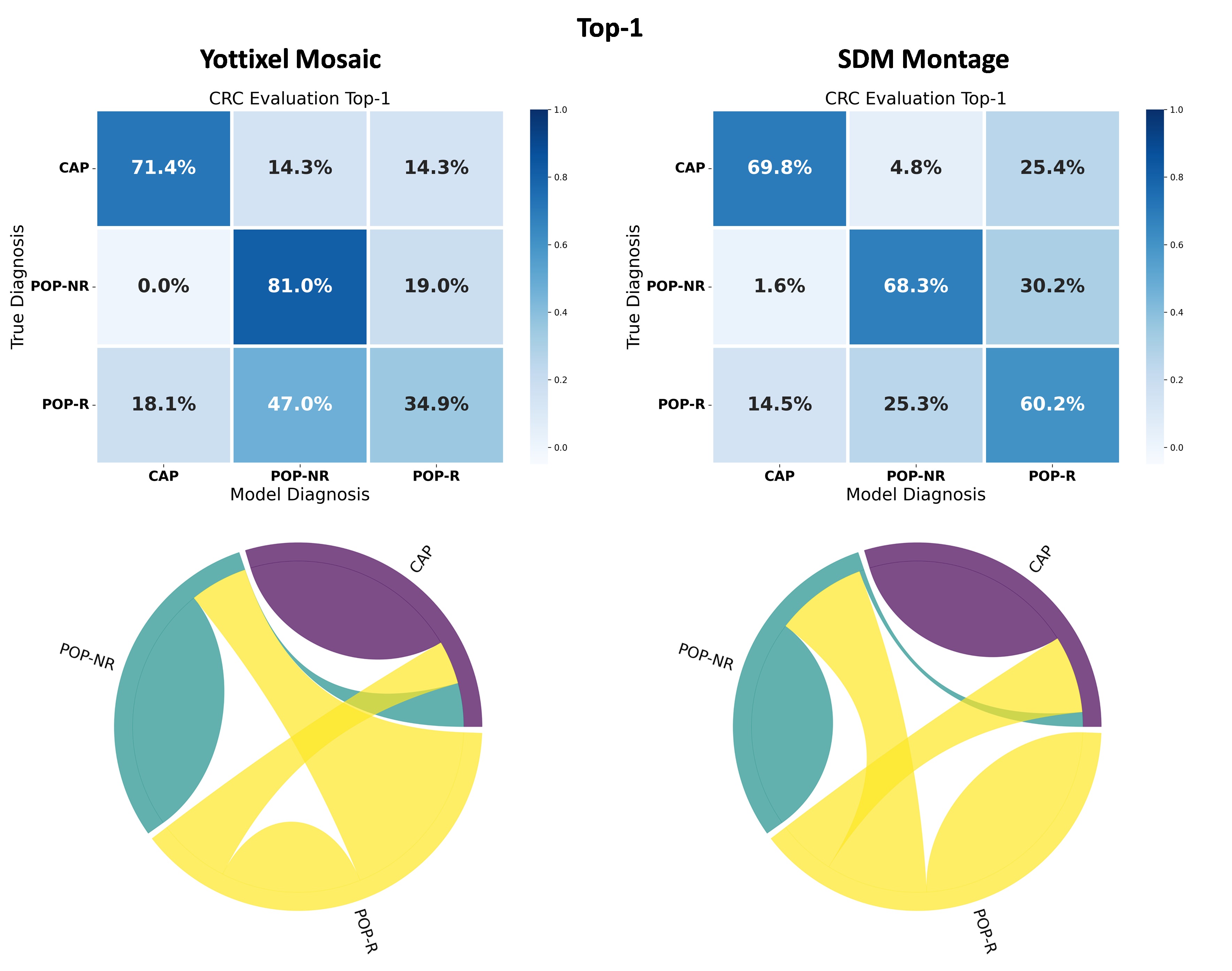}}
\caption{Confusion matrices and chord diagrams from Yottixel mosaic (left column), and SDM montage (right column). The evaluations are based on the top 1 retrieval when evaluating the CRC dataset.}
\label{fig:CRC_Chord1}
\end{figure*}

\begin{figure*}[t]
\centerline{\includegraphics[width =  1\textwidth]{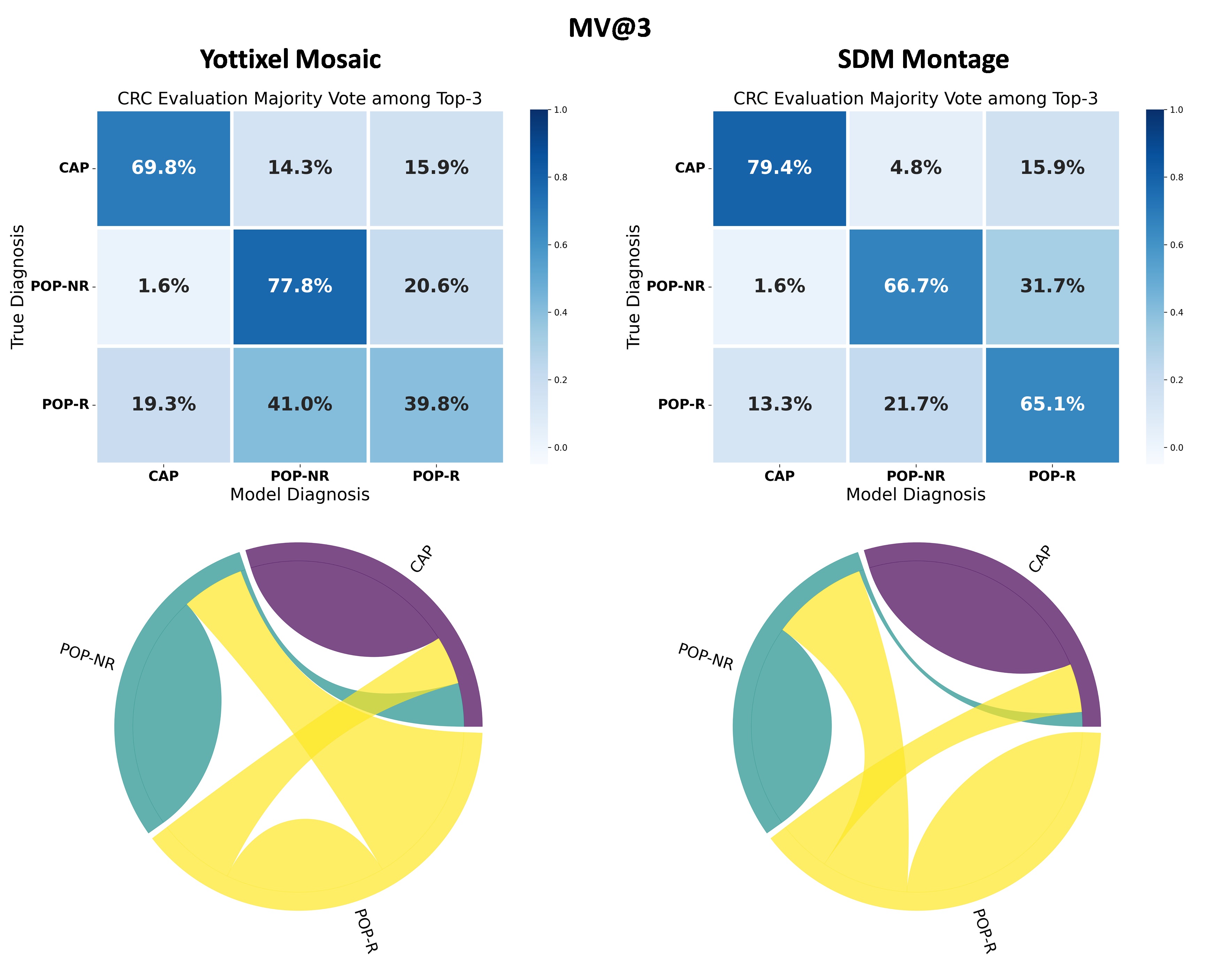}}
\caption{Confusion matrices and chord diagrams from Yottixel mosaic (left column), and SDM montage (right column). The evaluations are based on the majority of the top 3 retrievals when evaluating the CRC dataset.}
\label{fig:CRC_Chord3}
\end{figure*}

\begin{figure*}[t]
\centerline{\includegraphics[width =  1\textwidth]{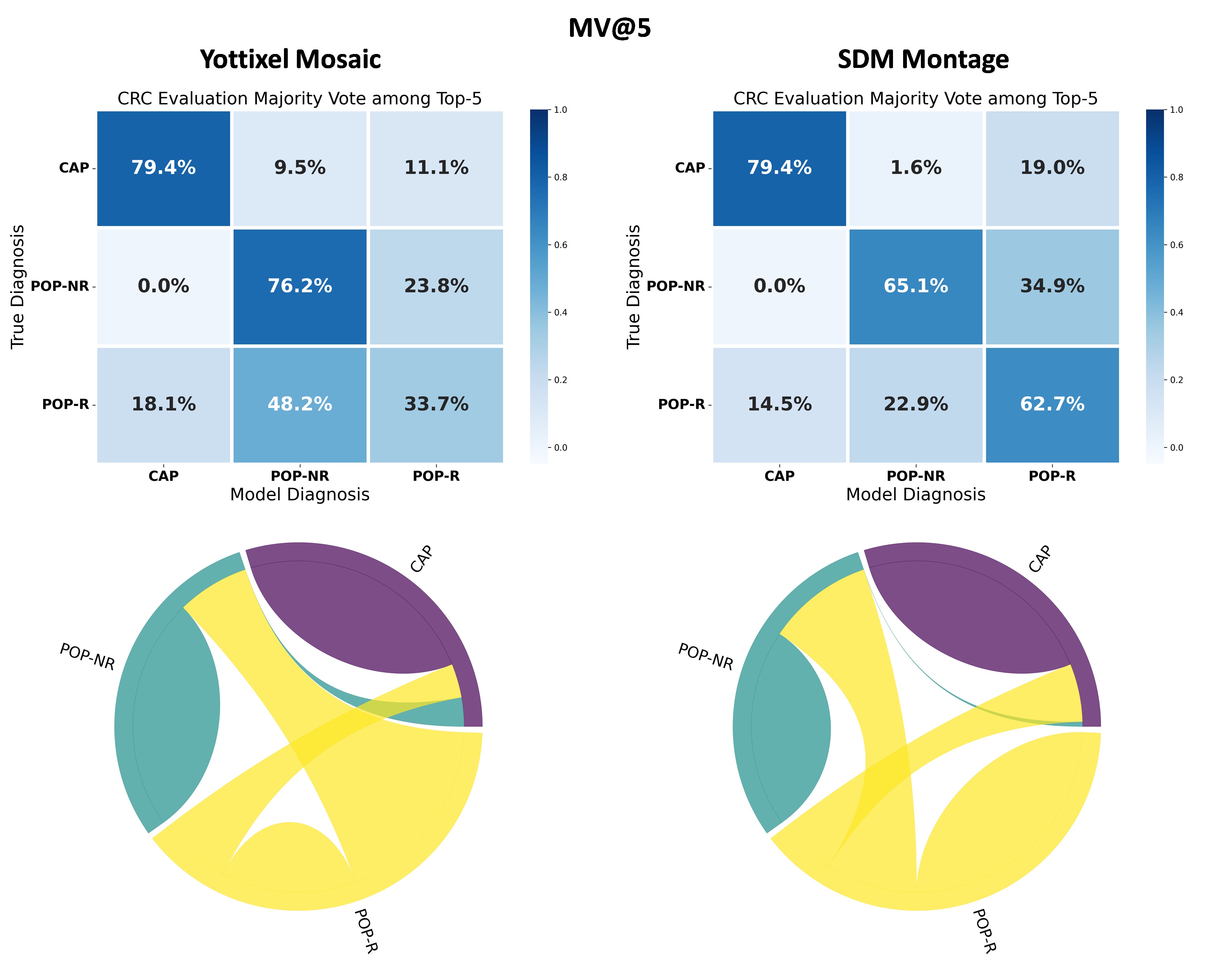}}
\caption{Confusion matrices and chord diagrams from Yottixel mosaic (left column), and SDM montage (right column). The evaluations are based on the majority of the top 5 retrievals when evaluating the CRC dataset.}
\label{fig:CRC_Chord5}
\end{figure*}

\begin{figure*}[t]
\centerline{\includegraphics[width =  1\textwidth]{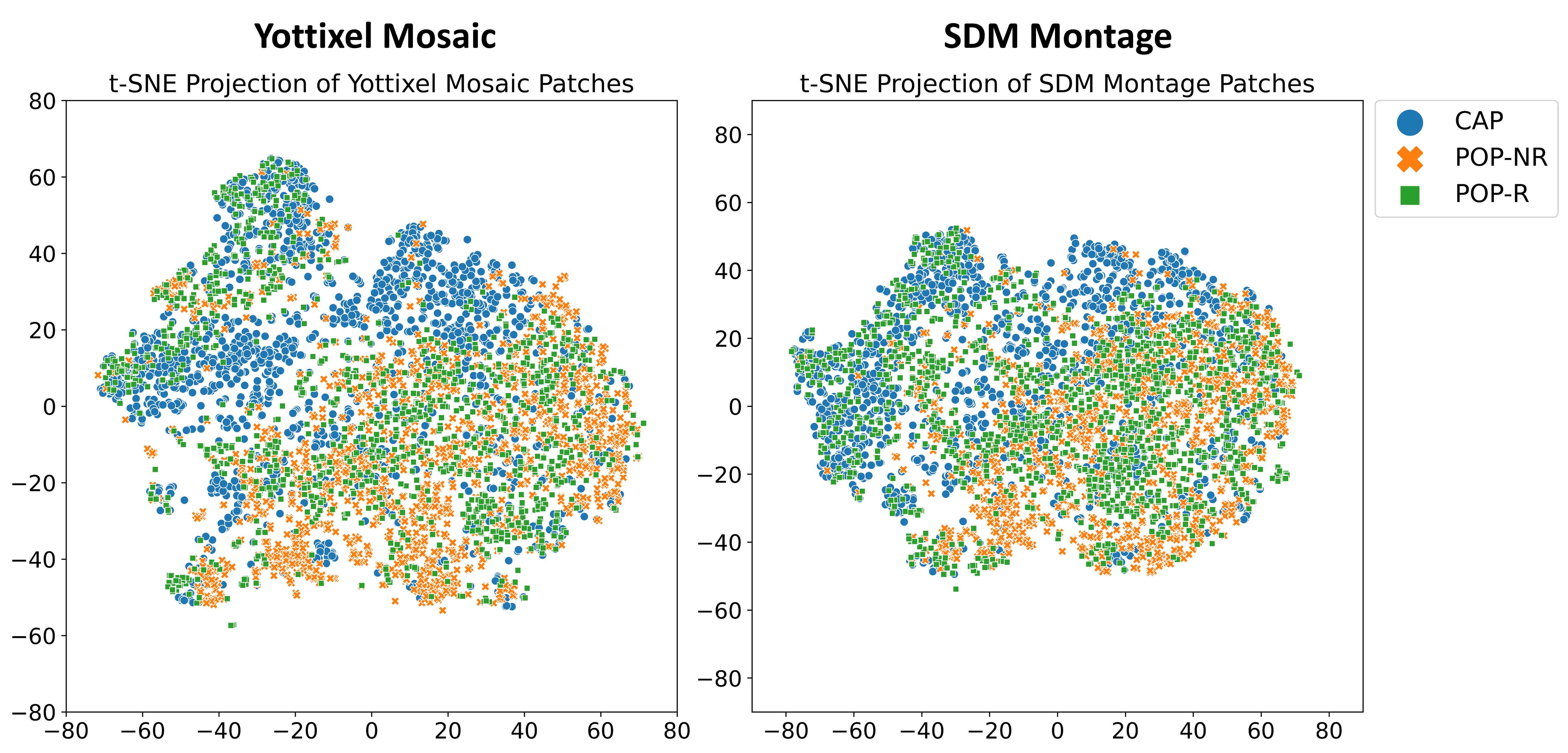}}
\caption{The t-SNE projection displays the embeddings of all patches extracted from the CRC dataset using Yottixel's mosaic (left) and SDM's montage (right).}
\label{fig:CRC_TSNE}
\end{figure*}

To assess the effectiveness of the SDM montage in comparison to Yottixel's mosaic, we conducted a leave-one-out evaluation to retrieve the most similar cases using the CRC dataset. The evaluation criteria encompass multiple retrieval scenarios, including the top-1, MV@3, and MV@5. The results, including accuracy, macro average, and weighted average scores at the top-1, MV@3, and MV@5 levels, are presented in Figure~\ref{fig:CRC_Accuracy}. Table~\ref{tab:CRC_results} shows the detailed statistical results including precision, recall, and f1-score. Moreover, confusion matrices and chord diagrams at Top-1, MV@3, and MV@5 retrievals are shown in Figure~\ref{fig:CRC_Chord1}, \ref{fig:CRC_Chord3}, and ~\ref{fig:CRC_Chord5}, respectively. In addition to the traditional accuracy metrics, we conducted a comparative examination of the number of patches extracted per WSI by each individual method. For a visual depiction of this distribution across the complete dataset, we refer to the boxplots provided in Figure~\ref{fig:CRC_Patch_boxplot}. To visually illustrate the extracted patches, we used t-SNE projections, as demonstrated in Figure~\ref{fig:CRC_TSNE}.

\begin{table*}[t]
\centering
\resizebox{\textwidth}{!}{\begin{tabular}{l|l|lll|lll|lll|c}
\hline
&  & \multicolumn{3}{c|}{\textbf{Top-1}}    & \multicolumn{3}{c|}{\textbf{MV@3}}     & \multicolumn{3}{c|}{\textbf{MV@5}} &  \\ \hline
& \begin{tabular}[c]{@{}l@{}}Primary\\ Diagnoses\end{tabular} & Precision & Recall & f1-score & Precision & Recall & f1-score & Precision & Recall & f1-score & Slides \\ \hline
\multirow{3}{*}{\rotatebox[origin=c]{0}{\textbf{\begin{tabular}[c]{@{}c@{}}Yottixel\\ Mosaic~\cite{kalra2020Yottixel}\end{tabular}}}} 
& CAP  & 0.75      & 0.71   & 0.73     & 0.72      & 0.70   & 0.71     & 0.77      & 0.79   & 0.78     & 63    \\
& POP-NR  & 0.52      & 0.81   & 0.63     & 0.53      & 0.78   & 0.63     & 0.51      & 0.76   & 0.61     & 63         \\
& POP-R   & 0.58      & 0.35   & 0.44     & 0.59      & 0.40   & 0.47     & 0.56      & 0.34   & 0.42     & 83        \\ \hline
&\textbf{Total Slides}&&&&&&&&&&  \textbf{209}\\ \hline
\multirow{3}{*}{\rotatebox[origin=c]{0}{\textbf{\begin{tabular}[c]{@{}c@{}}SDM\\ Montage\end{tabular}}}}  
& CAP     & 0.77  & 0.70 &  0.73   & 0.81  & 0.79 & 0.80   &   0.81     &  0.79   & 0.80 &   63    \\
& POP-NR  & 0.64  & 0.68 &  0.66  & 0.67  & 0.67 &  0.67      & 0.67    &  0.65   &  0.66 &  63      \\ 
& POP-R  & 0.59   & 0.60 &  0.60  & 0.64   & 0.65 &  0.65      & 0.60   &  0.63    &  0.62 &   83     \\ \hline
&\textbf{Total Slides}&&&&&&&&&&  \textbf{209}\\ \hline
\end{tabular}}
\caption{Precision, recall, f1-score, and the number of slides processed for each sub-type are shown in this table using Yottixel mosaic, and SDM montage. The evaluations are based on the top 1 retrieval, the majority among the top 3 retrievals, and the majority among the top 5 retrievals using the CRC dataset.}\label{tab:CRC_results}
\end{table*}

During our experimentation, the SDM montage manifested a marked performance superiority over the Yottixel mosaic. Specifically, we observed enhancements in the macro-average of F1-scores by +6\%, +9\%, and +9\% for top-1, MV@3, and MV@5 retrievals, respectively. From an accuracy perspective, the SDM method demonstrated increments of +6\%, +10\%, and +8\% for the top-1, MV@3, and MV@5 retrievals, respectively. These results emphasize the SDM method's adeptness in assimilating and representing critical data effectively within the retrieval paradigm, as delineated in the referenced Figure~\ref{fig:CRC_Accuracy}. Furthermore, an additional noteworthy benefit of implementing the SDM montage method comes to the forefront when examining Figure~\ref{fig:CRC_Patch_boxplot}, which depicts the number of selected patches. In contrast to the Yottixel mosaic, SDM proves to be more resource-efficient by opting for a smaller patch selection. This not only leads to storage conservation but also eliminates the redundancy and the necessity for an empirical determination of the optimal patch count to select.

\clearpage\subsection{Private -- Alcoholic Steatohepatitis \& Non-alcoholic Steatohepatitis (ASH \& NASH) of Liver}
\label{sec:Exd_Results_Liver}
Liver biopsy slides were acquired from patients who had been diagnosed with either Alcoholic Steatohepatitis (ASH) or Non-Alcoholic Steatohepatitis (NASH) at our hospital. The ASH diagnosis was established through a comprehensive review of patient records and expert assessments that considered medical history, clinical presentation, and laboratory findings. For the NASH group, liver biopsies were selected from a cohort of morbidly obese patients undergoing bariatric surgery. All of the biopsy slides were digitized at $40\times$ magnification and linked to their respective diagnoses at the WSI level (see Table~\ref{tab:liver} for more details).

\begin{table}[t] 
\centering
\begin{tabular}{llr}
\hline
\begin{tabular}[c]{@{}l@{}}Primary Diagnoses\end{tabular} & Acronyms & Slides \\ \hline
Alcoholic Steatohepatitis                                 & ASH     & 150     \\
Non-alcoholic Steatohepatitis                              & NASH    & 158     \\
Normal                                                      & Normal     & 18     \\ \hline
\end{tabular}
\caption{Information related to the Liver dataset, inclusive of the respective acronyms and the number of slides associated with each primary diagnosis.}\label{tab:liver}
\end{table}

\begin{figure*}[t]
\centerline{\includegraphics[width =  1\textwidth]{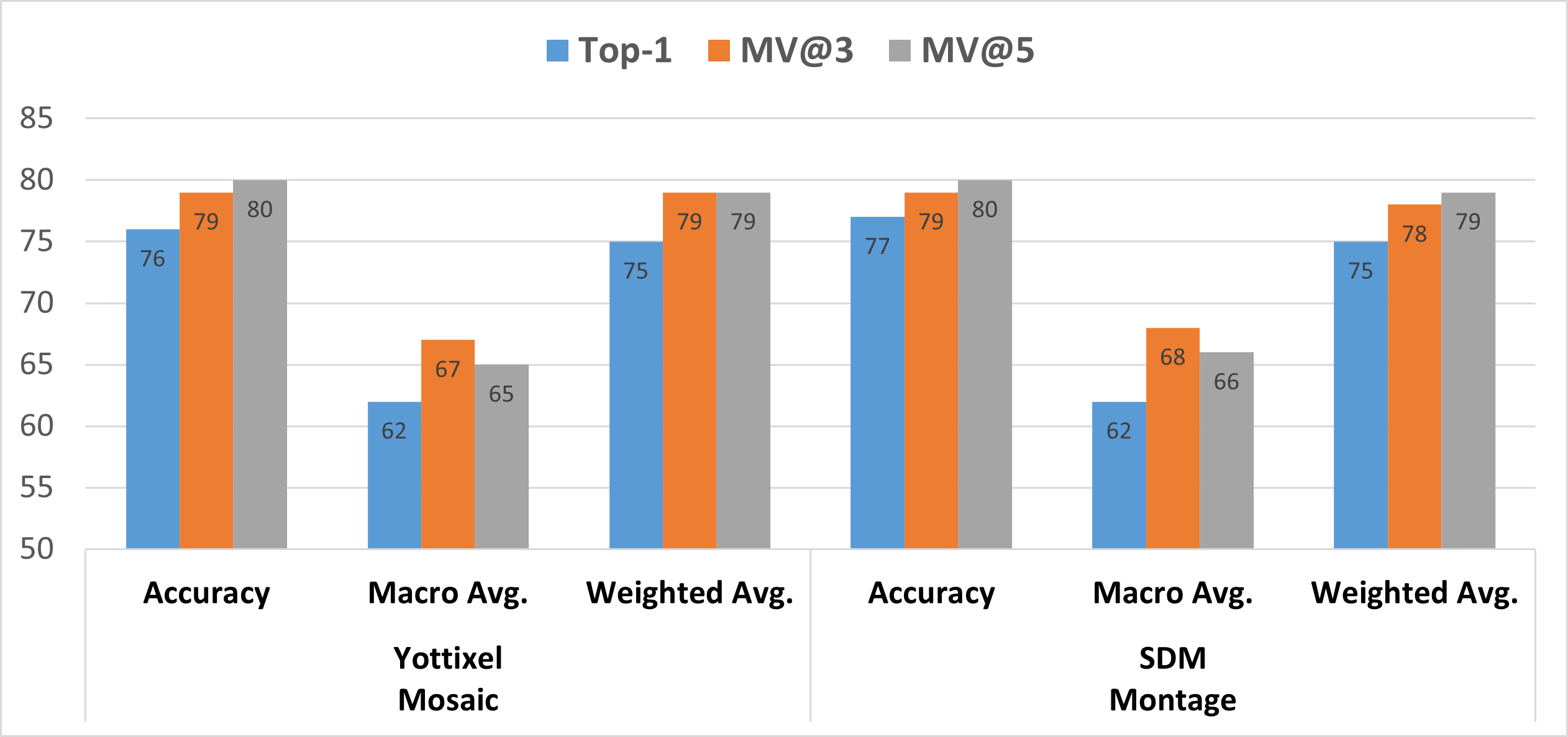}}
\caption{Accuracy, macro average of f1-scores, and weighted average of f1-scores are shown from Yottixel mosaic, and SDM montage. The evaluations are based on the top 1 retrieval, the majority among the top 3 retrievals, and the majority among the top 5 retrievals in the Liver dataset.}
\label{fig:Liver_Accuracy}
\end{figure*}

\begin{figure*}[t]
\centerline{\includegraphics[width =  0.6\textwidth]{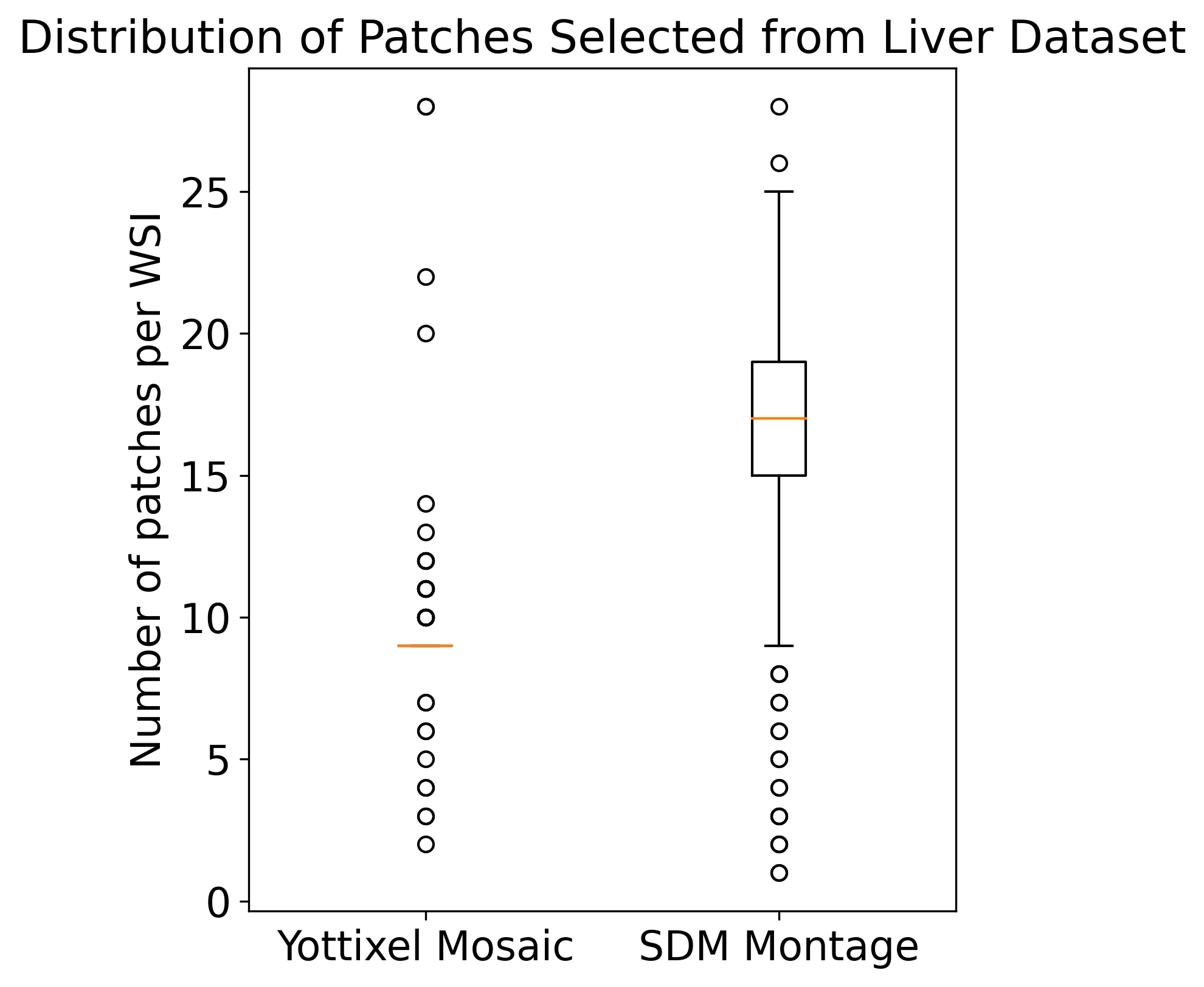}}
\caption{The boxplot illustrates the distribution of patches selected for each WSI in the Liver dataset from both the Yottixel Mosaic and SDM Montage. Additionally, it provides statistical measures for these distributions. Specifically, for the Yottixel Mosaic, the median number of selected patches is $9\pm3$. Conversely, for the SDM Montage, the median number of selected patches is $17\pm4$.}
\label{fig:Liver_Patch_boxplot}
\end{figure*}

\begin{figure*}[t]
\centerline{\includegraphics[width =  1\textwidth]{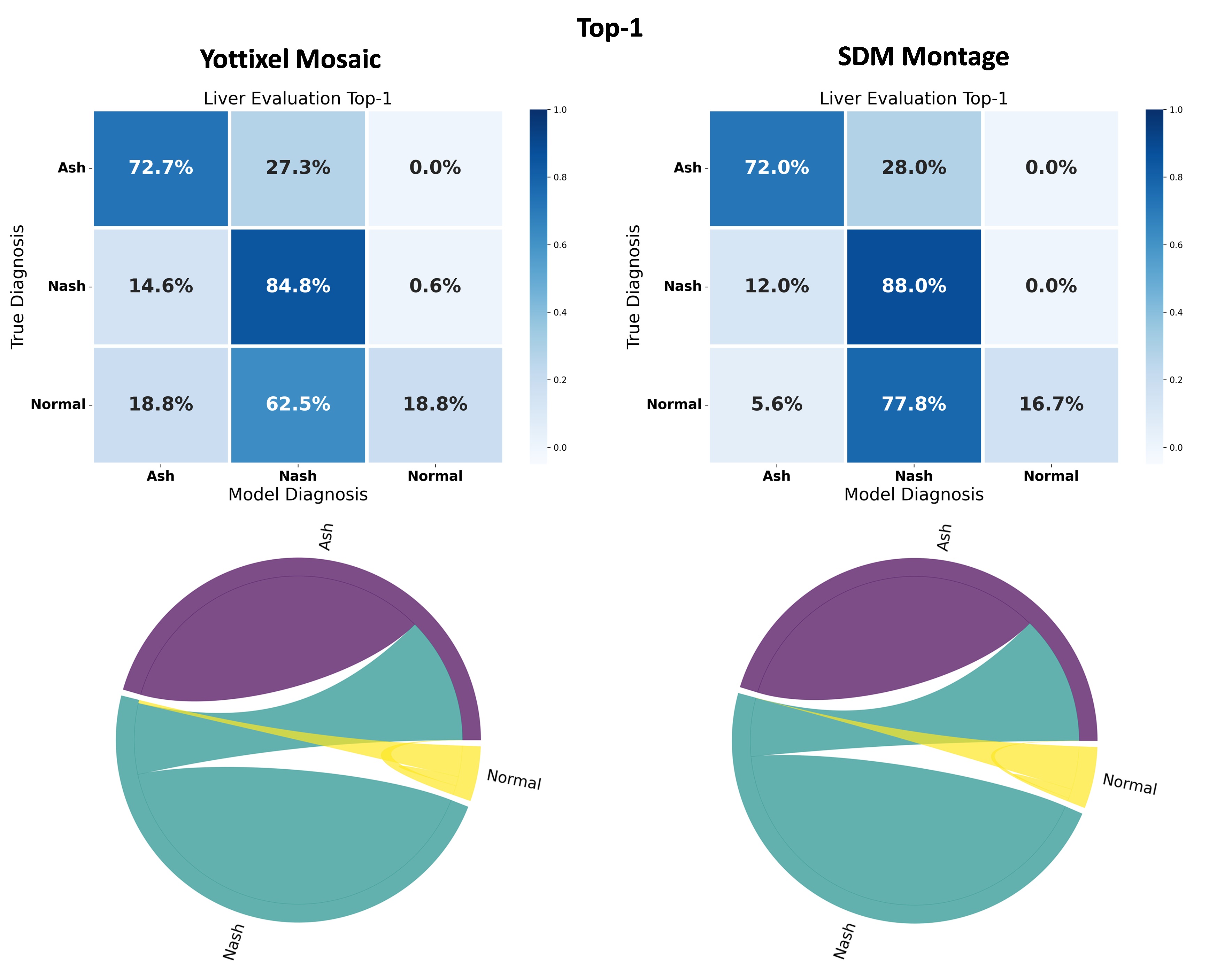}}
\caption{Confusion matrices and chord diagrams from Yottixel mosaic (left column), and SDM montage (right column). The evaluations are based on the top 1 retrieval when evaluating the Liver dataset.}
\label{fig:Liver_Chord1}
\end{figure*}

\begin{figure*}[t]
\centerline{\includegraphics[width =  1\textwidth]{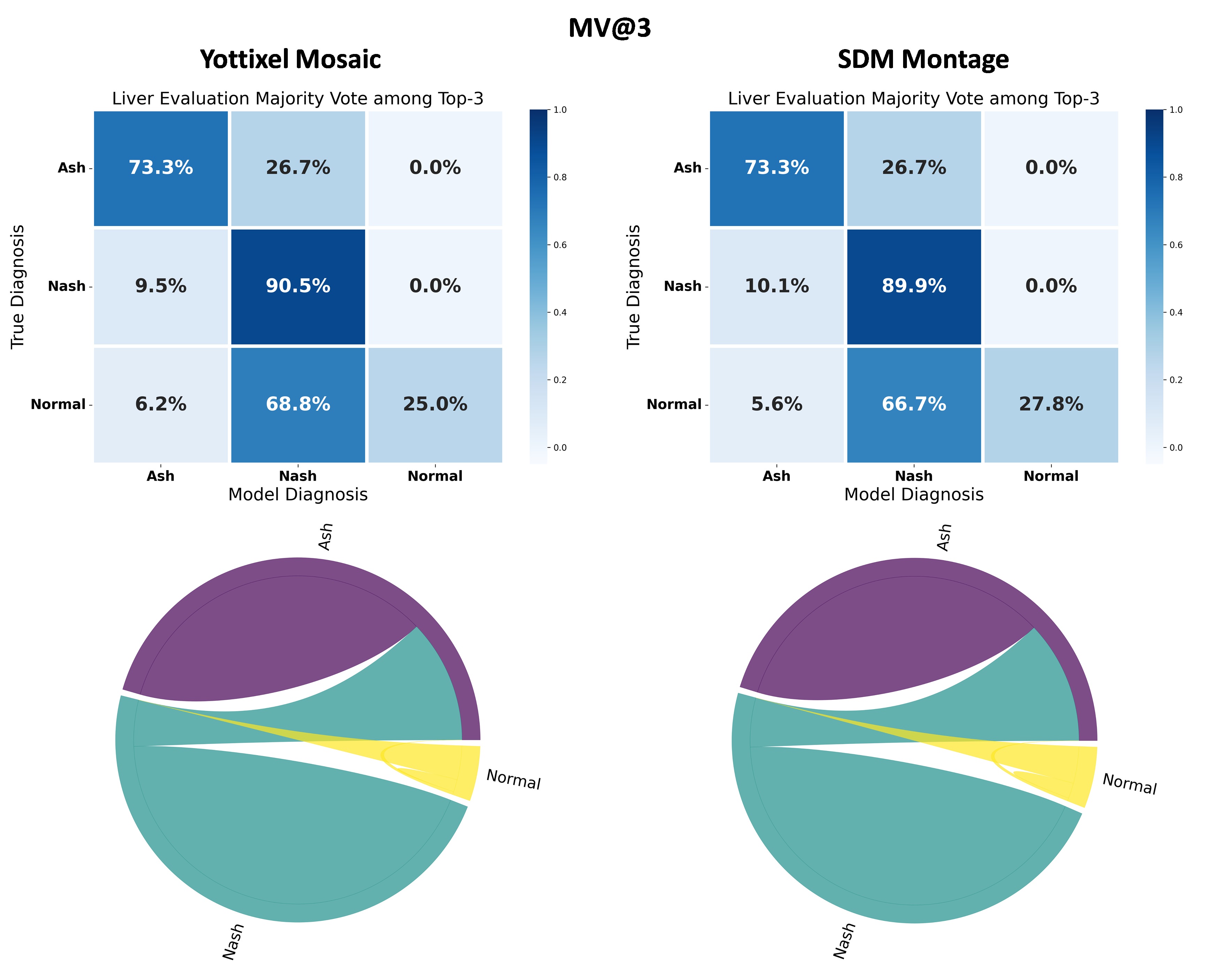}}
\caption{Confusion matrices and chord diagrams from Yottixel mosaic (left column), and SDM montage (right column). The evaluations are based on the majority of the top 3 retrievals when evaluating the Liver dataset.}
\label{fig:Liver_Chord3}
\end{figure*}

\begin{figure*}[t]
\centerline{\includegraphics[width =  1\textwidth]{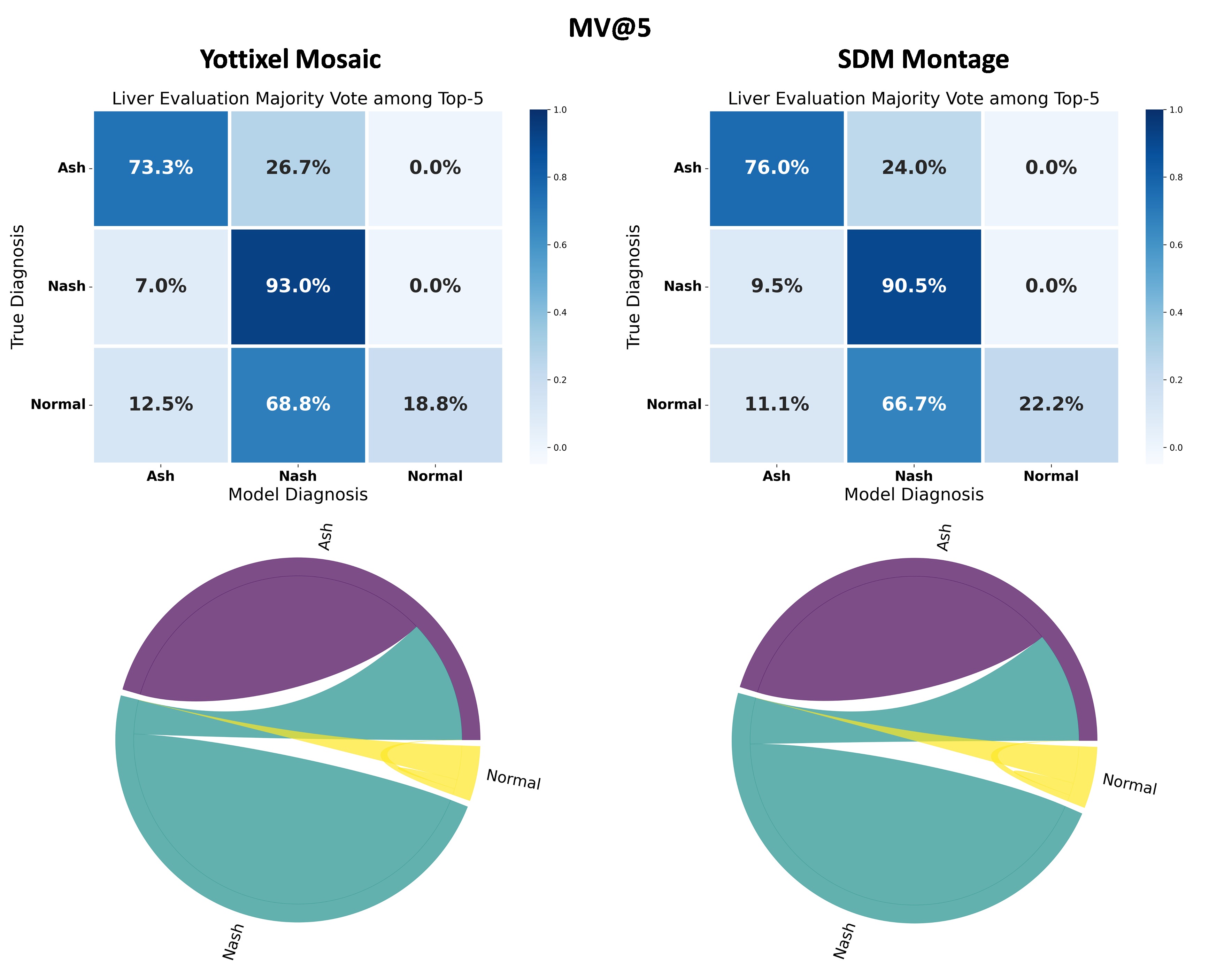}}
\caption{Confusion matrices and chord diagrams from Yottixel mosaic (left column), and SDM montage (right column). The evaluations are based on the majority of the top 5 retrievals when evaluating the Liver dataset.}
\label{fig:Liver_Chord5}
\end{figure*}

\begin{figure*}[t]
\centerline{\includegraphics[width =  1\textwidth]{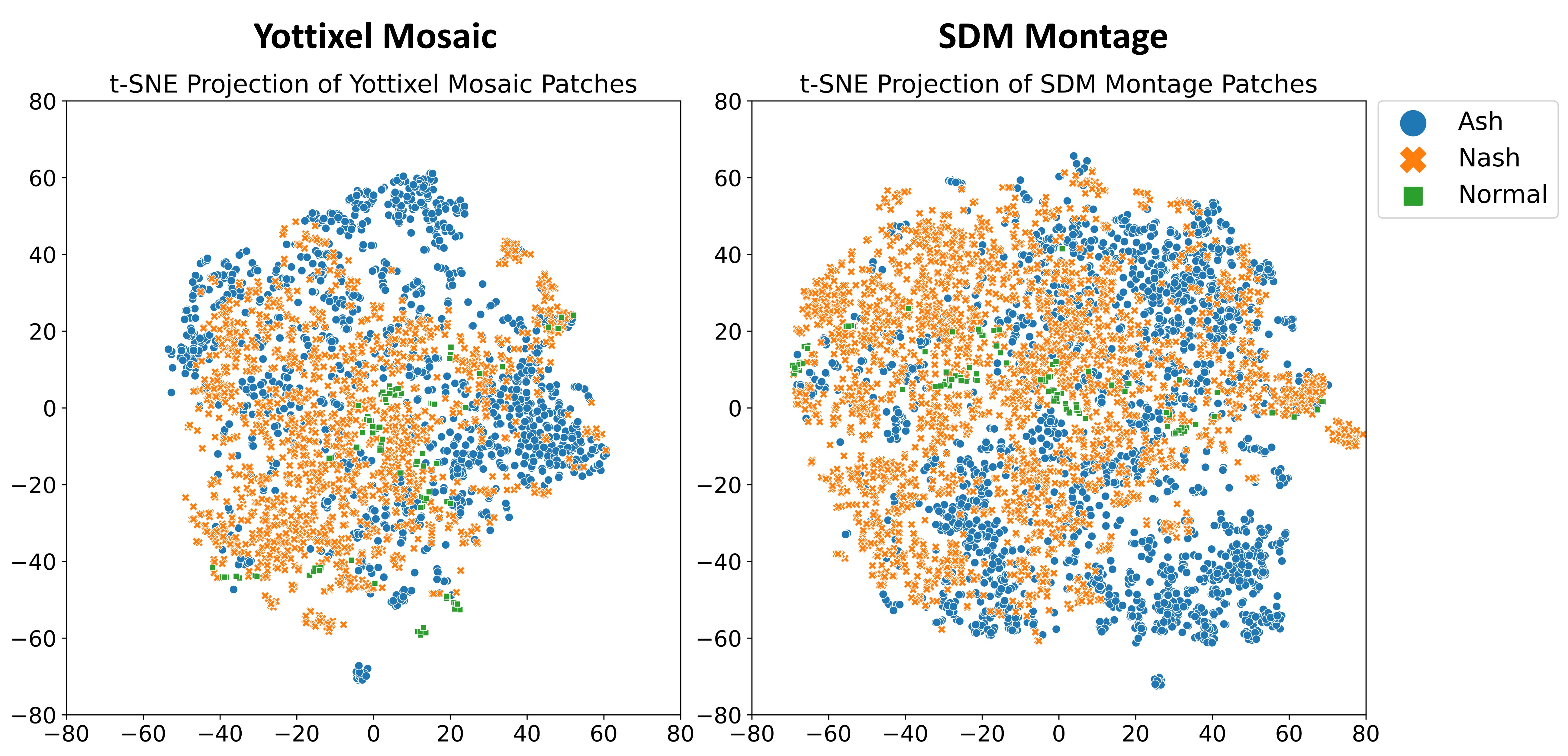}}
\caption{The t-SNE projection displays the embeddings of all patches extracted from the Liver dataset using Yottixel's mosaic (left) and SDM's montage (right).}
\label{fig:Liver_TSNE}
\end{figure*}

To assess the effectiveness of the SDM montage in comparison to Yottixel's mosaic, we conducted a leave-one-out evaluation to retrieve the most similar cases using the Liver dataset. The evaluation criteria encompass multiple retrieval scenarios, including the top-1, MV@3, and MV@5 retrievals. The results, including accuracy, macro average, and weighted average scores at the top-1, MV@3, and MV@5 levels, are presented in Figure~\ref{fig:Liver_Accuracy}. Table~\ref{tab:Liver_results} shows the detailed statistical results including precision, recall, and f1-score. Moreover, Confusion matrices and chord diagrams at Top-1, MV@3, and MV@5 retrievals are shown in Figure~\ref{fig:Liver_Chord1}, \ref{fig:Liver_Chord3}, and ~\ref{fig:Liver_Chord5}, respectively. In addition to these accuracy metrics, a comparative analysis of the number of patches extracted per WSI by each respective method is also presented in Figure~\ref{fig:Liver_Patch_boxplot} for a visual representation of the distribution over the entire dataset. To visually illustrate the extracted patches, we used t-SNE projections, as demonstrated in Figure~\ref{fig:Liver_TSNE}.

\begin{table*}[t]
\centering
\resizebox{\textwidth}{!}{\begin{tabular}{l|l|lll|lll|lll|l}
\hline
&  & \multicolumn{3}{c|}{\textbf{Top-1}}    & \multicolumn{3}{c|}{\textbf{MV@3}}     & \multicolumn{3}{c|}{\textbf{MV@5}} &  \\ \hline
& \begin{tabular}[c]{@{}l@{}}Primary\\ Diagnoses\end{tabular} & Precision & Recall & f1-score & Precision & Recall & f1-score & Precision & Recall & f1-score & Slides \\ \hline
\multirow{3}{*}{\rotatebox[origin=c]{0}{\textbf{\begin{tabular}[c]{@{}c@{}}Yottixel\\ Mosaic~\cite{kalra2020Yottixel}\end{tabular}}}} 
& Ash  & 0.81  & 0.73   & 0.76     & 0.87  & 0.73   & 0.80  & 0.89 & 0.73   & 0.81 & 150  \\
& Nash  & 0.72  & 0.85   & 0.78     & 0.74  & 0.91   & 0.81 & 0.74 & 0.93   & 0.83 & 158    \\
& Normal  & 0.75 & 0.19   & 0.30     & 1.00  & 0.25   & 0.40 & 1.00 & 0.19   & 0.32 & 16    \\ \hline
&\textbf{Total Slides}&&&&&&&&&&  \textbf{324}\\ \hline
\multirow{3}{*}{\rotatebox[origin=c]{0}{\textbf{\begin{tabular}[c]{@{}c@{}}SDM\\ Montage\end{tabular}}}}  
& Ash     & 0.84  & 0.72 &  0.78   & 0.87  & 0.73 & 0.79   &   0.87  &  0.76   & 0.81 &   150    \\
& Nash  & 0.71  & 0.88 &  0.79    & 0.73  & 0.90 &  0.81   & 0.75    &  0.91   &  0.82 &  158      \\ 
& Normal  & 1.00   & 0.17 &  0.29  & 1.00   & 0.28 &  0.43  & 1.00   &  0.22    &  0.36 &   18     \\  \hline
&\textbf{Total Slides}&&&&&&&&&&  \textbf{326}\\ \hline
\end{tabular}}
\caption{Precision, recall, f1-score, and the number of slides processed for each sub-type are shown in this table using Yottixel mosaic, and SDM montage. The evaluations are based on the top 1 retrieval, the majority among the top 3 retrievals, and the majority among the top 5 retrievals in the Liver dataset.}\label{tab:Liver_results}
\end{table*}

In our empirical assessments, the SDM approach displayed performance metrics closely aligned with the Yottixel mosaic. This similarity in performance was especially pronounced in the MV@-3 and MV@-5 retrieval outcomes. Notably, there was an enhancement of +1\% in the macro-average of F1-scores when employing the SDM technique. The nuanced differences and advantages of the SDM approach over the Yottixel mosaic in specific retrieval scenarios are further elucidated in the referenced Figure~\ref{fig:Liver_Accuracy}. From an accuracy standpoint, the SDM method exhibited a marginal improvement of one percentage point for top-1 retrieval. Nonetheless, its performance remained largely analogous to that of the Yottixel mosaic when evaluated at MV@3 and MV@5 retrieval metrics as seen in Figure~\ref{fig:Liver_Accuracy}. Moreover, our observations have unveiled an intriguing aspect of Yottixel's behavior in contrast to SDM. It shows that Yottixel processed a total of 324 WSIs, while SDM successfully processed all 326 WSIs.  From a detailed examination of the box plot presented in Figure~\ref{fig:Liver_Patch_boxplot}, it becomes evident that for fine needle biopsies — a predominant category within the Liver dataset — the Yottixel 5\% strategy tends to opt for fewer patches relative to the SDM method.  

\clearpage\subsection{Private -- Breast Cancer (BC)}
\label{sec:Exd_Results_BC}
Breast tumor slides were acquired from patients at our hospital. There are 16 different subtypes of breast tumors were employed in this experiment. All of the biopsy slides were digitized at $40\times$ magnification and linked to their respective diagnoses at the WSI level (see Table~\ref{tab:BC_Mayo} for more details).
\begin{table}[t] 
\centering
\resizebox{1\columnwidth}{!}{\begin{tabular}{llr}
\hline
\begin{tabular}[c]{@{}l@{}}Primary Diagnoses\end{tabular} & Acronyms & Slides \\ \hline
Adenoid Cystic Carcinoma & ACC     & 3     \\
Adenomyoepthelioma       & AME    & 4     \\
Ductal Carcinoma In Situ   & DCIS   & 10   \\
\begin{tabular}[c]{@{}l@{}}Ductal Carcinoma In Situ, - \\Columnar Cell Lesions Including - \\ Flat Epithelial Atypia, - \\ Atypical Ductal Hyperplasia\end{tabular}  & DCIS, CCLIFEA, ADH  & 3  \\
Intraductal Papilloma, Columnar Cell Lesions  & IP, CCL   & 3 \\
Invasive Breast Carcinoma of No Special Type & IBC NST  & 3  \\
Invasive Lobular Carcinoma   & ILC  & 3  \\ 
\begin{tabular}[c]{@{}l@{}}Lobular Carcinoma In Situ + \\Atypical Lobular Hyperplasia\end{tabular} & LCIS + ALH& 2\\
\begin{tabular}[c]{@{}l@{}}Lobular Carcinoma In Situ, - \\Flat Epithelial Atypia, - \\ Atypical Lobular Hyperplasia\end{tabular} & LCIS, FEA, ALH&2\\
Malignant Adenomyoepithelioma &MAE&4\\
Metaplastic Carcinoma&MC&5\\
Microglandular Adenosis&MGA&2\\
Microinvasive Carcinoma&MIC&2\\
Mucinous Cystadenocarcinoma&MCC&5\\
Normal Breast&Normal&21\\
Radial Scar Complex Sclerosing Lesion&RSCSL&2\\
\hline
\end{tabular}}
\caption{Detailed information related to the BC dataset, inclusive of the respective acronyms and the number of slides associated with each primary diagnosis.}\label{tab:BC_Mayo}
\end{table}

\begin{figure*}[t]
\centerline{\includegraphics[width =  1\textwidth]{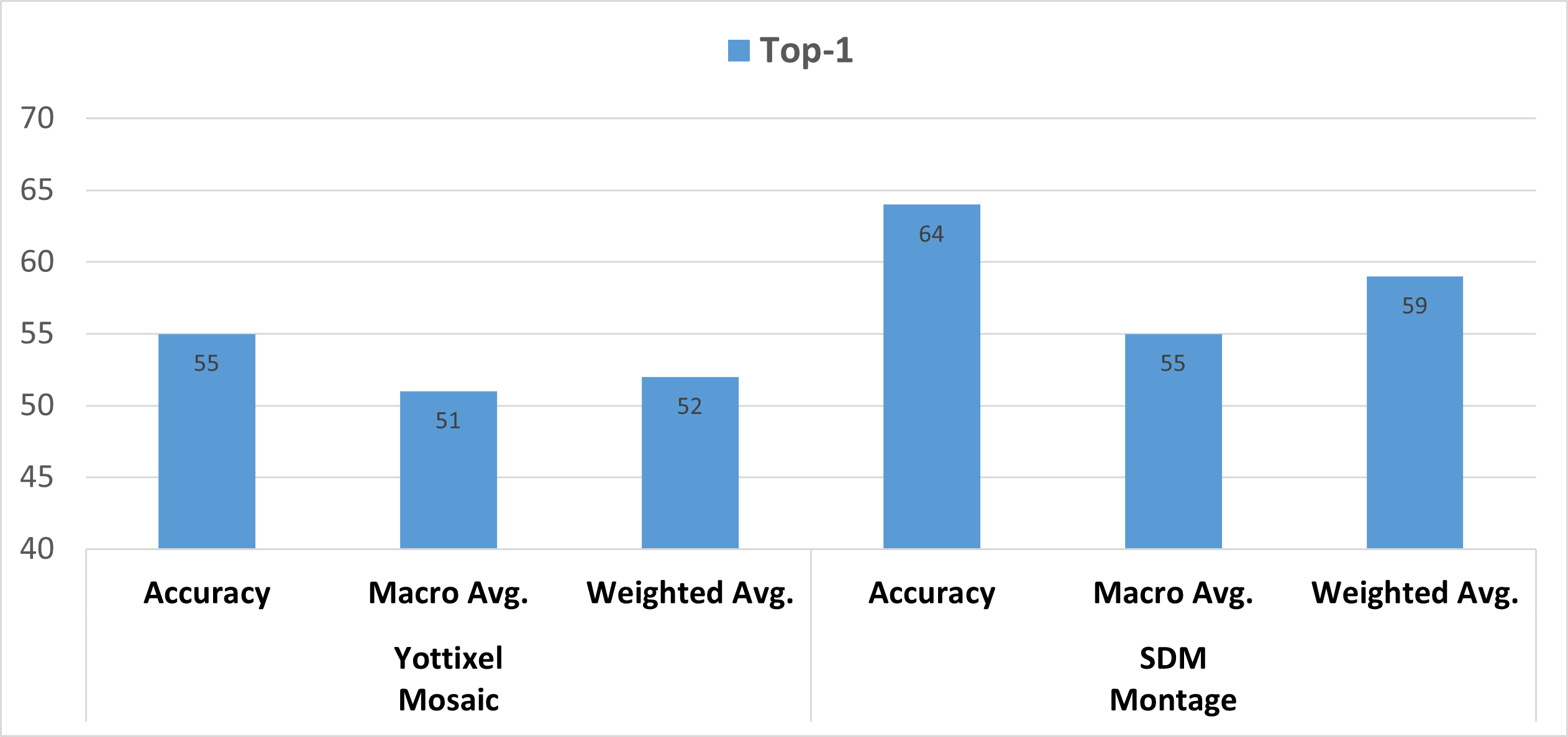}}
\caption{Accuracy, macro average of f1-scores, and weighted average of f1-scores are shown from Yottixel mosaic, and SDM montage. The evaluations are based on the top 1 retrieval in the Breast Cancer dataset.}
\label{fig:BC_Accuracy}
\end{figure*}

\begin{figure*}[t]
\centerline{\includegraphics[width =  0.6\textwidth]{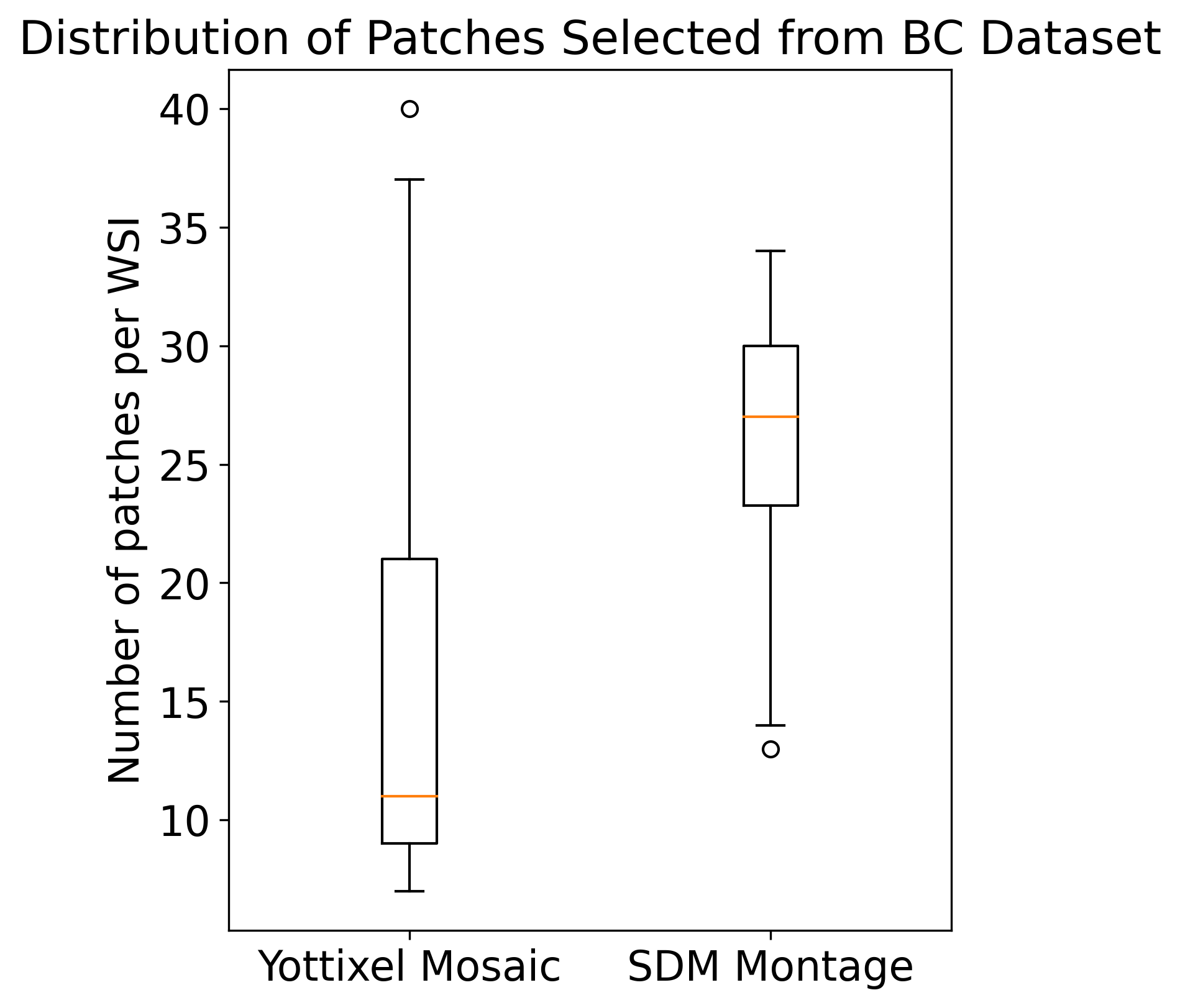}}
\caption{The boxplot illustrates the distribution of patches selected for each WSI in the Breast Cancer (BC) dataset from both the Yottixel Mosaic and SDM Montage. Additionally, it provides statistical measures for these distributions. Specifically, for the Yottixel Mosaic, the median number of selected patches is $11\pm9$. Conversely, for the SDM Montage, the median number of selected patches is $27\pm5$.}
\label{fig:BC_Patch_boxplot}
\end{figure*}

\begin{figure*}[t]
\centerline{\includegraphics[width =  1\textwidth]{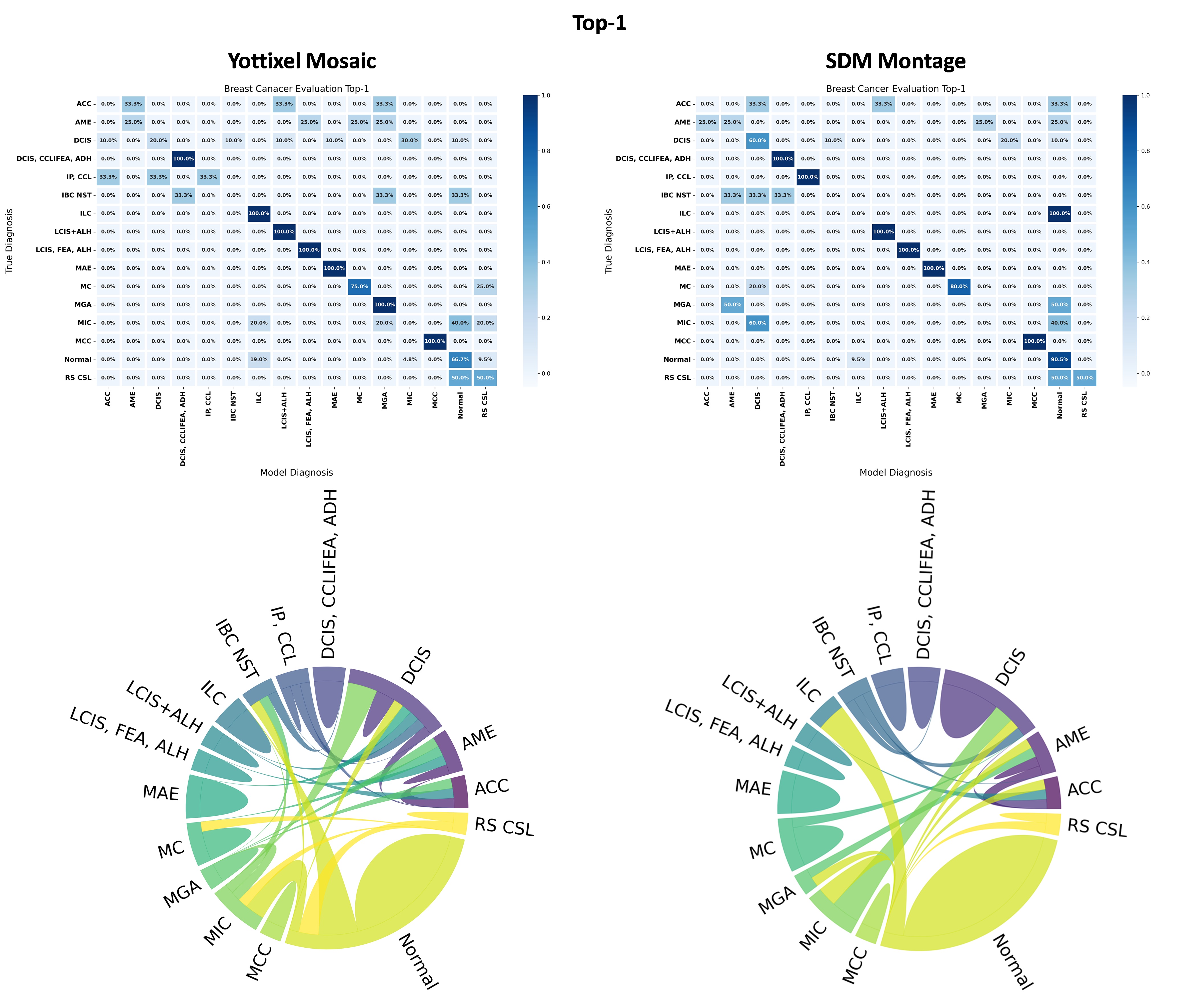}}
\caption{Confusion matrices and chord diagrams from Yottixel mosaic (left column), and SDM montage (right column). The evaluations are based on the top 1 retrieval from the BC dataset.}
\label{fig:BC_Chord1}
\end{figure*}

\begin{figure*}[t]
\centerline{\includegraphics[width =  1\textwidth]{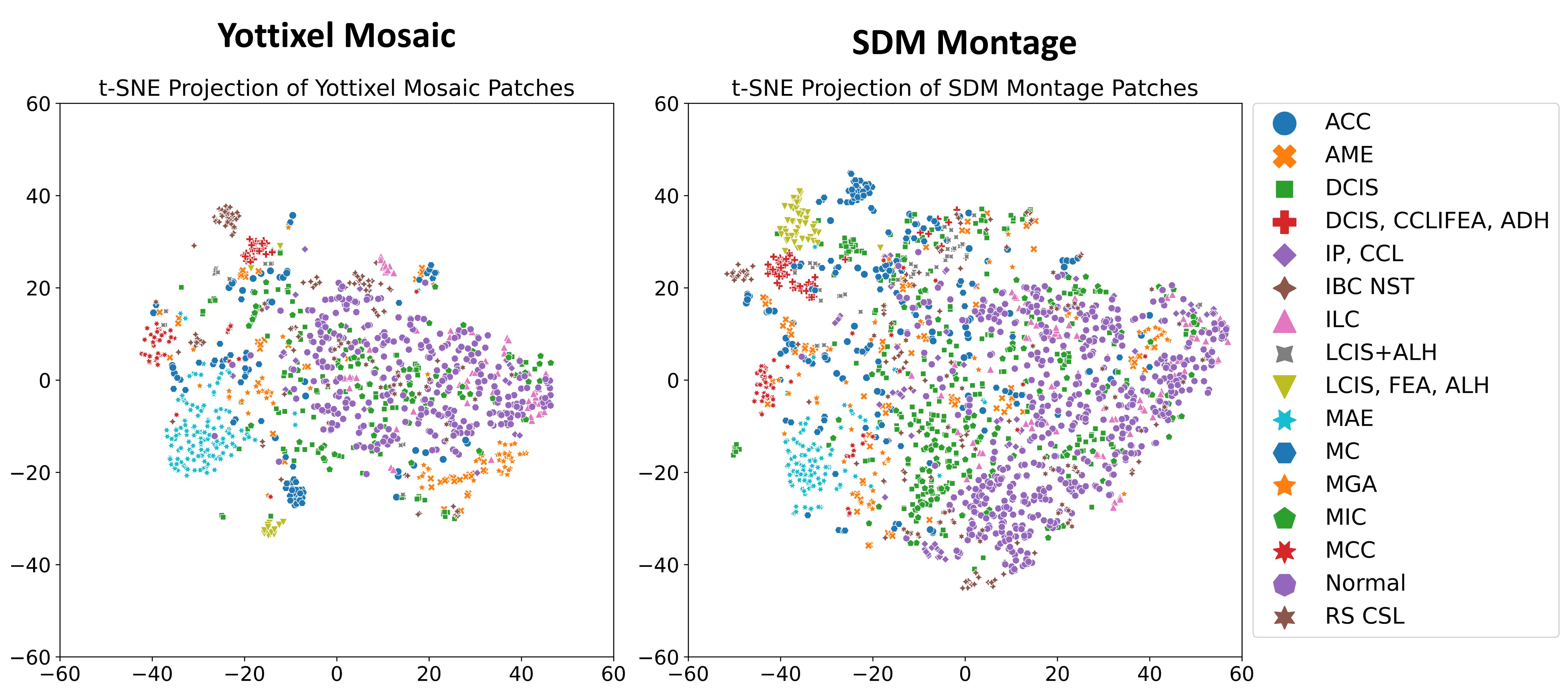}}
\caption{The t-SNE projection displays the embeddings of all patches extracted from the BC dataset using Yottixel's mosaic (left) and SDM's montage (right).}
\label{fig:BC_TSNE}
\end{figure*}

To assess the performance of the SDM's montage against Yottixel's mosaic, we conducted a leave-one-out evaluation to retrieve the most similar cases using the BC dataset. The evaluation criteria encompass the top-1 retrieval. The results, including accuracy, macro average, and weighted average scores at the top-1 are presented in Figure~\ref{fig:BC_Accuracy}. Table~\ref{tab:BC_results} shows the detailed statistical results including precision, recall, and f1-score. Moreover, Confusion matrices and chord diagram at top-1 are shown in Figure~\ref{fig:BC_Chord1}. In addition to these accuracy metrics, a comparative analysis of the number of patches extracted per WSI by each respective method is also presented in Figure~\ref{fig:BC_Patch_boxplot} for a visual representation of the distribution over the entire dataset. To visually illustrate the extracted patches, we used t-SNE projections, as demonstrated in Figure~\ref{fig:BC_TSNE}.

Our experimental findings showcased the superior performance of SDM, particularly evident in the top-1 retrieval result by +9\% in accuracy, +4\% in macro avg. of f1-scores, and +7\% in weighted average as illustrated in Figure~\ref{fig:BC_Accuracy}. Furthermore, our observations shed light on an intriguing aspect of Yottixel's behavior in comparison to SDM. Specifically, it has come to our attention that Yottixel displays a proclivity for overlooking certain WSIs within the dataset. To elaborate, our analysis reveals that Yottixel processed a total of 73 WSIs, whereas SDM demonstrated a more comprehensive approach by successfully processing all 74 WSIs. This observation underscores the robustness and completeness of the SDM method in handling the entire dataset, further emphasizing its merits in WSI analysis and retrieval applications.

\begin{table*}[t]
\resizebox{\textwidth}{!}{\begin{tabular}{l|llll|llll}
\hline
                                                            & \multicolumn{4}{c|}{\textbf{Yottixel Mosaic~\cite{kalra2020Yottixel}}} & \multicolumn{4}{c}{\textbf{SDM Montage}} \\ \hline
                                                            & \multicolumn{4}{c|}{\textbf{Top-1}}           & \multicolumn{4}{c}{\textbf{Top-1}}       \\ \hline
\begin{tabular}[c]{@{}l@{}}Primary Diagnoses\end{tabular} & Precision   & Recall   & f1-score   & Slides  & Precision  & Recall  & f1-score & Slides \\ \hline
ACC                                                         & 0.00        & 0.00     & 0.00       & 3       & 0.00       & 0.00    & 0.00     & 3      \\
AME                                                         & 0.50        & 0.25     & 0.33       & 4       & 0.33       & 0.25    & 0.29     & 4      \\
DCIS                                                        & 0.67        & 0.20     & 0.31       & 10      & 0.50       & 0.60    & 0.55     & 10     \\
DCIS, CCLIFEA, ADH                                          & 0.75        & 1.00     & 0.86       & 3       & 0.75       & 1.00    & 0.86     & 3      \\
IP, CCL                                                     & 1.00        & 0.33     & 0.50       & 3       & 1.00       & 1.00    & 1.00     & 3      \\
IBC NST                                                     & 0.00        & 0.00     & 0.00       & 3       & 0.00       & 0.00    & 0.00     & 3      \\
ILC                                                         & 0.38        & 1.00     & 0.55       & 3       & 0.00       & 0.00    & 0.00     & 3      \\
LCIS + ALH                                                  & 0.50        & 1.00     & 0.67       & 2       & 0.67       & 1.00    & 0.80     & 2      \\
LCIS, FEA, ALH                                              & 0.67        & 1.00     & 0.80       & 2       & 1.00       & 1.00    & 1.00     & 2      \\
MAE                                                         & 0.80        & 1.00     & 0.89       & 4       & 1.00       & 1.00    & 1.00     & 4      \\
MC                                                          & 0.75        & 0.75     & 0.75       & 4       & 1.00       & 0.80    & 0.89     & 5      \\
MGA                                                         & 0.33        & 1.00     & 0.50       & 2       & 0.00       & 0.00    & 0.00     & 2      \\
MIC                                                         & 0.00        & 0.00     & 0.00       & 5       & 0.00       & 0.00    & 0.00     & 5      \\
MCC                                                         & 1.00        & 1.00     & 1.00       & 2       & 1.00       & 1.00    & 1.00     & 2      \\
Normal                                                      & 0.74        & 0.67     & 0.70       & 21      & 0.66       & 0.90    & 0.76     & 21     \\
RSCSL                                                       & 0.20        & 0.50     & 0.29       & 2       & 1.00       & 0.50    & 0.67     & 2      \\ \hline
\textbf{Total Slides}                                       &             &          &            & 73      &            &         &          & 74     \\ \hline
\end{tabular}}
\caption{Precision, recall, f1-score, and the number of slides processed for each sub-type are shown in this table using Yottixel mosaic, and SDM montage. The evaluations are based on the top 1 retrieval in the Breast Cancer dataset.}\label{tab:BC_results}
\end{table*}

\end{document}